\documentclass{CUP-JNL-EDS}

\usepackage{latexsym}
\usepackage{graphicx}
\usepackage{physics}
\usepackage{anyfontsize}
\usepackage{placeins}
\usepackage{multicol,multirow}
\usepackage{amsmath,amssymb,amsfonts}
\usepackage{mathrsfs}
\usepackage{amsthm}
\usepackage{rotating}
\usepackage{appendix}
\usepackage[authoryear]{natbib}
\usepackage{ifpdf}
\usepackage[T1]{fontenc}
\usepackage{gensymb}
\usepackage{nicefrac}
\usepackage[caption=false,font=normalsize,labelfont=sf,textfont=sf]{subfig}
\usepackage{newtxmath}
\usepackage{textcomp}%
\usepackage{xcolor}%
\usepackage{hyperref}
\usepackage{lipsum}

\mathchardef\mhyphen="2D 

\begin{document}
\newcommand\hyph{\mathop{SPAC\mhyphen}}
\begin{Frontmatter}

\title[Article Title]{TemperatureGAN: Generative Modeling of Regional Atmospheric
Temperatures}

\author*[1]{Emmanuel Balogun}\orcid{0000-0002-8514-4315}\email{ebalogun@stanford.edu}
\author[2]{Ram Rajagopal}
\author*[1]{Arun Majumdar}\email{amajumdar@stanford.edu}

\address[1]{\orgdiv{Mechanical Engineering}, \orgname{Stanford University}, \orgaddress{\city{Stanford}, \state{CA},  \country{USA}}}

\address*[2]{\orgdiv{Civil and Environmental Engineering}, \orgname{Stanford University}, \orgaddress{\city{Stanford}, \state{CA},  \country{USA}}}




\authormark{Balogun et al.}


\abstract{Stochastic generators are useful for estimating climate impacts on various sectors. Projecting climate risk in various sectors, e.g., energy systems, requires generators that are accurate (statistical resemblance to ground-truth), reliable (do not produce erroneous examples), and efficient. Leveraging data from the North American Land Data Assimilation System, we introduce TemperatureGAN, a Generative Adversarial Network conditioned on months, regions, and time periods, to generate 2m above ground atmospheric temperatures at an hourly resolution. We propose evaluation methods and metrics to measure the quality of generated samples. We show that TemperatureGAN produces high-fidelity examples with good spatial representation and 
temporal dynamics consistent with known diurnal cycles.} 
\end{Frontmatter}
%
\section[Introduction]{Introduction}
Over the past decade, the growing impacts of climate change have been disproportionate among regions and populations \cite{Diffenbaugh_Burke_2019}. The frequency and severity of extreme events are impacted by climate change and pose a major threat to ecosystems, power systems, agriculture, and people \citep{liang_wu_chambers_schmoldt_gao_liu_liu_sun_kennedy_2017, henry_pratson_2017, somanathan_somanathan_sudarshan_tewari_2021}. For power systems, outages and curtailments can last for days because of events like heat waves or wildfires impacting generation and transmissions units. In \cite{thiery_et_al}, the authors estimate that the generation born in 2020 will experience a two to seven fold increase in heat waves compared to people born in 1960. It has been shown that inequities can be spatially and temporally heterogeneous and are correlated with socioeconomic factors like race. For instance, \cite{brockway_conde_callaway_2021}, show that in California, regions with more black population have lower solar grid hosting capacities, thus potentially limiting access to solar PV, which can act as a key home-level resiliency tool. 

As the power sector transitions to cleaner energy technologies, fueled by the large-scale deployment renewable energy resources, there is a massive opportunity to make the grid more resilient for everyone. Thus, it is important to identify who are most at risk to the effects of climate change. It is very challenging to accurately (spatially and temporally) forecast specific occurrences and attributes of significant weather events more than a couple days to weeks in advance, which makes preparing for these events challenging. 

General Circulation Models (GCMs) provide spatially coarse examples of plausiblefuture climate states but lack the spatial resolution needed to study local phenomena. Characterizing the tails of temperature distributions requires generating many examples, as rarer events may occur only once in many samples, making computationally efficient sampling paramount. Stochastic weather generators (SWG) are commonly used for generating statistically plausible examples of weather data \citep{semenov2008simulation}. SWGs commonly comprise statistical models (Markov Chains, Gaussian fits, exponential fits, etc.) that are fit to observational weather data; \cite{wilks1999weather} talks in detail about some of these models and applications.  LARS-WG \citep{semenov2002stochastic}, a popular stochastic weather generator based on the series weather generator \citep{RACSKO199127}, was created to simulate data at a single site, generating maximum and minimum daily temperatures, but not at hourly resolutions. Generating atmospheric conditions at hourly resolutions can be useful for grid reliability studies.

This motivates a method that can provide spatially and temporally resolved (hourly resolution) examples of atmospheric conditions that are physically realistic (i.e. reasonably fall within or near historically observed trends) to a specific region and time; we develop such an approach utilizing Generative Adversarial Networks (GANs) \citep{GAN}. A GAN is a deep learning (DL) generative modeling framework in which two groups of Artificial Neural Networks (ANN) are trained continuously in a repeated sequence. It comprises the discriminator/critic networks which are trained to distinguish between true and fake samples and the generator networks whose goal is to produce samples that come from the same distribution as the true samples, or resemble the true samples. As both networks train over time, the critic learns to be better at distinguishing between fake and true samples, while the generator produces fake samples that are harder to distinguish. Mathematically, this results in a minimax expression with both the critic and generator networks trying to optimize conflicting objectives.  



Deep generative models have gained prominence due to their ability to capture complex distributions without explicit statistical parameterization, yielding methods that generally outperform their more explicit counterparts which rely on stronger prior assumptions on the data distribution; the authors in \cite{buechler2021evgen} show an example of this. SWGs have been leveraged by various research communities for decades now, thus, developing new methods that complement or improve on these models will drive the research community forward. In this paper, we propose an approach to modelling conditional regional temperatures utilizing GANs and propose metrics for evaluation.

The rest of this paper is organized as follows. In section \ref{problem}, we describe the problem, state our contributions, propose a framework for approaching the problem, and discuss some related works. In section \ref{methods}, we describe the methodology in detail, including data treatment, model architecture, and training. In section \ref{experiments}, we show results from experiments and we evaluate of the model's outputs, and finally, we conclude in section \ref{conclude}, with outlook on future work.

\section{Problem statement}\label{problem}
Understanding the distribution of impacts of temperatures on smaller regions or specific communities requires data at spatial resolutions finer than what GCMs currently provide, and the plausible states derived from GCMs do not provide the temporal resolution that some studies i.e. a power distribution system reliability or resilience study, will require. Additionally, because GCMs produce snapshots of the entire earth at once, it can be very inefficient and memory intensive to produce enough samples to study smaller regions of interest. In this work, we aim to:
\begin{enumerate}
    \item  Motivate and develop an an approach for utilizing GANs to reliably generate daily surface temperature maps at an hourly temporal resolution at low computational cost.
    \item Show that our model can be leveraged for period, month, and region-based sampling by conditioning the model on those priors during training time; to the best of our knowledge, no other work has used GANs in the regime for month and region-based sampling.
    \item Propose metrics and methods to evaluate our approach specifically for this task and measure performance and validate the quality of the generative model's outputs empirically. The metrics can be leveraged in other domains for evaluating generative models of similar outputs.
\end{enumerate}

GANs are most commonly used for image generation and have seen significant development in conditional and controllable generation \citep{DCGAN, bigGAN, styleGAN} and more nascent video generation as in \cite{clark_donahue_simonyan_2019} and \cite{xia_mitchell_ek_sheffield_cosgrove_wood_luo_alonge_wei_meng_et}. In this work, we aim to produce 24-hour temperature maps, which can be viewed as a video generation task. 
We propose a conditional GAN that can generate (2m above ground) atmospheric temperatures, conditioned on region, month, and time period. This can potentially provide inputs to other impact assessment models \citep{gleick1987development, moriondo2011climate, semenza2012climate, siebert2014impact} that estimate the downstream impacts of weather and the changing climate, by capturing region-based distribution of temperatures in a generative fashion. We display the overarching framework in Figure \ref{fig:model_framework} below.

\begin{figure}[h]
    \centering
    \includegraphics[width=0.8\linewidth]{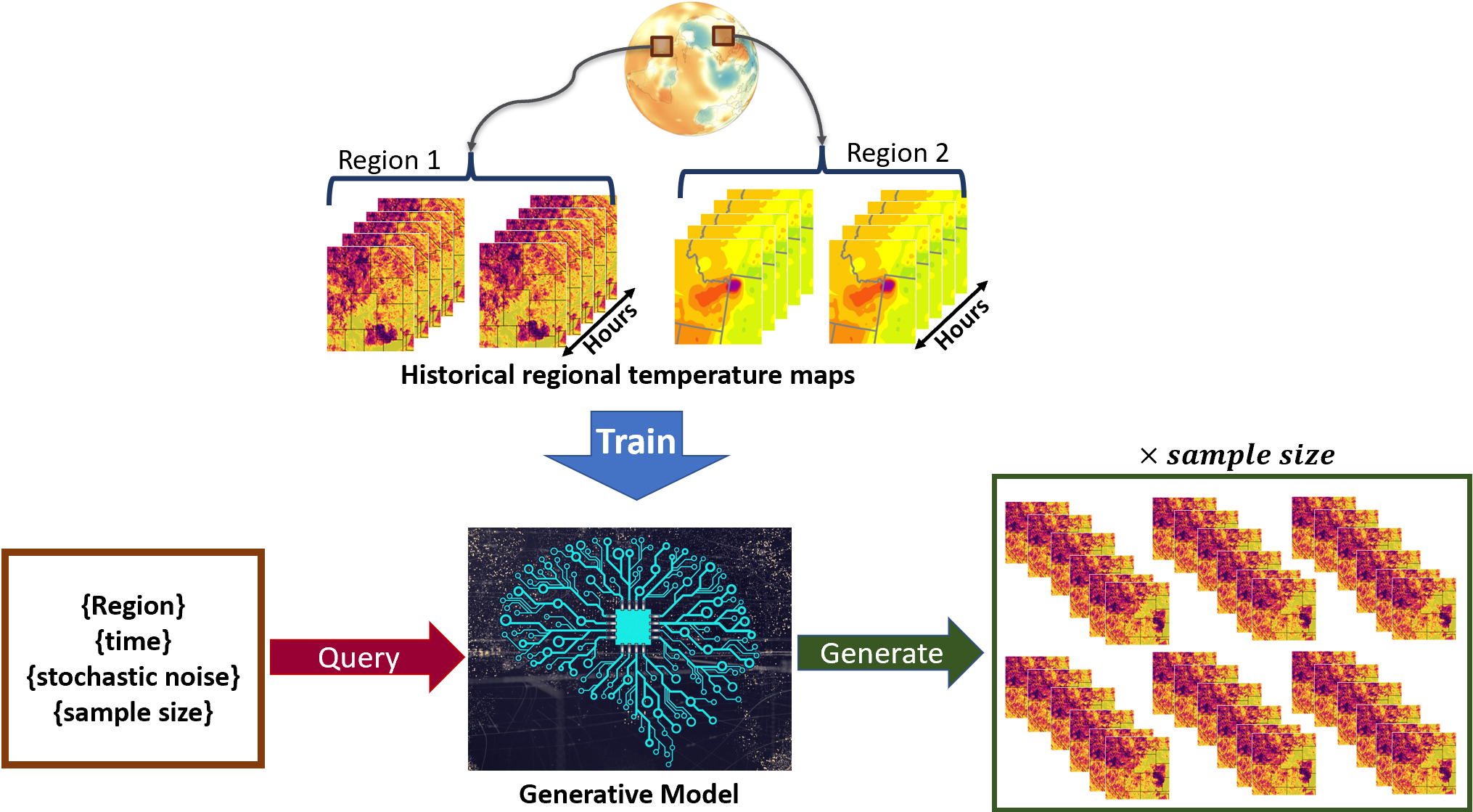}
    \caption{TemperatureGAN model framework. Regional temperature maps are passed as input to discriminator during the training, however the generator never sees the training data}
    \label{fig:model_framework}
\end{figure}

\subsection{Related works}
In recent years, there has been meaningful progress in applying deep learning techniques to problems related to weather and climate. The authors in \cite{bihlo2021generative} use an ensemble of GANs to predict the future (one year) weather using the European Centre for Medium-Range Weather Forecasts (ECMWF) ERA5 ($0.25\degree \times 0.25\degree$ spatial and 3-hour time resolution) reanalysis data from the prior 4 years, and evaluate their models using root-mean-squared error (RMSE), anomaly correlation coefficients (ACC), and the continuous ranked probability scores \citep{zamo2018estimation}, which are commonly used to evaluate forecasts. \cite{meng2021physics} develop a physics-informed GAN for sea subsurface temperature prediction at $0.25\degree \times 0.25\degree$ spatial and daily mean resolutions. In \cite{keisler}, the authors utilize Graph Neural Networks (GNN) for global weather forecasts. The model uses local information to step the current 3D atmospheric state forward by six hours, and show results that rivals state of the art forecasting models such as the ECMWF model, while reducing computational cost; some months later, \cite{lam2022graphcast} (also GNN) and \cite{bi2022pangu} introduced global weather forecasting models that surpassed the state of the art physics-based numerical weather prediction model on most metrics. In \cite{bhatia2021exgan}, the authors adopt a gradual distribution shifting and resampling approach to model extreme precipitation, but they evaluate the outputs using Fr\'echet Inception Distance (FIT) proposed by \cite{heusel_ramsauer_unterthiner_nessler_klambauer_hochreiter_2017} which does not explain the temporal veracity of the generated rainfall (distribution). Additionally, we suggest that FIT is not appropriate for understanding the model's performance on generating physical variables, because the intermediate layer from which scores are obtained does not produce values with units that have an easily discernible physical meaning. In \cite{DeepClimGAN}, the authors use a GAN to model average daily temperatures from the Coupled Model Intercomparison Project Phase (CMIP) data, taking an autoregressive approach, and evaluate their model by visually comparing the histogram plot of true and generated data, which can be useful, but is not rigorous and can be biased by the human eye. Additionally, they do not produce samples at an hourly resolution, but rather a daily average temperature, which may not be sufficient for some downstream applications (i.e. reliability assessments). They also acknowledge that more work needs to be done to thoroughly evaluate their GAN. In \cite{npg-28-347-2021}, the authors use a GAN to model the daily mean climate variables at a 2.8\degree ~resolution and propose some approaches to evaluation, which include Principal Component Analysis (PCA), Wasserstein distance, and visual inspection, recognizing that traditional methods for evaluating GANs for natural image synthesis community may not suffice. More recently, \cite{izumi2022super} leverage existing GANs for super-resolution of sea surface temperature, and evaluate the models by comparing the outputs with high resolution optimum interpolation sea surface temperature (OISST), using the learned perceptual image patch similarity (LPIPS) and RMSE as metrics. 

Although recent work has shown that GANs have the potential for modeling climate variables, utilizing GANs in this regime is still in its infancy. Consequently, there is a dearth of intuitive and consistent evaluation metrics/benchmarks specific to this application. Evaluation metrics that can be adopted by machine learning (ML), power, and climate communities, are critical for comparing generative models and their outputs, especially for researchers working at the confluence of these fields.

\section[main]{Methodology}\label{methods}
TemperatureGAN's task is analogous to video generation; we aim to produce 2D spatial maps of temperatures that iterate through multiple time steps. Consequently, such spatial maps are amenable to estimating the probability, $\hat{P}(T|R,M,k)$, where $\{T, R, M, k\}$ represent temperature, region, month, and period. This can be relevant for other downstream applications, such as empirically estimating power system resilience over a whole utility service territory under plausible ambient conditions \cite{billinton1999application, billinton2006predicting, saraiva1996generation, li2013reliability}. Though they have shown reliability and accuracy for weather forecasting, physics-based models are computationally expensive, thus, cannot efficiently create ensembles with enough members for simulating multiple realizations rapidly.

\subsection[model]{Data}
Direct station observations (e.g. the Automated Surface Observation System in the US) are reliable, accurate, and can have decades of historical records, but are spatially sparse. Satellites can provide spatially gridded observations of some atmospheric variables, but data may not exist for long time periods. Reanalysis datasets blend these sources together along with forecasting/simulation models to produce spatially and temporally consistent maps of climate variables. The dataset used for this work is the Mosaic Land Surface Model Forcing Temperatures data from the North American Land Data Assimilation System (NLDAS) \citep{xia_mitchell_ek_sheffield_cosgrove_wood_luo_alonge_wei_meng_et}. The dataset is well-suited for this task as it provides fine spatial and temporal resolution relevant to power systems studies. NLDAS (a collaborative effort between NOAA, NASA, and others) integrates satellite and surface-level observational data with reanalysis datasets and includes multiple surface state and flux variables in North America from 1979 to near-present day. This data has a $0.125 \degree \times 0.125\degree$ ($13.8 \times 11$ km) spatial resolution (taken from an equidistant cylindrical projection) at an hourly timescale and has a size of about 700GB for the contiguous US. Training using data at a $0.125 \degree \times 0.125\degree$ resolution, aids the possibility for a model to capture fine spatiotemporal dependencies that a physics-driven model cannot without a costly downscaling effort. We limit the scope of this work to the United States (US) West Coast region due to computational resources, and regions without data are zero-padded. We aggregate the raw data both spatially and temporally, as single grid point measurements are not sufficient for generating distributions with a high degree of confidence.

\subsubsection{Spatial Data Aggregation}
To aggregate the data spatially, $1\degree\times1\degree$ ($111 \times 88$ km) grids are grouped as a single labelled region; region $R$ is a grid of 64 grid points ($8\times8$) from the original dataset. This aggregation increases the number of samples for each region compared to the original NLDAS grid. This implicitly assumes that samples from each $1\degree \times 1\degree$ region comes from a conditional distribution $p(T|R)$. regions are indexed based on their relative (integer) positions from the SW corner of the dataset which corresponds to a position of (1,1). The GAN is conditioned on these relative integer positions during training. Note that this data aggregation may limit the downstream application for this model. For example, Independent System Operators (ISO) will cover large regions, usually larger than $1\degree \times 1\degree$, thus studying power risks to weather at that scale using this spatial extent will not be straightforward. Nonetheless, it is possible to model spatial extents larger $1\degree \times 1\degree$ as this will mainly modify the data-engineering step.

\subsubsection{Temporal Data Aggregation}
\begin{figure}[h]
    \centering
    \includegraphics[width=0.6\linewidth]{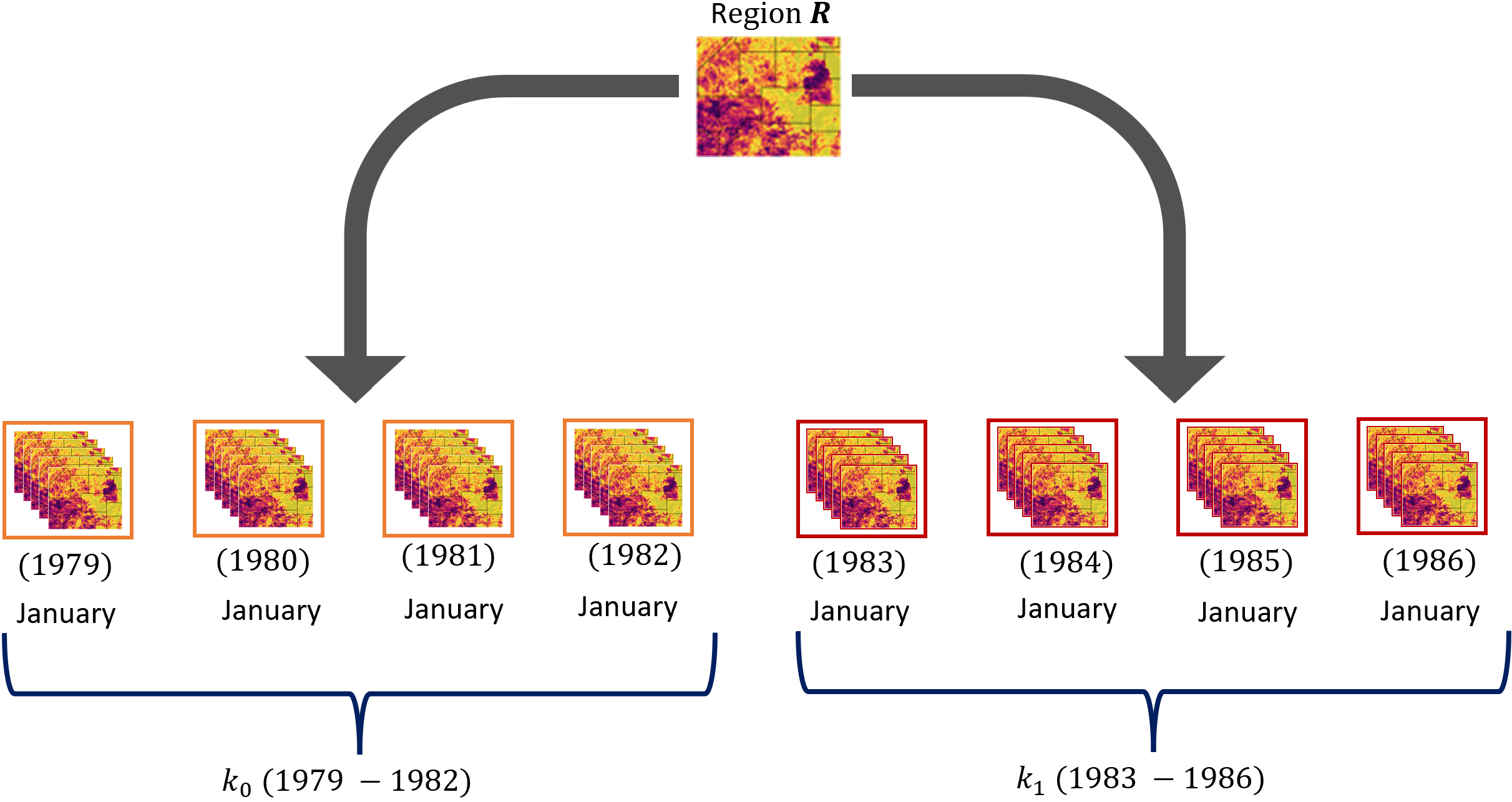}
    \caption{Image depiction of data aggregation scheme for a specific region $R$ and for the chosen month $M$ of January. For a 4-year period $k_i$, all the 24-hour time-series (31 for each year because January has 31 days) within all four months of January are grouped into the same bucket (making it a total of 124 examples) and have the same labels}
    \label{fig:temp_agg}
\end{figure}

Temporally, we posit that temperature distributions are non-stationary over sufficiently long periods \cite{donat2012shifting}. The data is aggregated such that each example is a 24-hour (daily) temperature map; one can imagine each example as a video with 24 frames. We introduce the idea of periods—a period is a stipulated number of years for which one can assume that the overall climate does not change significantly. To make this concrete, we aggregate the data into 24-hour daily time-series, then group all the 24-hour time series by their respective months, and finally, the same month within the elected period is also thrown into the same bucket.  In this work, we elect 4-year periods. This makes the implicit assumption that for a 4-year period $k_i$, at a given region $R$, and month, $M$, the diurnal cycles come from the same distribution, or are independently and identically distributed (IID). For example, if the entire historically observed record spans 1979 to present day, then the first period, $k_0$, will encompass observations from 1979--1982, inclusive, the second period $k_2$, spans 1983--1986 inclusive, the third period $k_3$ spans 1987--2000, etc. Selecting quadrennial (4-year) periods has not been rigorously justified, as one might elect 1-, 2-, or 5-year periods instead. The choice of number of years within a period constitutes a design trade-off. If there are too few years (for example, one year period) then the modelling assumption departs too far from known climatology and will be difficult to evaluate empirically with confidence. If there are too many years, one might unintentionally average out temporal effects or years/periods of significant temperature distribution shifts or dilation---we find 4 years to be a reasonable choice and defer more thorough investigation to future work. Figure \ref{fig:temp_agg} describes the temporal data aggregation scheme.

Adopting $k_i$, where $i\in\{0, 1, 2, ..., n\}$, offers the ability to smooth over inter-year macro scale climate events (e.g. El Ni\~no) in the ground-truth observations and capture medium/longer term trends. Another important motivation for this modeling decision is that it lends itself well to evaluation, because by aggregating over regions and years we have more data per sample, making parametric and non-parametric goodness-of-fit tests for evaluating the generated samples from the model more tractable.

\subsection{Model Architecture and training}
Video generation \citep{clark_donahue_simonyan_2019, saito_matsumoto_saito_2017, wang2020imaginator, chu2020learning, gur2020hierarchical, gupta2022rv} is one of the most challenging GAN applications. In \cite{clark_donahue_simonyan_2019}, the authors propose a Dual-Video-Discriminator to handle the memory bottleneck for video datasets (terming the model DVD-GAN), using two discriminators---a spatial and temporal discriminator. Spatial discriminator $D_s$ inspects an individual video frame (a static image) for texture quality and \textit{temporal discriminator} $D_t$ for penalizes the generated frame-by-frame transitions. For TemperatureGAN, we also leverage two discriminator/critic networks, but take a different approach from DVD-GAN. We build a Convolutional Neural Network (CNN)-based temporal discriminator $D_t$, and rather than training on videos (in our case a video is a spatio-temporal time series of temperature values) our model is trained on \textit{temporal gradients}, distinguishing it from DVD-GAN. Training on the temporal gradients separately guides the model to focus on the learning the daily (hourly) diurnal cycles and to produce hourly (or temporal) temperature transitions that are consistent with the ground-truth's diurnal cycles. Thus, for a given 24-hour sample, we have 23 gradients $\frac{\partial T}{\partial t}$. We use the Wasserstein loss with a gradient penalty (GP) \citep{GAN-Improved} as a soft constraint to satisfy the 1-Lipschitz continuity \citep{gouk2021regularisation}. We also experimented with directly constraining the layer weights via spectral normalization introduced in \cite{spec_norm} to satisfy the 1-Lipschitz continuity conditions and found it to perform poorly, thus was not further pursued. The loss functions are:

\begin{equation}
     L_{D, temporal} = \underset{\tilde{\mathbf{T}}\sim\mathbb{P}_{g}}{\mathbb{E}}\left[D_{t}\left(\mathbf{\frac{\partial \tilde{T}}{\partial t}}\right)\right] - \underset{\mathbf{T}\sim\mathbb{P}_{r}}{\mathbb{E}}\left[D_{t}\left(\mathbf{\frac{\partial T}{\partial t}}\right)\right] +\lambda_{GP}\underset{\hat{\mathbf{T}}\sim\mathbb{P}_{\hat{\mathbf{T}}}}{\mathbb{E}}\left[\left(\norm{\nabla_{\hat{\mathbf{T}}}D_t\left(\mathbf{\frac{\partial{\hat{T}}}{\partial{t}}}\right)}_{2}-1\right)^2\right]
     \label{L_temporal}
\end{equation}

\begin{equation}
     L_{D, spatial} = \underset{\tilde{\mathbf{T}}\sim\mathbb{P}_{g}}{\mathbb{E}}\left[D_{s}\left(\mathbf{\tilde{T}}\right)\right] - \underset{\mathbf{T}\sim\mathbb{P}_{r}}{\mathbb{E}}\left[D_{s}\left(\mathbf{T}\right)\right] +\lambda_{GP}\underset{\hat{\mathbf{T}}\sim\mathbb{P}_{\hat{\mathbf{T}}}}{\mathbb{E}}\left[\left(\norm{\nabla_{\hat{\mathbf{T}}}D_s(\hat{\mathbf{T}})}_{2}-1\right)^2\right]
     \label{L_spatial}
\end{equation}

\begin{equation}
    L_{G} = \underset{\tilde{\mathbf{T}}\sim\mathbb{P}_g}{{-\mathbb{E}}\left[D_{s}\left(\mathbf{\tilde{T}}\right)\right] - {\mathbb{E}}\left[D_{t}\left(\mathbf{\frac{\partial \tilde{T}}{\partial t}}\right)\right]}
    \label{G_loss}
\end{equation}

$\mathbf{\tilde{T}}\sim\mathbb{P}_g$ represents examples sampled from the GAN (the generator) and $\mathbf{T}\sim\mathbb{P}_r$ represents examples sampled from the real (observed) data. The discriminators/critics and generator seek to minimize their respective losses. Observe that the first two terms in Equations \eqref{L_temporal} and \eqref{L_spatial} instruct the discriminator D to maximize the gap between the expected values of the true and fake samples for the temperature gradients and the temperature values—this is done by minimizing the loss functions in \eqref{L_temporal} and \eqref{L_spatial}. The objective of the discriminators $D_t$ and $D_s$ is to minimize Equations \eqref{L_temporal} and \eqref{L_spatial}, because by minimizing these losses, it is encouraged to assign higher scores to true (observed) data and lower scores to generated (“fake”) examples. However, the objective of the generator $G$ is to minimize Equation \eqref{G_loss}, which means it is encouraged to produce examples that will yield high scores from the discriminator, suggesting that it attempts to produce samples that resemble those from the observed (ground-truth) data, thereby implicitly estimating the true probability distribution $\mathbb{P}_r$. The final terms in Equations \eqref{L_temporal} and \eqref{L_spatial} highlight an important concept regarding the training stability of Wasserstein GANs (WGANs), which is the notion of Lipschitz continuity. For training stability, the loss functions of the discriminator should be 1-Lipschitz continuous for the Wasserstein distance approximation to be valid; this constrains how quickly the models’ parameters can change during training. In other words, enforcing Lipschitz continuity ensures that the Discriminator's loss does not grow too quickly such that the Generator cannot learn. $\nabla_{\mathbf{\hat{T}}}D(\mathbf{\hat{T}})$ is the gradient of the discriminator's outputs with respect to its inputs, which are temperature maps. We follow a similar convention as in \cite{GAN-Improved}, where the authors define $\mathbb{P}_{\mathbf{\hat{T}}}$ as sampling uniformly along straight lines between pairs of points sampled from the data distribution $\mathbb{P}_{r}$ and the generator distribution $\mathbb{P}_g$. $\lambda_{GP}$ is a hyperparameter for weighting how much importance the model places on the final terms representing Lipschitz continuity in Equations \eqref{L_temporal} and \eqref{L_spatial}; we elect $\lambda_{GP}=1$ for training. 

\begin{figure}[h!]
    \centering
    \includegraphics[width=0.8\linewidth]{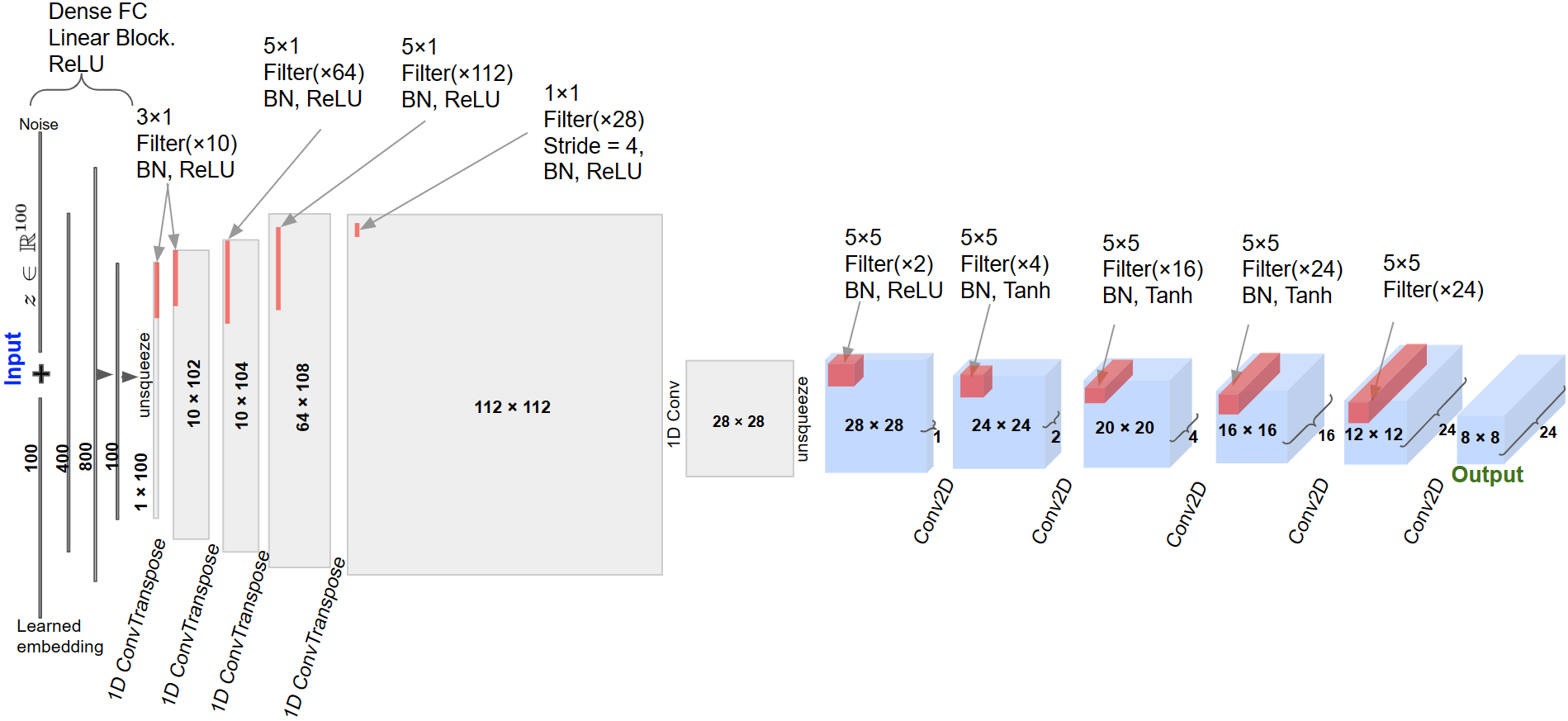}
    \caption{Generator $G$ architecture. The sampled noise is concatenated with the learned label embedding and passed through a dense, fully-connected (FC) linear block with Rectified Linear Unit (ReLU) activation functions. The output of the linear block is sent through series of convolution layers with batch normalization to obtain the desired ($8\times8\times24$) output shape}
    \label{fig:G_arch}
\end{figure}

\begin{figure}[h!]
    \centering
    \includegraphics[width=0.8\linewidth]{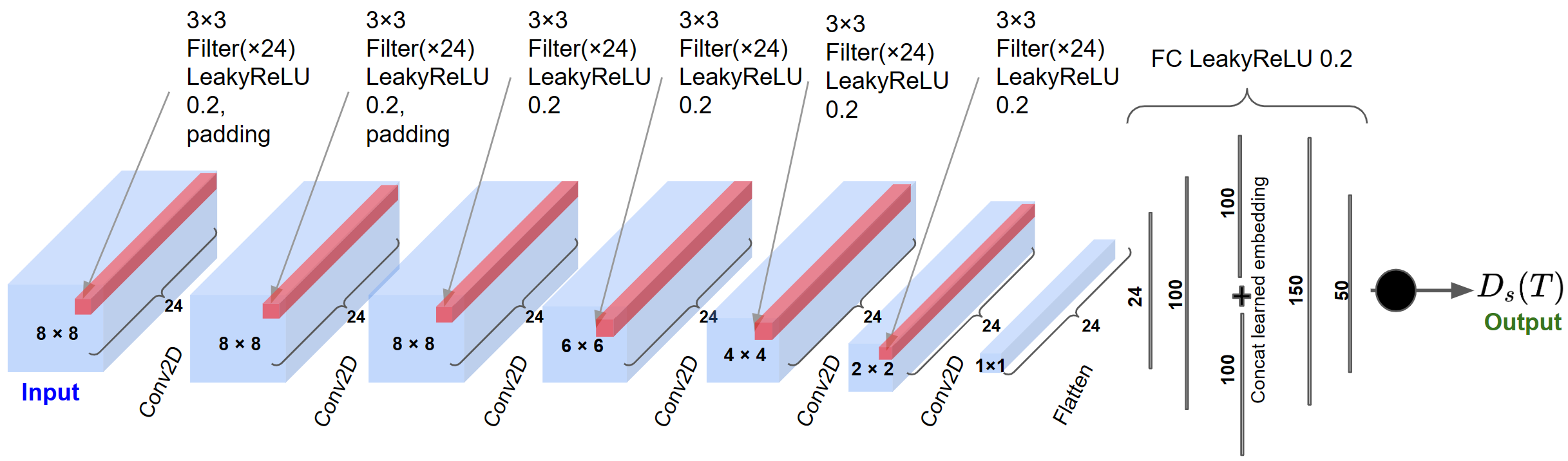}
    \caption{Spatial discriminator $D_{s}$ architecture. The inputs to the spatial discriminator $D_s$, are the $8\times8\times24$ temperature maps. The input is then passed through a series 2D convolution layers with the output of the prior layer serving as input into the following layer. The final convolution layer outputs a 2-dimensional $24\times1$ vector, which is flattened into a one-dimensional vector before it is passed through a dense FC linear block to produce a score}
    \label{fig:Ds_arch}
\end{figure}

\begin{figure}[h!]
    \centering
    \includegraphics[width=0.8\linewidth]{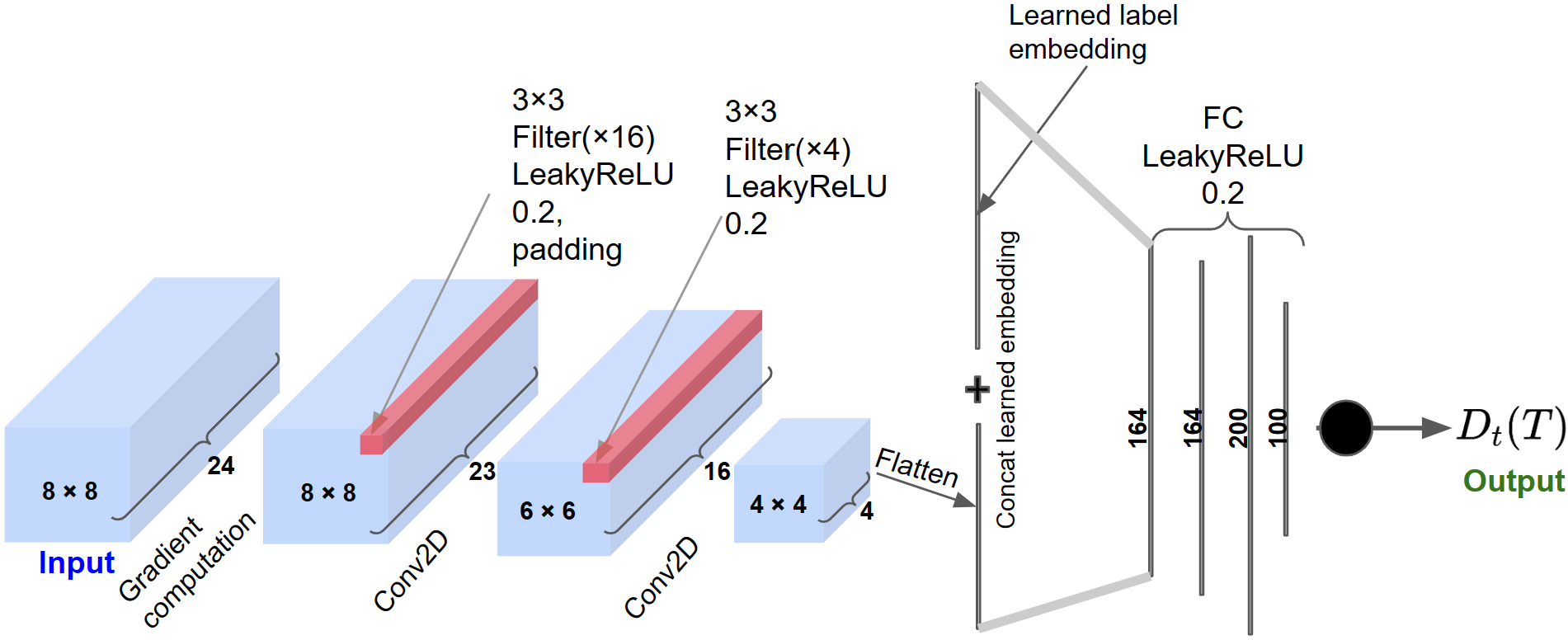}
    \caption{Temporal discriminator $D_{t}$ architecture. The inputs to the temporal discriminator $D_t$, are the $8\times8\times24$ temperature maps and then a gradient computation is followed, to compute $\frac{\partial{T}}{\partial{t}}$. The output from the convolution layers is flattened and passed through a series of fully-connected (FC) linear layers to produce a score}
    \label{fig:Dt_arch}
\end{figure}

Figures \ref{fig:G_arch}, \ref{fig:Ds_arch}, and \ref{fig:Dt_arch} show the architectures of the Generator and discriminator neural networks. They are convolution-based networks.


\FloatBarrier

The generator is an arguably modest 562,206 parameter model. We normalize and standardize the data and train the GAN for 2000 epochs using a batch size of 4096 and ADAM optimizer \citep{kingma2014adam} ($\beta_1 = 0.5, \beta_2 = 0.99$) for gradient descent, with an exponential learning rate (LR) decay every 100 epochs. We train on the first 8 periods (1979-2010), about 2.7 million examples, where each example is 3-dimensional (lon $\times$ lat $\times$ time). The model takes 4 days to train for 2000 epochs on a single 80GB NVIDIA A100 GPU. Because we do not explicitly constrain the network outputs (constraining the output would imply we know the upper bound on temperature values), it is important to use activation functions that are bounded, especially at later layers of the generator $G$. We observed that using only ReLU/LeakyReLU layers in the generator could yield implausibly extreme samples which was rectified by the choosing more bounded activation functions (i.e $\tanh$).

\section{Experiments and evaluation}\label{experiments}
\begin{figure}[]
    \centering
    \includegraphics[width=0.7\linewidth]{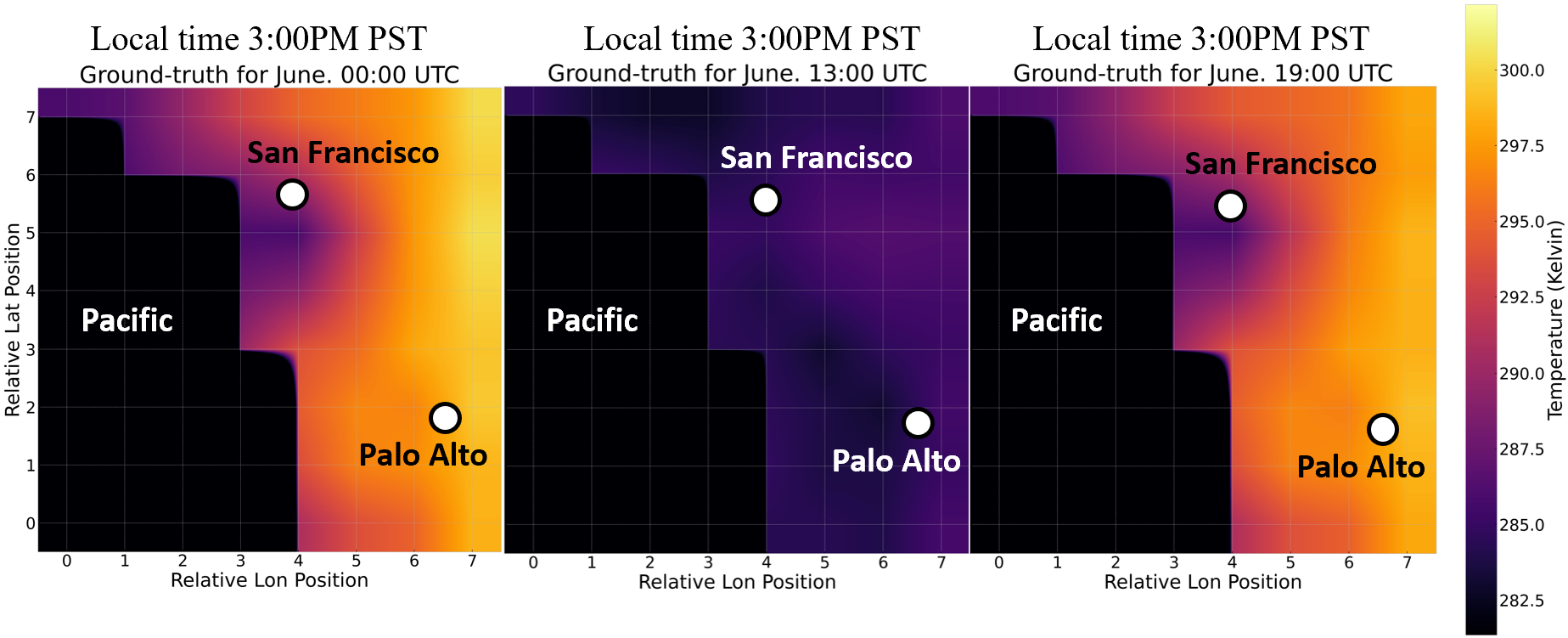}
    \includegraphics[width=0.7\linewidth]{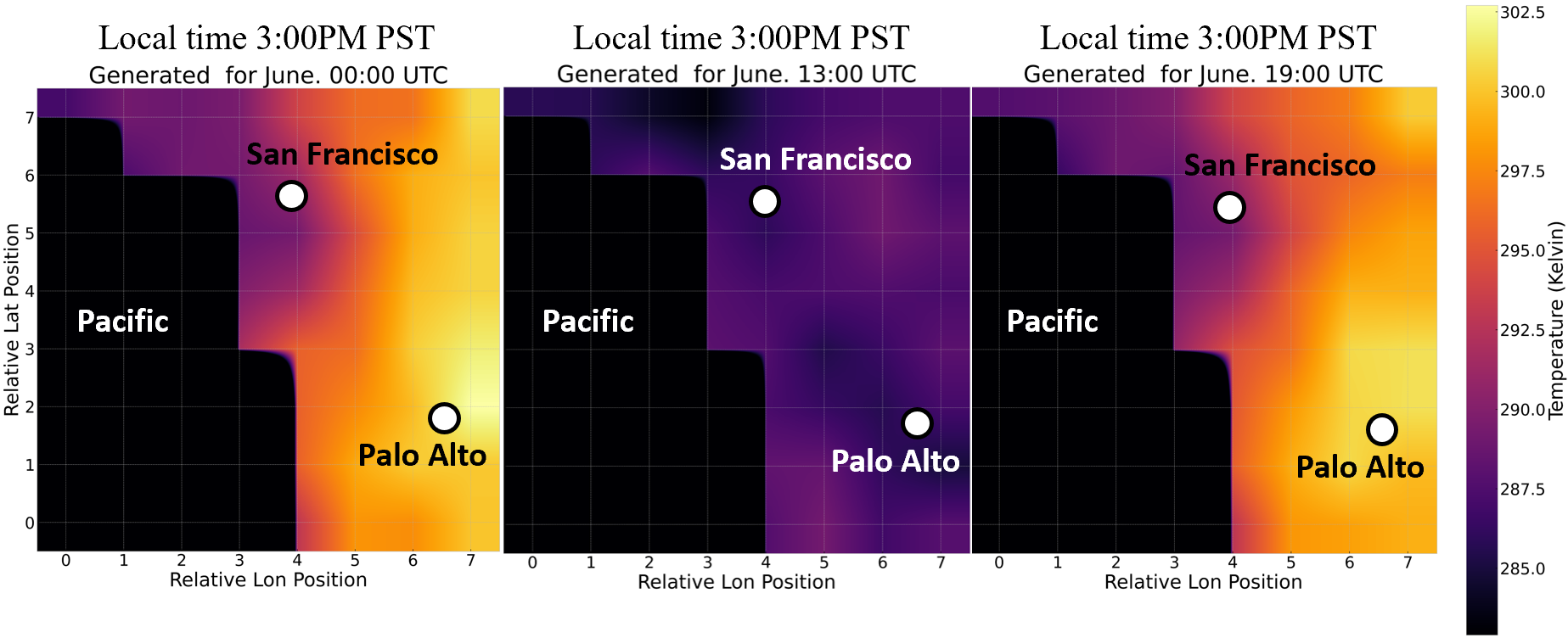}
    \caption{Ground-truth (top) and generated (bottom) hourly snapshots of samples of a summer day in California Bay Area (2011-2014)}
    \label{fig:PA_real_vs_generated}
\end{figure}

\begin{figure}[]
    \centering
    \includegraphics[width=0.7\linewidth]{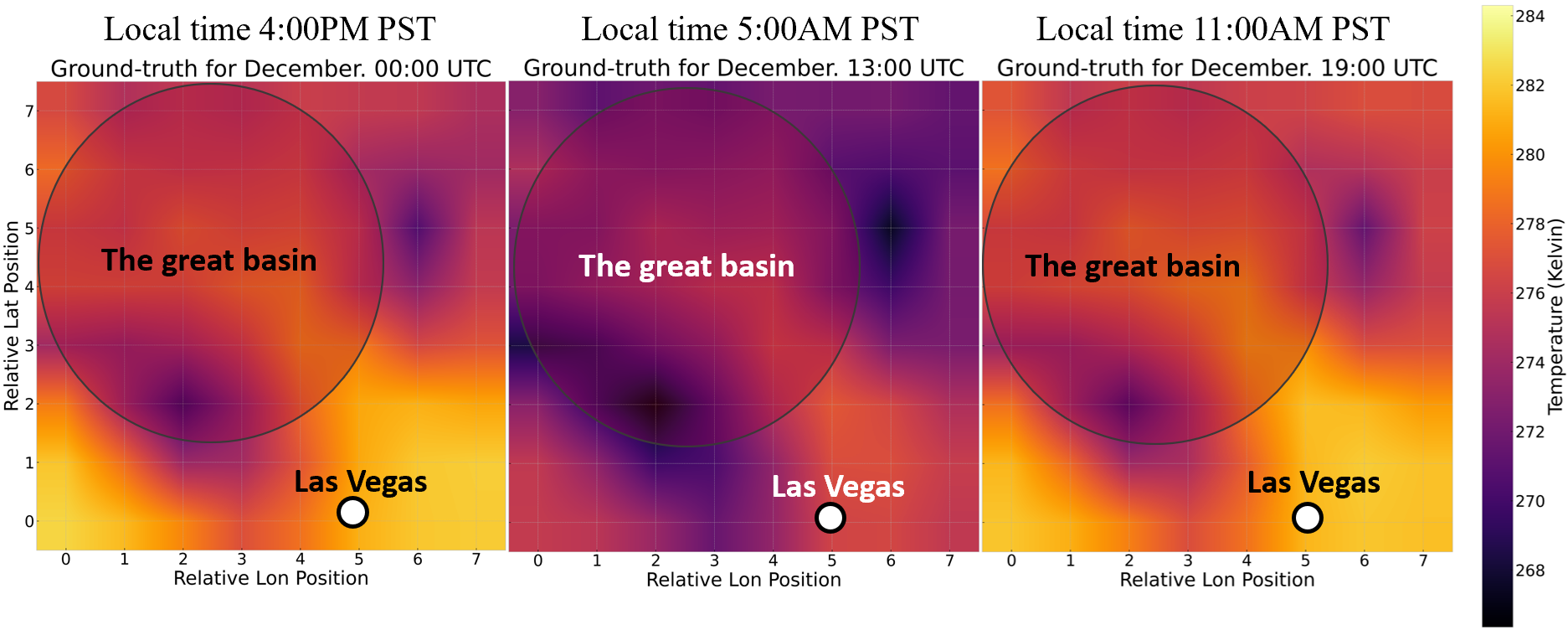}
    \includegraphics[width=0.7\linewidth]{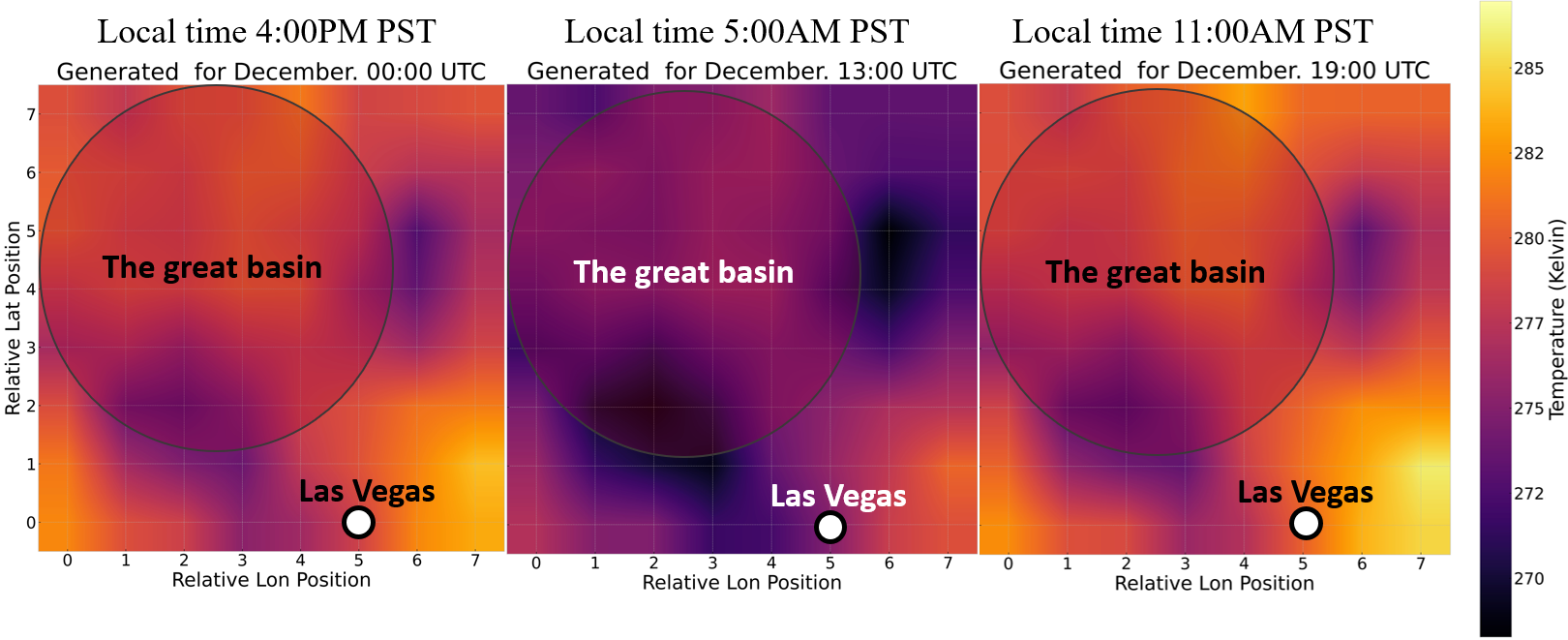}
     \caption{Ground-truth (top) and generated (bottom) hourly snapshots of samples of a winter day in Nevada. (2011-2014)}
    \label{fig:NV_real_vs_generated}
\end{figure}

\begin{figure}[h!]
    \centering
    \includegraphics[width=0.35\linewidth]{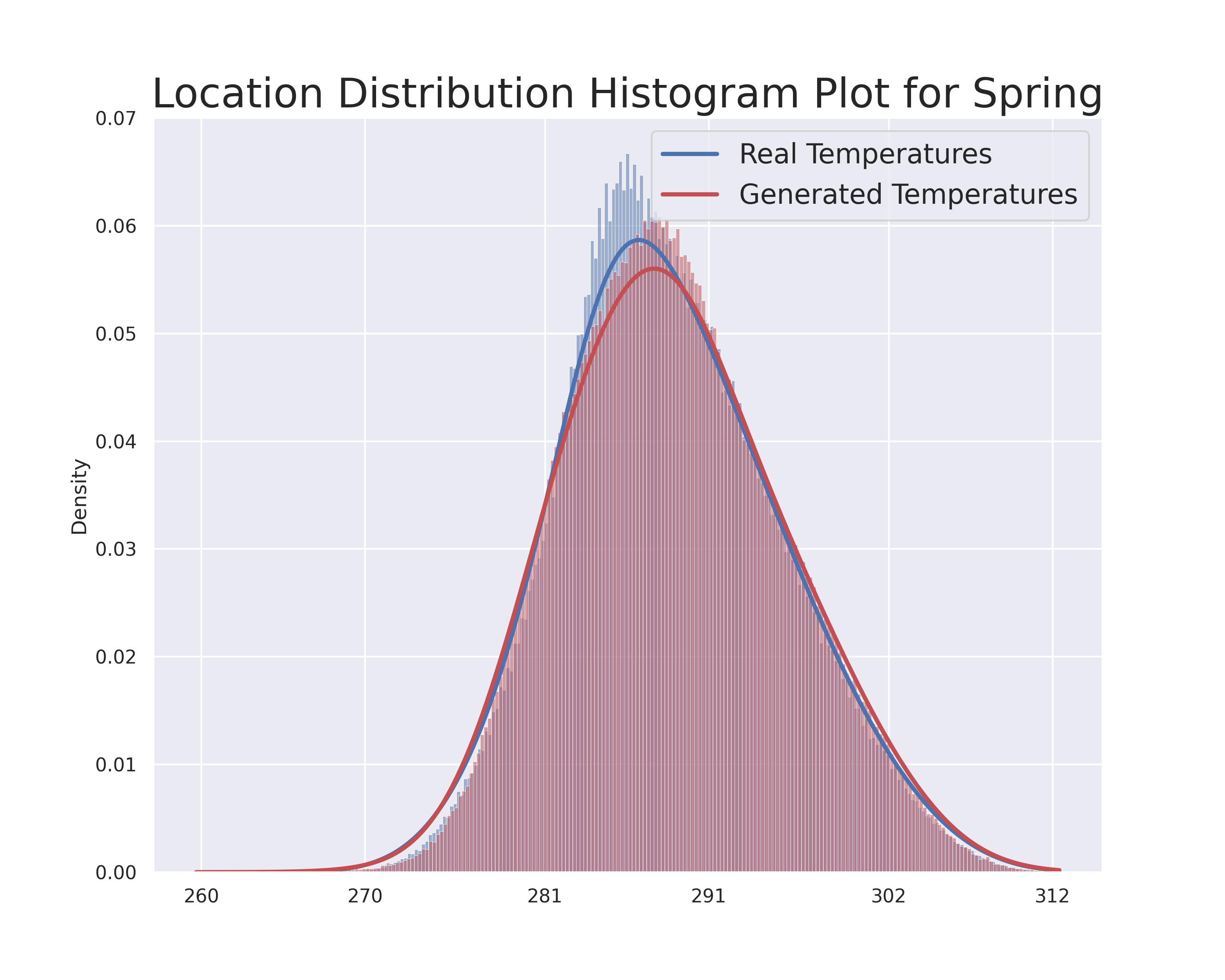}
    \includegraphics[width=0.35\linewidth]{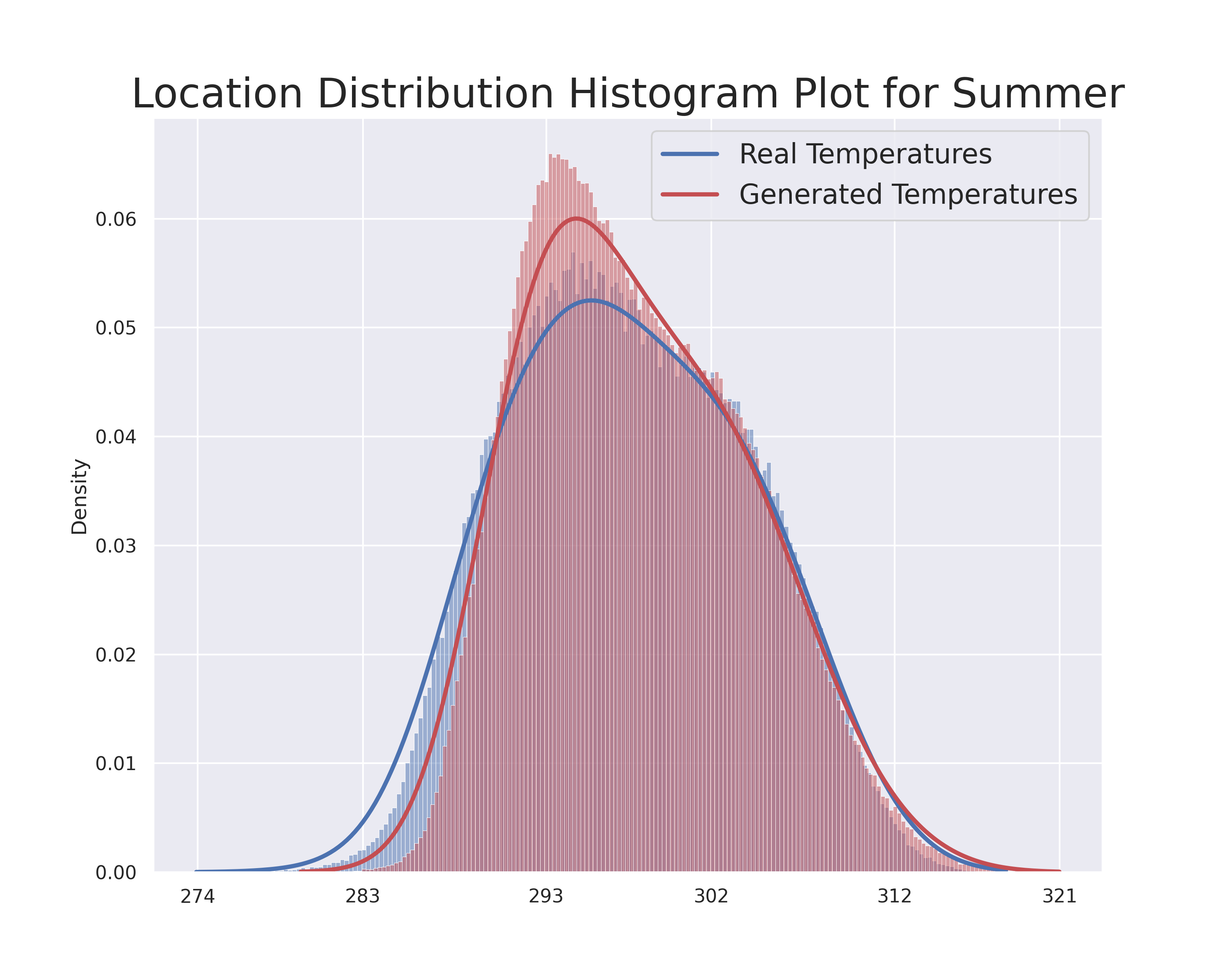}
    \includegraphics[width=0.35\linewidth]{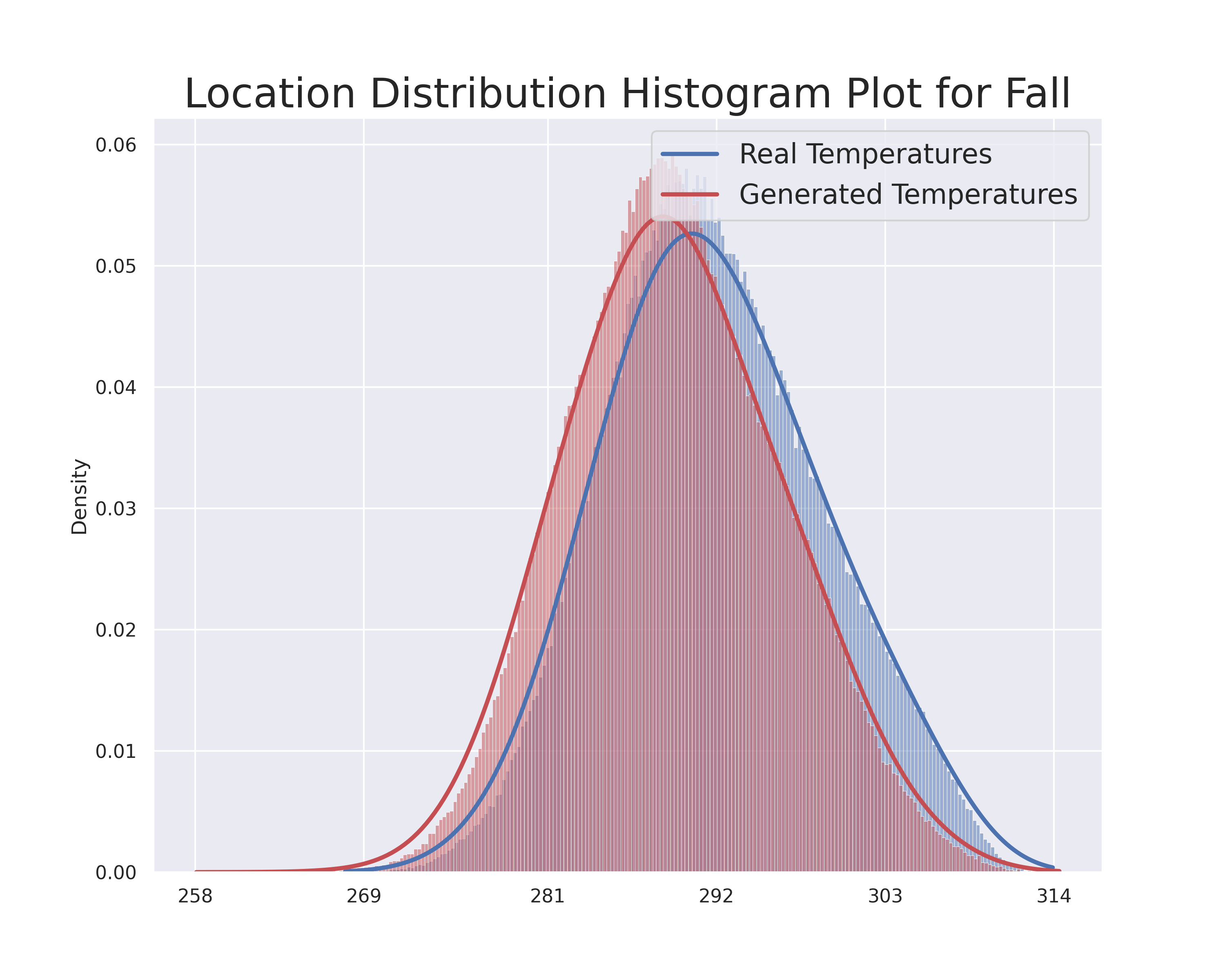}
    \includegraphics[width=0.35\linewidth]{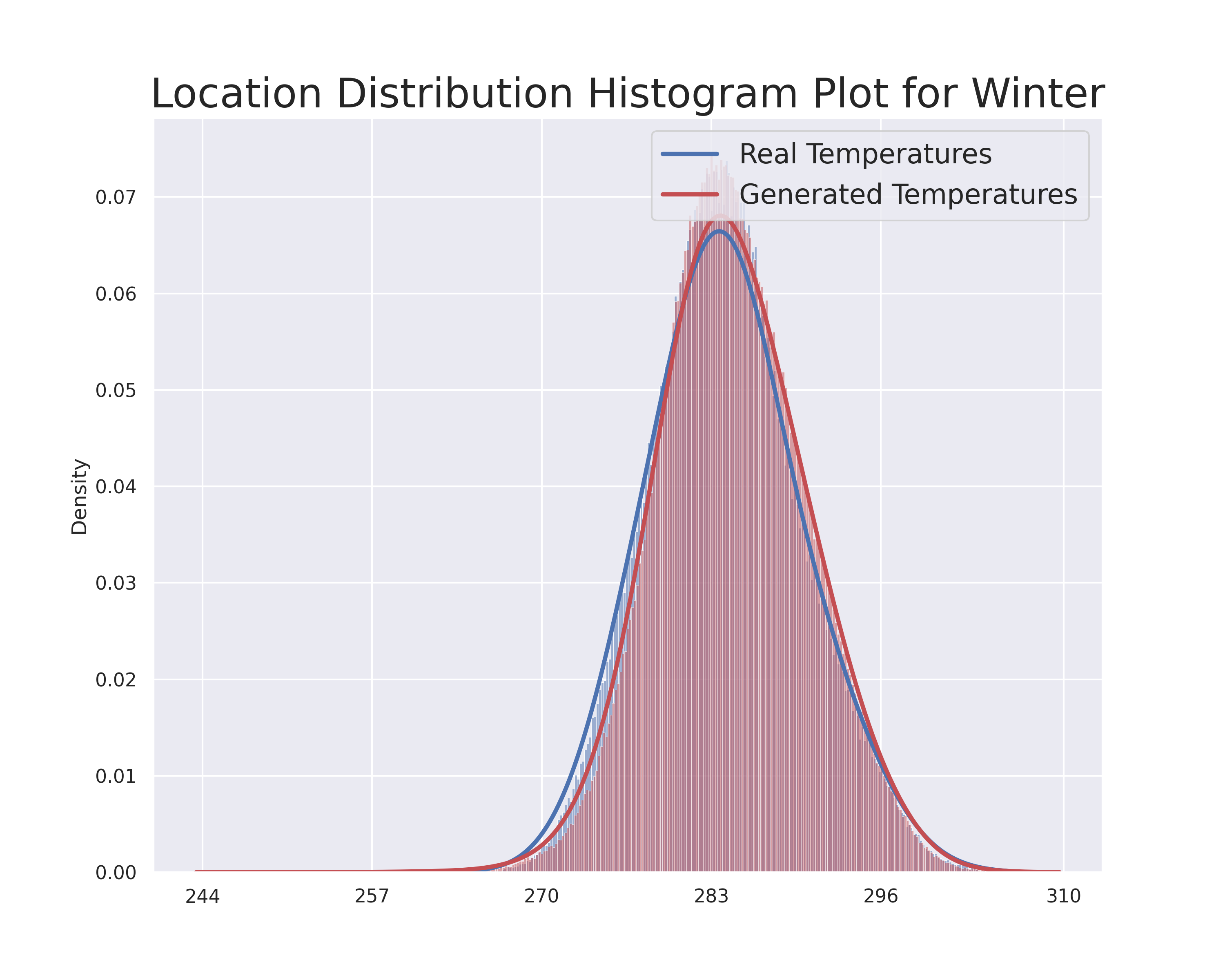}
    \caption{Histograms with kde plots comparing ground-truth (blue) and generated (red) empirical distribution plots for a $1\degree \times 1\degree$ region around the Los Angeles California (2011-2014)}
    \label{fig:base_lbl_dist_plots}
\end{figure}

Figures \ref{fig:PA_real_vs_generated}, \ref{fig:NV_real_vs_generated}, and \ref{fig:base_lbl_dist_plots} show results from TemperatureGAN compared to the ground-truth. The generated outputs are sampled from the model by passing in the input noise, region, month, and period labels as displayed in Figure \ref{fig:model_framework}. The displayed outputs are (random) representative days for the given month, region, and period. The displayed ground-truth data is similarly randomly sampled. The time zones for generated examples are in Universal Time Coordinate (UTC), and local times are included. More generated examples can be accessed via the hyperlink \href{https://drive.google.com/drive/folders/1-wj-6qX0RZXqlr1BHjPGFXcHC5Wtj0Uk?usp=sharing}{here}, and additional distribution plots are in Figures \ref{fig:base_lbl_dist_plots_LA} and \ref{fig:period_sampling} of Appendix \ref{distribution_plots}.


\subsection{Model evaluation}
Because generative models attempt to learn a probability distribution $\hat{p}(x, y)$ (or $\hat{p}(x)$ if labels are not observed) that estimates the true probability distribution $p(x, y)$ (where $x$ represents the data sample or its features and $y$ represents the labels/class) of data, it is important to develop methods/metrics for evaluation that quantify how well a model performs this task. This can be achieved by estimating the likelihood that an example $\hat{x} \sim\hat{p}(x)$ comes from the true distribution $p(x)$.

Using generative ML for climate variables differs from traditional ML tasks such as image, speech, or video synthesis, because models generate physical values for which a unit discrepancy has physical consequence. It is important to establish a simple yet efficient method for quantifying how well or poorly a model performs on any given climate variable, offering standard baselines that future models can be compared to. Metrics should ideally (1) be efficiently computed, (2) consistently track quality, and (3) be relevant and easily adopted by the ML, climate, and energy/power systems communities. We discuss and propose ideas for evaluation below.

If the true distribution $p(x)$ of $x$ is known or $p(x)$ approximately admits a certain functional form (e.g. Gaussian), one can compute the distance between the estimates of the sufficient statistics of the generated $\hat{p}(x)$ and true distribution $p(x)$. A metric that uses this approach is the Fr\'echet Inception Distance (FID) \citep{heusel_ramsauer_unterthiner_nessler_klambauer_hochreiter_2017}, which is commonly used on GANs. FID implicitly assumes that the intermediate feature vectors extracted from images using an Inception V3 model trained on the ImageNet data set, come from a multivariate normal distribution. That is, $X_r \sim \mathcal{N}(\mathbf{\mu_r}, \mathbf{\Sigma_r})$ and $X_g \sim \mathcal{N}(\mathbf{\mu_g}, \mathbf{\Sigma_g})$ with $(\mathbf{\mu_r}, \mathbf{\Sigma_r})$ and $(\mathbf{\mu_g}, \mathbf{\Sigma_g})$ as the mean-covariance pairs for the ground-truth and generated images, respectively. In some other cases, the distribution $p(x)$ may not be continuous, which can make evaluation more challenging. In discontinuous cases, one may take the approach of piece-wise evaluation, if there exists a continuous form of the distribution $p_{q}(x)$ for a certain interval $\mathbf{q_0 \leq q \leq q_1}$. And, if the distribution within these intervals admits a known functional form, then a weighted average of the distances between the sufficient statistics for all intervals can be adopted; this approach is limited to real-valued distributions.

In some other cases, a functional form of $p$ is unknown, thus a non-parametric goodness of fit (e.g. Kolmogorov–Smirnov statistic) test can be adopted, or entropy-based (Kullback Leibler (KL) and Jensen-Shannon (JS)) \citep{kullback1951information, JS} methods can be leveraged. The JS divergence is generally accepted as the symmetric distance measure borne from the KL divergence, thus can be formally considered a metric. For evaluating a generative model when the functional form of $p$ is unknown, one can empirically estimate the JS-divergence by taking the following steps:

\begin{enumerate}
    \item Choose the number of bins $n$, which decides how many quantile intervals the datapoints will be placed into,
    \item Sort the data and place every datapoint/sample into a bin corresponding to its quantile range within the dataset,
    \item Empirically calculate $D_{JS}(\mathbf{P}||\mathbf{Q})$ for each bin and then compute the average JS-divergence.
\end{enumerate}
We discuss and formalize the evaluation of TemperatureGAN for the rest of this section.

\subsubsection{Q-Q Envelopes}\label{QQ_plots}

Q-Q plots are typically leveraged to examine the plausibility that two separate datasets come from the same distribution. They can also help discern the distribution quantiles for which for uncertainty is higher.

\begin{figure}[h!]
    \centering 
    \subfloat[]{\includegraphics[width=0.8\linewidth]{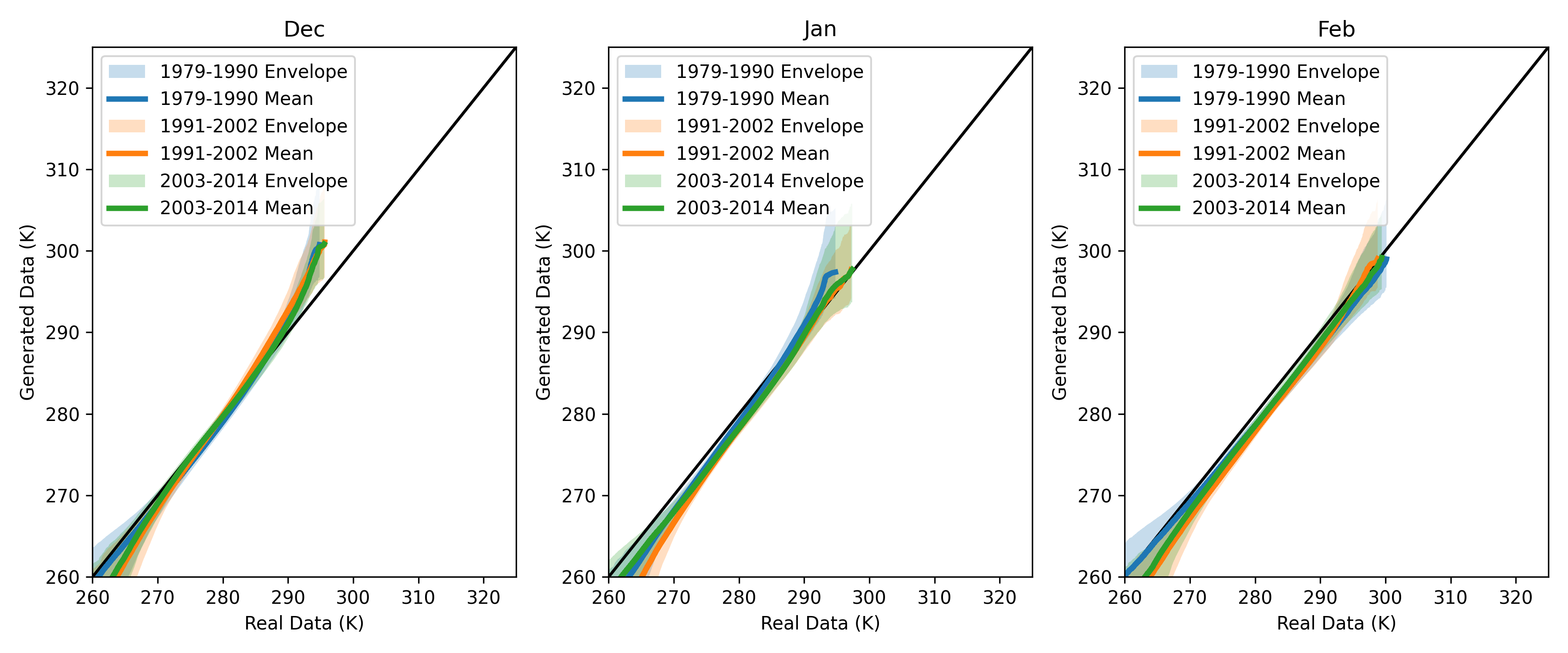}\label{fig:qq_nv_winter}}\
    \subfloat[]{\includegraphics[width=0.8\linewidth]{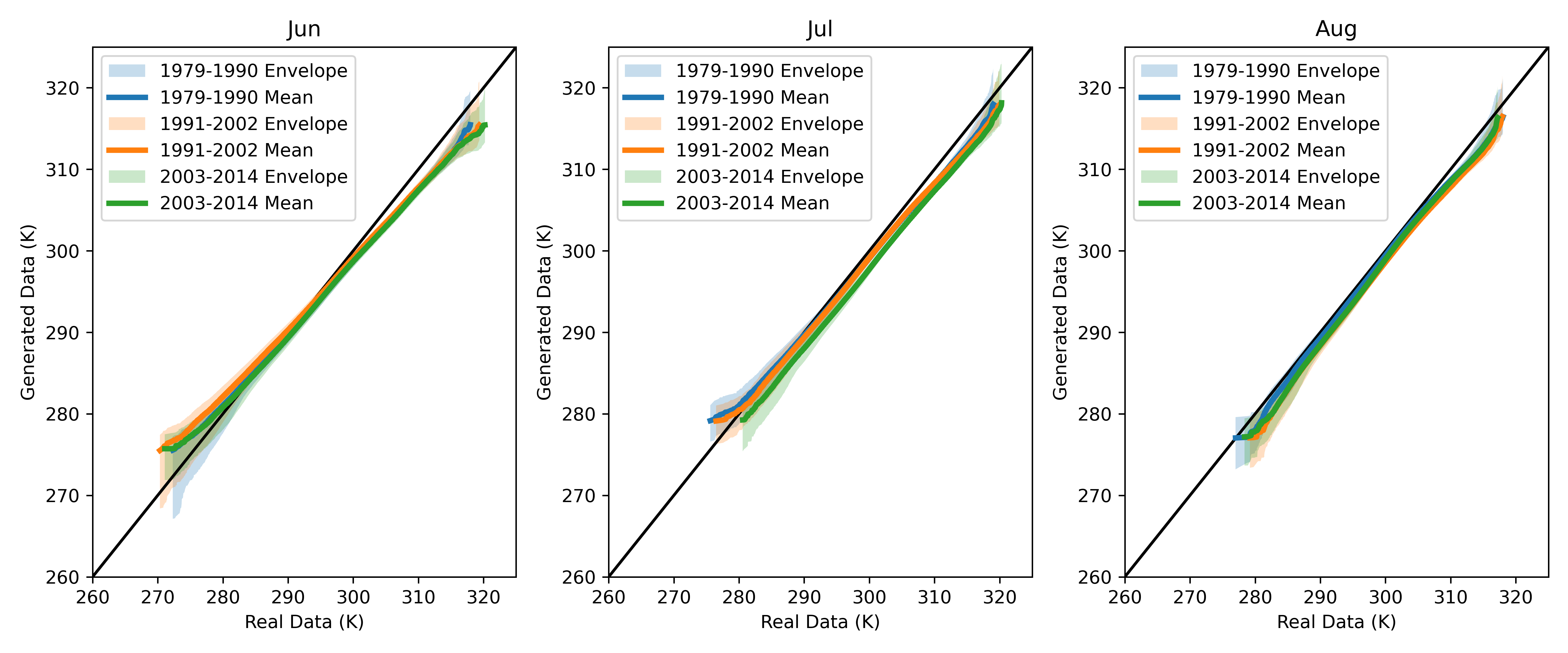}\label{fig:qq_nv_summer}}
    \caption{Q-Q Plot envelopes for winter (\ref{fig:qq_nv_winter}) and summer (\ref{fig:qq_nv_summer}) in Nevada region}
    \label{fig:qq_nv}
\end{figure}

For each of the Q-Q plot envelopes in Figure~\ref{fig:qq_nv} within each period (labelled by the legends), TemperatureGAN is sampled 100 times (to generate 100 realizations), while the ground-truth contains one realization (we can only observe a single realization). The plot elucidates a few things. A key observation across all seasons is that the bulk of the distribution produces a tighter envelope around the constant (black) line, while the envelope at the tails are wider. This is not too surprising, because the observations at the tails are sparse, thus the spread (envelope) around the constant line is observed to be wider. Empirical evidence also suggests that the temperature maps generated are bounded, which is important for representing plausible physical temperature states. The samples generated from TemperatureGAN may be leveraged to expand the potential realizations that are plausible within a given region, which can aid robust planning for energy transition planning agencies. More Q-Q envelopes are displayed in Appendix~\ref{more_QQ_plots}.

\subsubsection{Baseline}
Because we group $1\degree \times 1\degree$ regions as one region $R$, the spatial temperature patterns are not critical for analyzing the overall effects on that region, but the temperature distributions are critical. However, for capturing more granular, local effects within a region, the spatial patterns become critical. For spatial representation evaluation, we propose a baseline model. Because daily temperatures typically exhibit cyclical (diurnal) patterns, we assume that for a given hour of the day, the temperatures within a ($1\degree \times 1\degree$) region overall follow a normal distribution; that is $T(M, R, k, t) \sim \mathcal{N}(\mu, \sigma)$, where $M, R, k, t$ represent, month, region, period, and hour-of-day, respectively. For each hour, we compute the empirical spatial means and variances of the ground-truth data. Thus, we have 24-dimensional mean and standard deviation vectors $\hat{\mu}_{S}, \hat{\sigma}_{S} \in \mathbb{R}^{24}$, and can then generate multiple examples using these statistics.  This yields a fairly simple model that generates temperature maps quickly. This baseline is compared to TemperatureGAN in the following section.

\subsubsection{Spatial Pixel-wise Average Correlation Distance  (SPAC'D)}
We introduce SPAC'D (pronounced "spaced"). To evaluate the spatial representation, we leverage the idea of covariance. Specifically, we calculate the L1-norm distance of the Pearson product-moment correlation coefficient matrices \citep{benesty2009pearson} between the ground-truth and generated data samples. It is worth noting that the choice of the L1-norm is not arbitrary. We elect the L1-norm because it ensures the metric has a fixed range, making it suitable for frames with varying length and width; an intuitive description of SPAC'D is included in Appendix \ref{appendixSPACD}. SPAC'D estimates how well the pixel-wise correlations (or relationships), in the ground-truth are replicated by the generator. It has a range $[0, 2]$, with the quality of spatial representation increasing with decreasing value. Additionally, it is resolution-invariant, meaning the spatial resolution (or size) of the samples being evaluated does not alter its range. We define SPAC'D below with $\tilde{T}\sim\mathbb{P}_g$ and $T\sim\mathbb{P}_r$, where the subscripts $r$ and $g$ represent the generated and ground-truth data, respectively.

\begin{equation}
    SPAC'D = \frac{1}{N}\norm{\rho_{\tilde{T}, \tilde{T}} - \rho_{{T}, {T}}}_{1}
\end{equation}

where $ N = W \times D$ is the total number of pixels per frame with W and D representing the number of pixels in the x and y directions. $\norm{\cdot}_1$, a matrix L1-norm is the maximum sum of absolute values of the column vectors of a matrix. For any matrix $A$, the L1-norm is given by:

\begin{equation}
    \norm{A}_1 = \underset{x\neq0}{\sup}\frac{\norm{A x}_1}{\norm{x}_1} = \underset{j}{\max}\sum_{i=1}^{n} \abs{a_{ij}}
\end{equation}
$\rho$ is the correlation coefficient matrix with each entry given by Equation \eqref{corr_coeff} below.

\begin{equation}
    \rho_{x,y} = \frac{\mathbb{E}[(x - \mu_x)(y - \mu_y)]}{\sigma_x\sigma_y};
    {\sigma_x\sigma_y} > 0,
    \label{corr_coeff}
\end{equation}
where $x$ and $y$ represent the two features for which $\rho_{x, y}$ is calculated.

\begin{figure}[h!]
    \centering
    \includegraphics[width=0.3\linewidth]{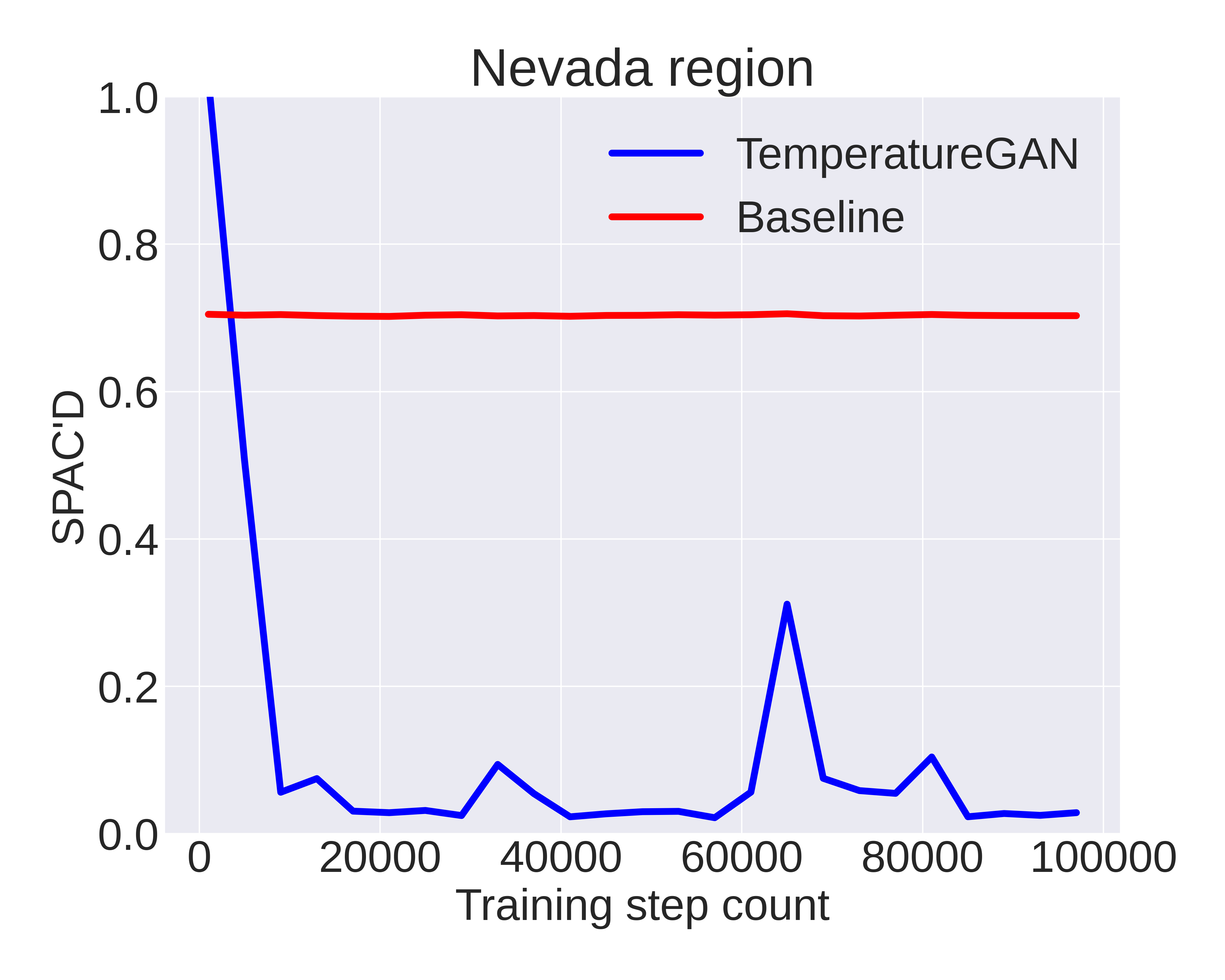}
    \includegraphics[width=0.3\linewidth]{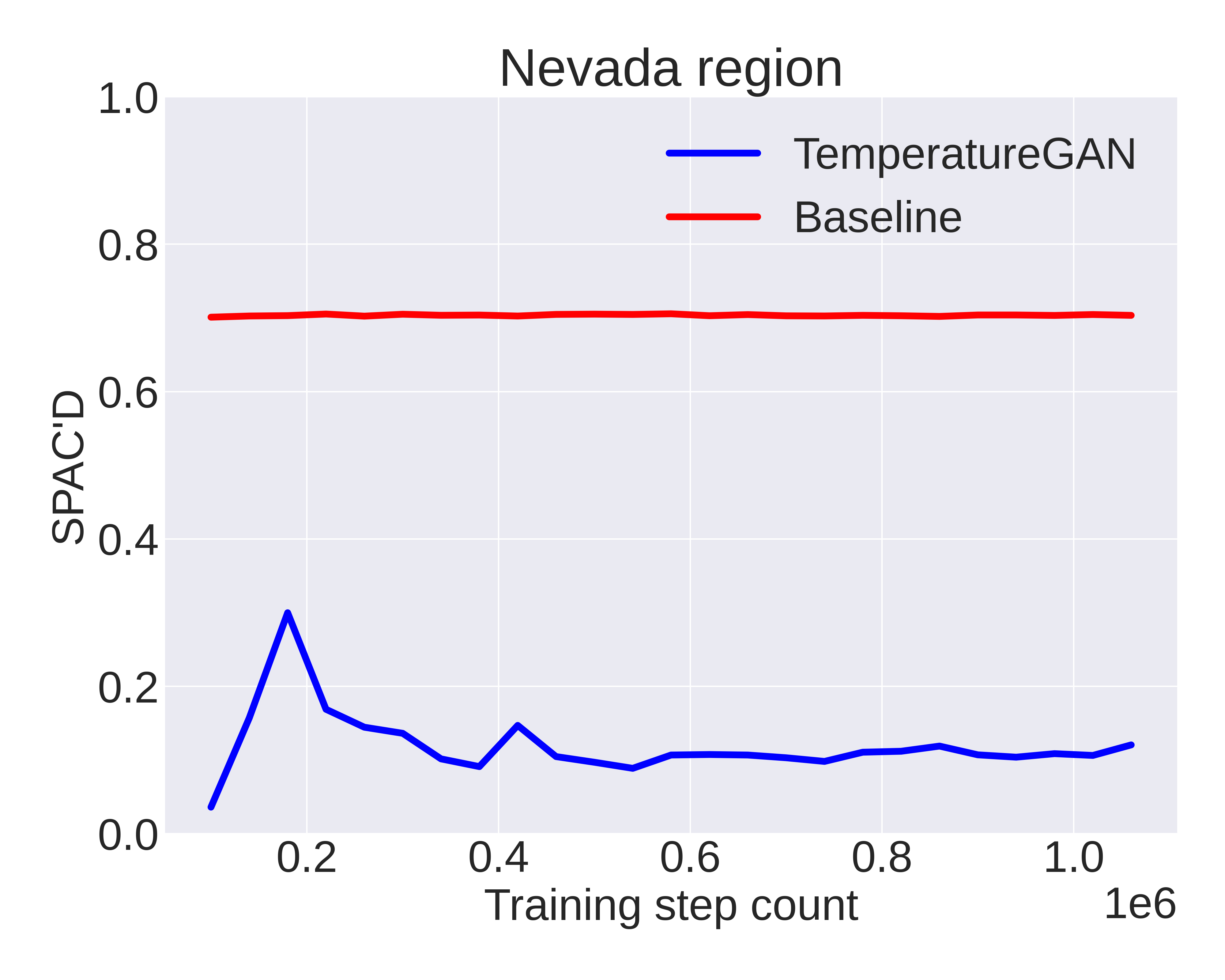}\\
    \includegraphics[width=0.3\linewidth]{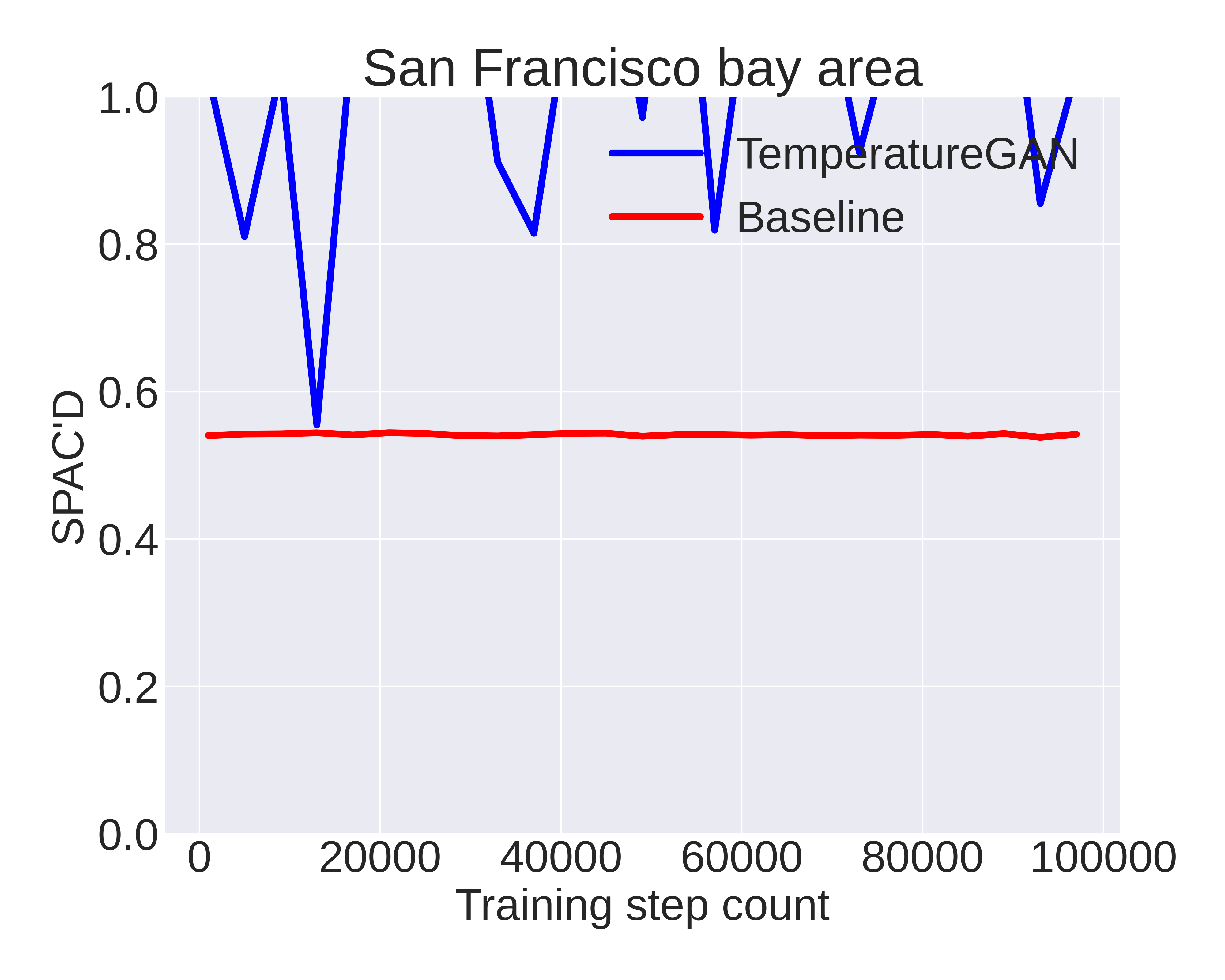}
    \includegraphics[width=0.3\linewidth]{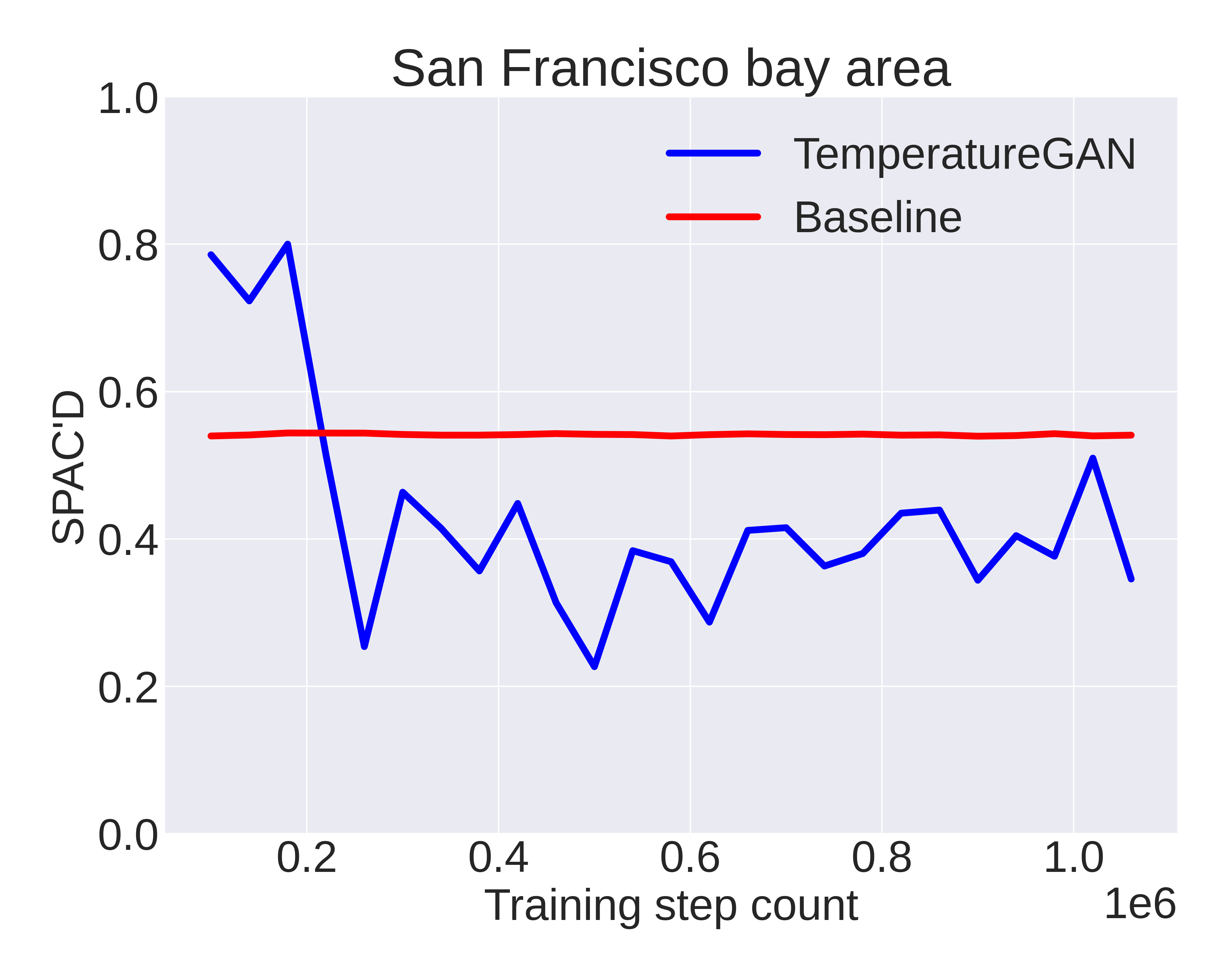}\\
    \includegraphics[width=0.3\linewidth]{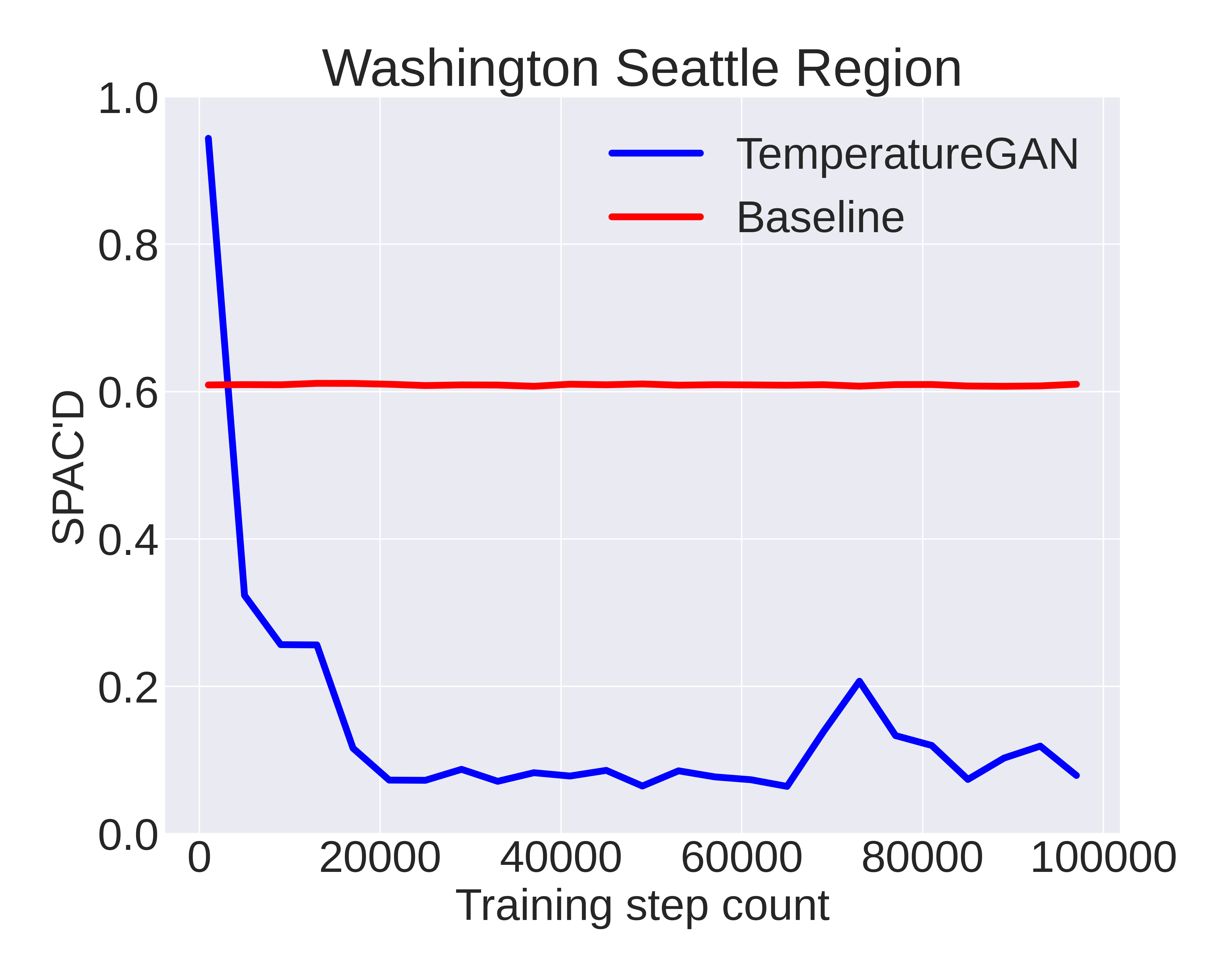}
    \includegraphics[width=0.3\linewidth]{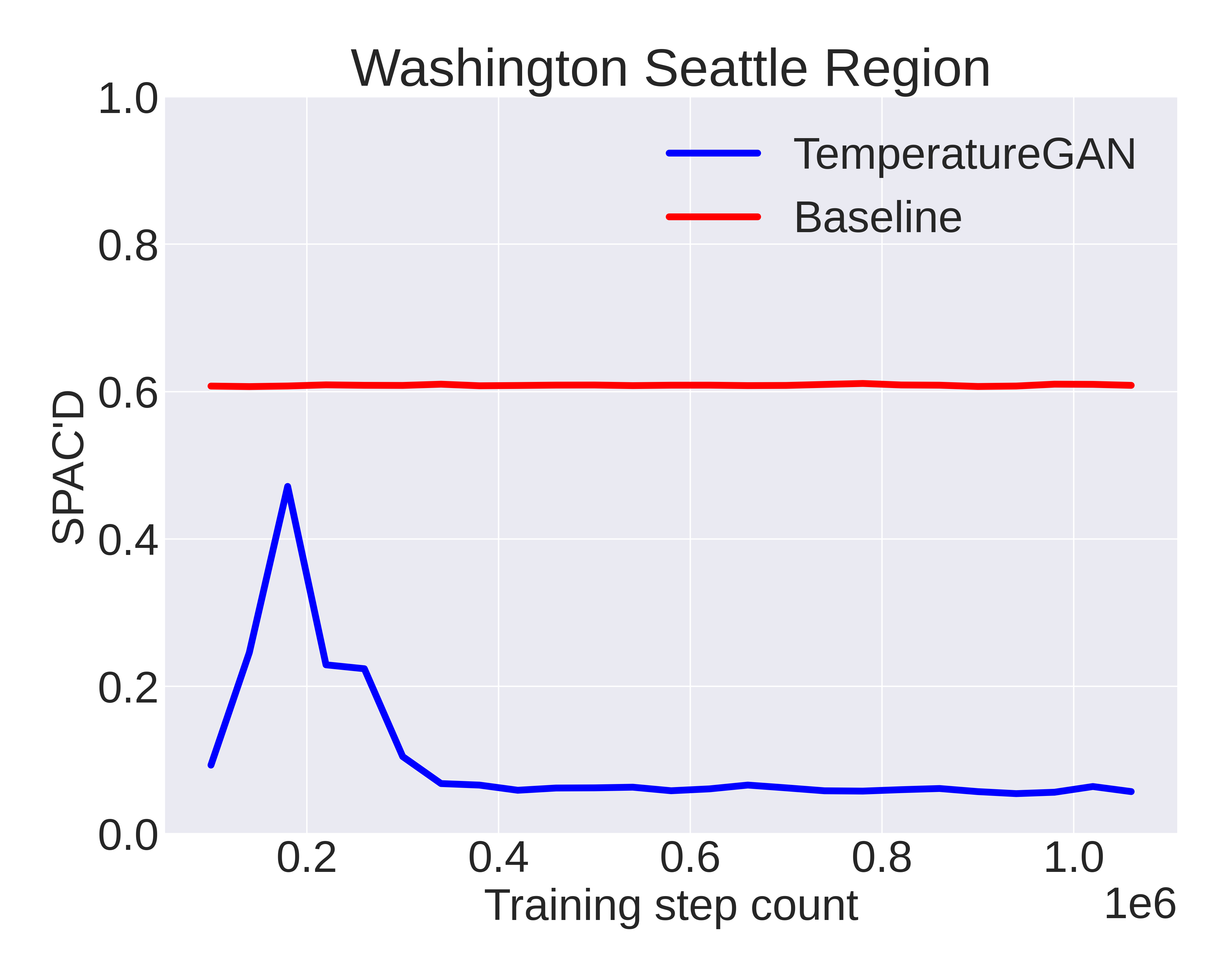}\\
    \includegraphics[width=0.3\linewidth]{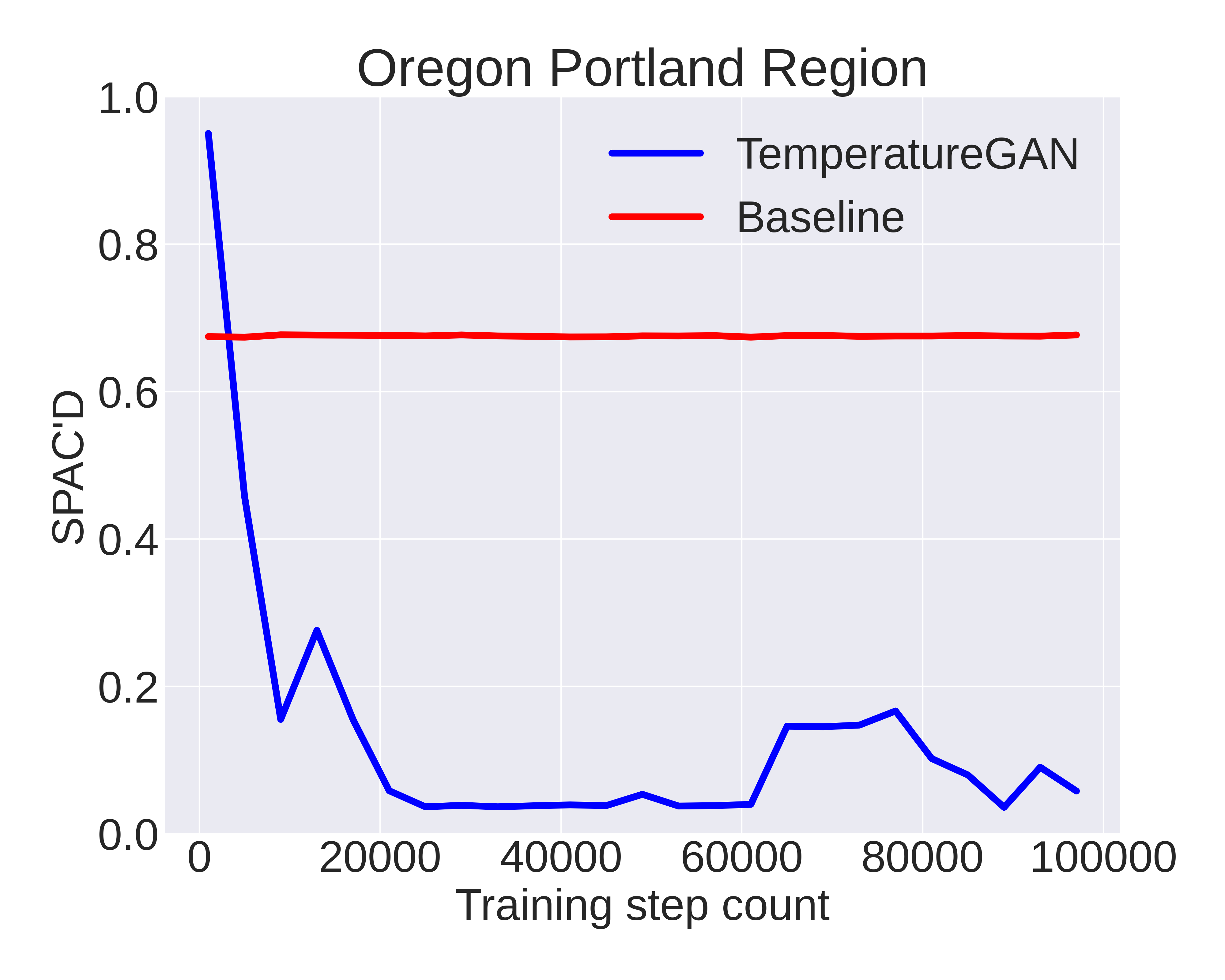}
    \includegraphics[width=0.3\linewidth]{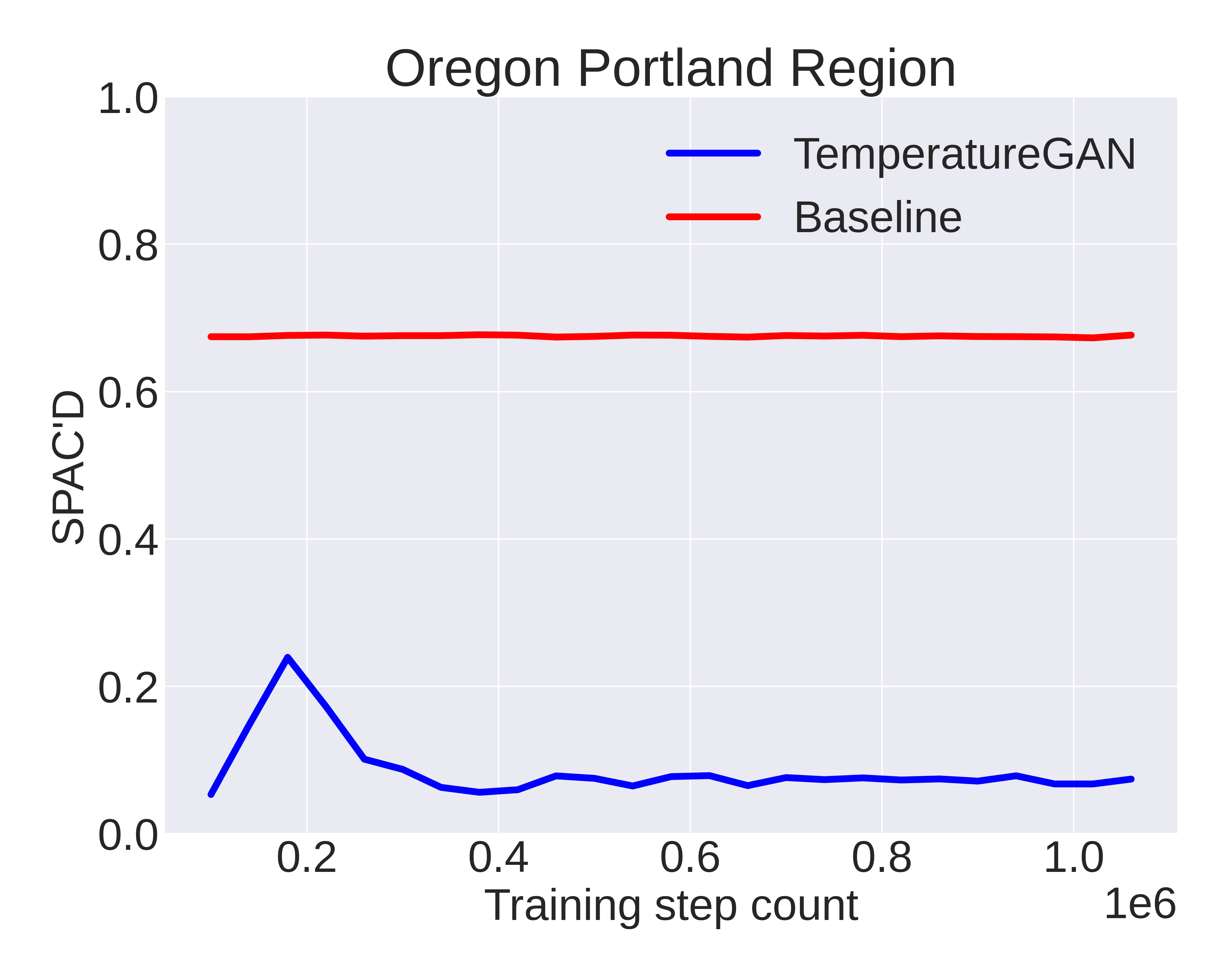}
    \caption{SPAC'D plots for Nevada, San Francisco Bay, Washington, and Oregon regions (top to bottom). The left plots represent initial training steps and the right plots display latter training steps. The red line is the baseline for which the model is compared with at various regions (1979 - 1982). It is evident that the model learns some of the structure of the temperature fields later into training}
    \label{fig:SPACD_plots}
\end{figure}

We call TemperatureGAN `$G$' and the baseline `$B$'. We sample temperature maps $T_G \sim G(z, M, R, k)$ and $T_B \sim B(\hat{\mu}_{S}, \hat{\sigma}_{S})$ (where each sample  $T_{G, B} \in \mathbb{R}^{24 \times 8 \times 8}$), from the GAN and baseline, respectively and report this metric for different regions. Because we are introducing this metric for the first time in this regime, we cannot compare it to other existing models, but it offers a baseline for future comparison. The plots in Figure \ref{fig:SPACD_plots} show the SPAC'D values over training steps. The initial examples produced by the GAN, display inferior SPAC'D values compared to the baseline. Because, during the earlier stages of training, the parameters of the neural network are in proximity to their randomly initialized values. However, later into training (see plots on the right and notice training steps), there is a significant decline (improvement) in the SPAC'D values. This shows that not only are the generated temperature ranges (distribution) accurate, the spatial temperature fields generated have structure.

\subsubsection{Fr\'echet Daily-mean Temperature Distance (FDTD)}
\label{FDTD_appendix}

In addition to evaluating spatial correlations, it is important to evaluate the temperature values being generated. By visual inspection of Figure \ref{fig:PA_real_vs_generated} and \ref{fig:NV_real_vs_generated}, observe that TemperatureGAN generates realistic temperatures for the given regions and months. However, we propose a metric to measure its performance. Daily mean temperatures are typically assumed to follow Gaussian distributions \citep{meehl2000introduction}. Thus, we compute the distance between the sufficient statistics of the ground-truth and generated data as parameterized by a Gaussian. For a certain region $R$, in a given month $M$, and period $k$, the FDTD is calculated by taking the daily average temperatures across every observation. Then, the central bulk of the data is estimated by a normal distribution. The bulk is chosen as the 10th to 90th percentile daily mean temperatures from the ground-truth data, excluding the tails, as they usually admit a different functional form, usually characterized by Generalized Extreme Value (GEV) or Generalized Pareto (GPD) distributions. Thus, for a set of daily mean temperatures $\mathbf{\bar{T}} = \{\bar{T_1}, \bar{T}_{2}, \bar{T}_{3}, ..., \}$, we select a subset $\mathbf{\bar{T}_{bulk}} \subset \mathbf{\bar{T}}$, fit a normal distribution to $\mathbf{\bar{T}_{bulk}}$, and compute the distance of the sufficient statistics of the generated examples from the ground-truth data as expressed in Equation \eqref{equ:FDTD}.

\begin{equation}\label{equ:FDTD}
    FDTD = \sqrt{\norm{\mu_r - \mu_g}_2^2 + \norm{\sigma_r - \sigma_g}_2^2}
\end{equation}
The subscripts $r$ and $g$ represent ground-truth and generated examples, respectively. $\mu_r$ and $\mu_g$ are ground-truth and generated data sample means respectively, and $\sigma_r$ and $\sigma_g$ are the ground-truth and generated data sample standard deviations, respectively. The results are reported in Tables \ref{FDTD_PA_0}, \ref{FDTD_WA_0}, and \ref{FDTD_NV_0}.

\begin{table}[h!]
  \caption{FDTD (K) for Period 0 (1979 - 1982), San Francisco Bay area. Avg. distance 0.4 K/\degree C}
\label{FDTD_PA_0}
  \centering
  \begin{tabular}{llllll}
    \toprule        
        Month & Real Mean & Generated Mean & Real STDEV & Generated STDEV & FDTD\\
        \hline
        January & 283.3407 & 282.5232 & 2.0201 & 2.0373 & 0.8177\\
        February & 284.0418 & 283.7031 & 1.9294 & 1.9711 & 0.3412\\
        March & 283.6841 & 283.8510 & 1.7114 & 1.8608 & 0.2240\\
        April & 284.6826 & 285.0927 & 2.1637 & 2.2705 & 0.4239\\
        May & 286.0623 & 286.5555 & 2.5437 & 2.5836 & 0.4949\\
        June & 288.3015 & 288.3525 & 3.2327 & 3.1262 & 0.1181\\
        July & 289.9192 & 290.3221 & 3.5523 & 3.5033 & 0.4059\\
        August & 289.8751 & 290.3902 & 3.3434 & 3.2968 & 0.5172\\
        September & 290.2416 & 290.4065 & 3.1372 & 3.1955 & 0.1749\\
        October & 288.4430 & 288.2218 & 2.5153 & 2.6922 & 0.2832\\
        November & 285.9702 & 285.4471 & 1.8756 & 2.0207 & 0.5428\\
        December & 284.4566 & 283.9878 & 1.7677 & 1.8846 & 0.4832\\
    \bottomrule
  \end{tabular}
\end{table}

\begin{table}
 \caption{FDTD (K) for Period 8 (2011 - 2014), San Francisco Bay area. Avg. distance 0.6459 K/\degree C (Note: model was not trained on data from this period).}
    \label{FDTD_PA_8}
\centering

\begin{tabular}{l l l l l l}
\toprule
Month & Real Mean & Generated Mean & Real STDEV & Generated STDEV & FDTD\\
\hline
January & 283.4979& 282.7405& 2.6653& 2.3138& 0.8350
\\
February & 283.1975& 283.1278& 2.6399& 2.3874& 0.2620
\\
March & 284.1885& 283.8586& 2.6148& 2.5037& 0.3481
\\
April & 285.4229& 285.2516& 3.1637& 2.9431& 0.2793
\\
May & 286.9157& 286.8570& 3.6637& 3.5449& 0.1325
\\
June & 288.8024& 288.6553& 4.2302& 3.6864& 0.5634
\\
July & 290.2967& 290.0674& 4.4900& 4.1238& 0.4321
\\
August & 290.3452& 290.8286& 4.2165& 3.8490& 0.6073
\\
September & 290.5490& 289.7114& 4.1093& 3.3952& 1.1007
\\
October & 289.1697& 288.3250& 3.5333& 2.7549& 1.1487
\\
November & 286.2279& 284.9245& 2.6540& 2.1943& 1.3821
\\
December & 283.8543& 284.0557& 2.8291& 2.2009& 0.6597\\
\bottomrule

\end{tabular}

\end{table}


\begin{table}[h!]
  \caption{FDTD (K) for Period 0 (1979 - 1982), Portland, Oregon Area. Avg. distance 0.4750 K/\degree C}
\label{FDTD_WA_0}
  \centering
  \begin{tabular}{llllll}
    \toprule        
        Month & Real Mean & Generated Mean & Real STDEV & Generated STDEV & FDTD\\
        \hline
        January & 276.6373 & 277.1791 & 2.7718 & 2.8205 & 0.5439\\
        February & 278.9277 & 277.8223 & 2.6563 & 2.4601 & 1.1227\\
        March & 280.4386 & 279.9317 & 2.5140 & 2.5122 & 0.5068\\
        April & 281.9535 & 281.9129 & 3.0382 & 2.9373 & 0.1087\\
        May & 285.1138 & 285.0656 & 3.1491 & 3.2220 & 0.0874\\
        June & 287.5332 & 287.7721 & 3.2011 & 3.4654 & 0.3562\\
        July & 290.5277 & 290.5180 & 3.3878 & 3.6399 & 0.2523\\
        August & 290.5566 & 290.5930 & 3.3165 & 3.5560 & 0.2422\\
        September & 288.9453 & 288.6372 & 3.3243 & 3.5749 & 0.3972\\
        October & 284.8474 & 284.9464 & 3.0092 & 3.2205 & 0.2333\\
        November & 280.1408 & 279.6556 & 2.5067 & 2.5025 & 0.4852\\
        December & 278.9350 & 277.5671 & 2.3930 & 2.3874 & 1.3680\\
    \bottomrule
  \end{tabular}
\end{table}

\begin{table}
\caption{{FDTD (K) for Period 8 (2011 - 2014), Portland, Oregon Area. Avg. distance 0.3655 K/\degree C (Note: model was not trained on data from this period)}}
\centering

\begin{tabular}{l l l l l l}
\toprule
Month & Real Mean & Generated Mean & Real STDEV & Generated STDEV & FDTD\\
\hline
January & 277.7386 & 277.4297 & 2.4842 & 2.4784 & 0.3089 \\
February & 277.6818 & 278.2201 & 2.3703 & 2.3218 & 0.5405 \\
March & 279.6807 & 279.9788 & 2.4901 & 2.4226 & 0.3056 \\
April & 281.8780 & 282.7694 & 2.9036 & 2.8649 & 0.8922 \\
May & 284.9553 & 285.6162 & 3.3554 & 3.3001 & 0.6633 \\
June & 287.5976 & 288.0747 & 3.1221 & 3.0724 & 0.4796 \\
July & 291.1722 & 291.0664 & 3.4644 & 3.4512 & 0.1065 \\
August & 291.9764 & 291.6701 & 3.4593 & 3.5061 & 0.3099 \\
September & 289.8309 & 289.6427 & 3.7360 & 3.5898 & 0.2383 \\
October & 284.7873 & 284.9691 & 2.8783 & 2.9510 & 0.1958 \\
November & 280.0583 & 280.2240 & 2.7787 & 2.7031 & 0.1821 \\
December & 277.1813 & 277.3305 & 2.5667 & 2.6334 & 0.1634 \\
\bottomrule

\end{tabular}

\end{table}

\subsubsection{Temporal Gradient Distribution Distance (TGDD)}
\begin{figure}[h!]
    \centering
    \includegraphics[width=4cm]{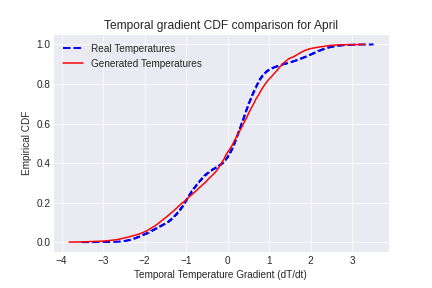}
    \includegraphics[width=4cm]{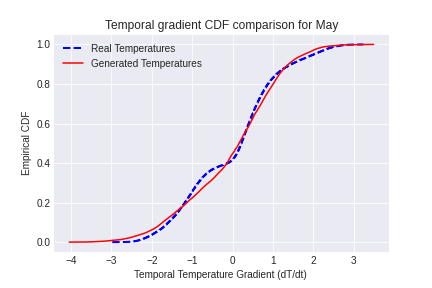}
    \includegraphics[width=4cm]{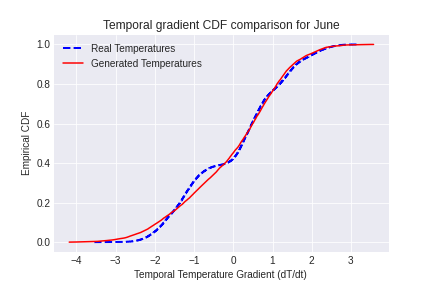}
    \includegraphics[width=4cm]{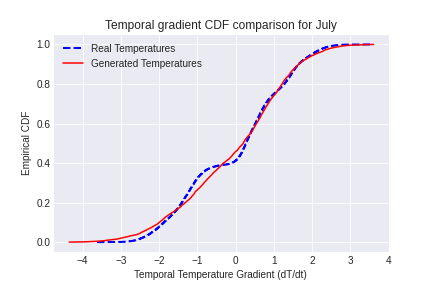}
    \includegraphics[width=4cm]{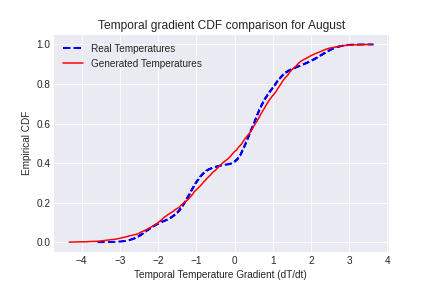}
    \includegraphics[width=4cm]{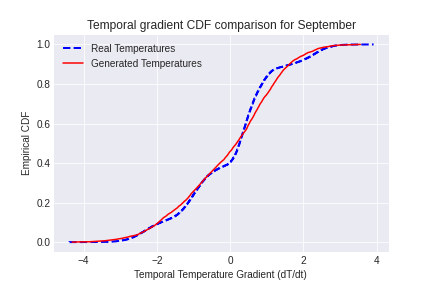}
    \caption{Temporal gradients distribution plots, Los Angeles (1979-1982)}
    \label{fig:TGDD}
\end{figure}

For some applications, generating realistic hourly temperature maps can be useful; for example, power system reliability or capacity sufficiency studies may benefit from this \citep{PANTELI2015259, perera2020quantifying}. Because we generate hourly spatial maps for any given day, we aim to estimate the integrity of the generated diurnal cycle. Visual inspection may be used to validate general cyclical patterns. However, we propose an approach to numerically estimating the model's performance. We obtain the distribution of temperature gradients by empirically estimating the cumulative density functions (CDFs) of the temperature hourly gradients $\mathbf{\frac{\partial T}{\partial t}}$ and $\mathbf{\frac{\partial \tilde{T}}{\partial t}}$, where  $\mathbf{T} \sim \mathbb{P}_{r}$ and $\mathbf{\tilde{{T}}} \sim \mathbb{P}_{g}$ and $\mathbb{P}_{r}$ and $\mathbb{P}_{g}$ represent the ground-truth and generated distributions (see Figure \ref{fig:TGDD}), respectively. To do this, we split the data samples into $n = 10$ bins, estimate $D_{JS}(p||q)$ for each bin,

\begin{equation}
    D_{JS}(p||q) = \frac{1}{2}D_{KL}\left(p||\frac{p + q}{2}\right) + \frac{1}{2}D_{KL}\left(q||\frac{p + q}{2}\right)
    \label{eqn:JS}
\end{equation}

after which we compute the average across all bins to obtain the metric we call TGDD,

\begin{equation}
    TGDD = \frac{1}{N}\sum_{i=1}^{N}D_{JS}(p_i||q_i),
    \label{eqn:TGDD_eqn}
\end{equation}

with a range $\left(0, \ln{2}\right)$, and improving with decreasing value. We report the monthly TGDD in Table \ref{tgdd-table}. Diurnal cycle patterns are displayed in Figure \ref{fig:Temporal_penalty} in Appendix \ref{temporal_integrity} for visual inspection.

\begin{table}[h!]
  \caption{TGDD values. $k_0$ (1979 - 1982)}
  \label{tgdd-table}
  \centering
  \begin{tabular}{lll}
    \hline        
        Month & San Francisco & Nevada\\
        \hline
        January & 0.0935 & 0.0257\\
        February & 0.0783 & 0.0123\\
        March & 0.0797 & 0.0096\\
        April & 0.0532 & 0.0133\\
        May & 0.0195 & 0.0314\\
        June & 0.0283 & 0.0574\\
        July & 0.0237 & 0.0267\\
        August & 0.0227 & 0.0193\\
        September & 0.0270 & 0.0169\\
        October & 0.0311 & 0.0184\\
        November & 0.0862 & 0.0232\\
        December & 0.0941 & 0.0374\\
    \hline
  \end{tabular}
\end{table}

\subsubsection{Comparison with existing Stochastic Weather Generator}

\begin{figure}[h!]
    \centering
    \includegraphics[width=5.5cm]{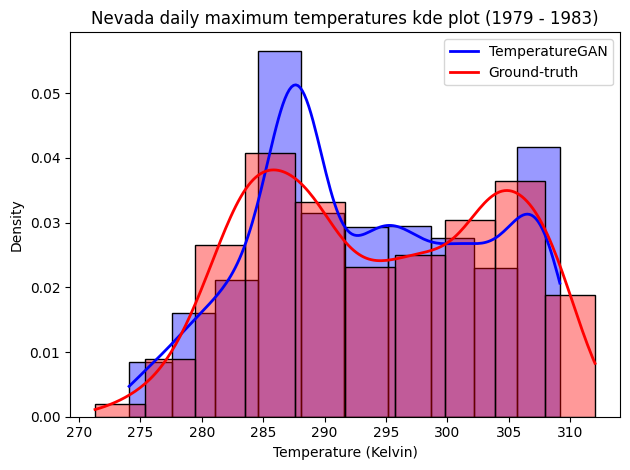}
    \includegraphics[width=5.5cm]{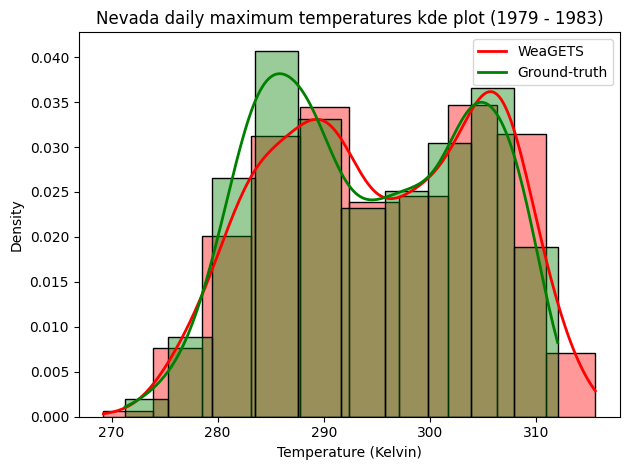}
    \caption{Histograms with kernel density estimate plots for maximum daily temperatures. Nevada (1979--1982)}
    \label{fig:WeaGETS}
\end{figure}


We conclude evaluations by comparing TemperatureGAN to \textit{WeaGETS}, an existing single-site stochastic weather generator (SWG) \citep{chen2010daily}, similar to \textit{WGEN}. We are limited in the breadth of comparsion as \textit{WeaGETS} only produces minimum and maximum daily temperatures for a given site. Because the ground-truth is 3-dimensional, we take the spatial average temperature of the $1\times1$ {region} and obtain the maximum and minimum temperatures for each day. \textit{WeaGETS} performs better than TemperatureGAN in this comparison. We posit that this is largely because the models were given fundamentally different tasks.
TemperatureGAN was developed to generate high-dimensional (3D) samples while \textit{WeaGETS} generates minimum and maximum temperatures for only one region only and is trained on one-dimensional data. \textit{WeaGETS} cannot learn temporal and spatial structure of disparate regions, which is important for more comprehensive energy systems assessments.

\begin{figure}[h!]
    \centering
    \includegraphics[width=6cm]{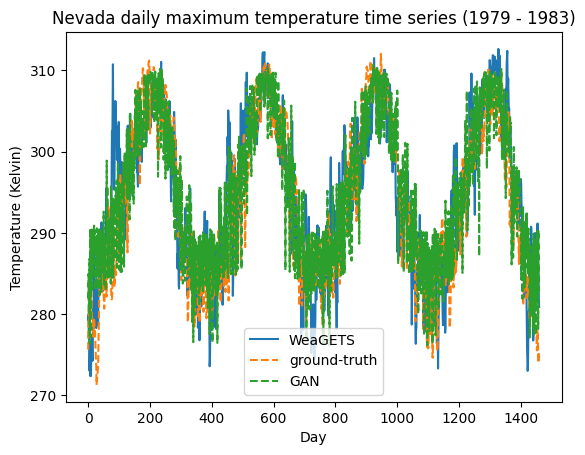}
    \includegraphics[width=6cm]{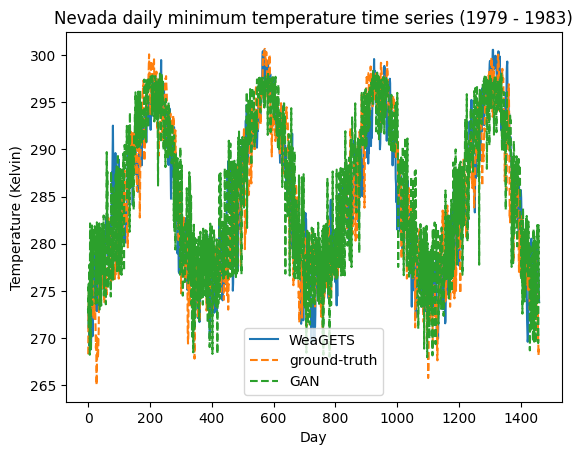}
    \caption{4-year timeseries plots of daily maximum temperatures for the ground truth, GAN, and WeaGETS. Nevada (1979--1982)}
    \label{fig:WeaGETS_timeseries}
\end{figure}

\section{Conclusion}\label{conclude}
In this paper, we introduced TemperatureGAN, a data-driven generative model that efficiently produces spatial temperature maps of a given {region}, month, and period at an hourly temporal resolution. We proposed metrics to evaluate this model and showed that it can learn and reproduce historically observed regional spatiotemporal temperature dynamics.

We discussed four metrics for evaluating the model performance, each serving an important purpose to evaluating the spatiotemporal integrity of the generated samples. Introduced for the first time is \textit{SPAC'D}, a bounded metric that measures the veracity of generated spatial fields. \textit{FDTD} is used to evaluate how well the model captures the distribution of daily average temperatures. \textit{TGDD} is used to scrutinize temporal integrity. Collectively, these metrics show that the GAN reasonably captures conditional temperature distributions. As discussed in the text, these metrics can be adopted for evaluating other models/approaches in this regime.

While this work leveraged only historical temperature measurements, we recognize that other input data streams may be useful. This approach is independent of input from global climate models (GCM) or other exogenous anthropogenic inputs. However, retrospective runs of GCMs could provide the GAN with a useful signal to better capture trends over time. We also recognize that regenerating accurate spatial gradients for multiple regions by virtue of its relative position is rather arbitrary and can be improved by including more physically meaningful priors for various regions, for example, topological maps. Topology maps at a finer spatial resolution than the base temperature data could be ingested by the GAN to improve both the spatial representation integrity and potentially, resolutions. Future work will leverage this.

Although TemperatureGAN can be used to generate regional temperatures, we caution against leveraging it for extreme value analysis for a few reasons. Firstly, the model has not been has not been rigorously evaluated within the regime of producing rare, extreme temperatures. Secondly, the 4-year period $k$ used in training TemperatureGAN may not produce a sufficient number of temperature observations to accurately capture the distribution of extreme temperature values. Because there are fewer samples, it may lead to high variance in the parameters for the extreme value models. Further work can be done to improve and validate the TemperatureGAN's performance at the tails of temperature distributions, however the current model demonstrates promising characteristics that can be built on for future work. Additionally, TemperatureGAN in its current state should not be applied to future climate as it has not been designed to handle nonstationarity; we defer this to future developments of TemperatureGAN.

There are many downstream applications for models that can produce realistic insights into regional temperature events, and its effects on communities, power distribution circuits, etc, especially in a \textit{quick} and \textit{scalable} fashion and TemperatureGAN provides a method to do this. 

\newpage
\begin{Backmatter}

\paragraph{Acknowledgments}
The authors would like to acknowledge Rob Buechler for his helpful feedback and suggestions throughout the project, and his assistance with initial data retrieval/processing. The authors would also like to thank Jehangir Amjad for helpful discussions and suggestions at the beginning of the project. The authors also thank Emily Gordon and Amina Ly for helpful discussions at the latter stages of revising this paper.

\bibliographystyle{apalike}
\bibliography{ref.bib}

\paragraph{Funding Statement}
Emmanuel Balogun was supported by the Stanford Data Science Scholarship and Chevron Energy Fellowship.

\paragraph{Competing Interests}
None

\paragraph{Data Availability Statement}
The original NLDAS-2 forcing data used in this paper is publically available and can be assessed \href{https://disc.gsfc.nasa.gov/datasets/NLDAS_FORA0125_H_002/summary?keywords=NLDAS}{here}. The modified data for training the model is publicly accessible on data mendeley. Model training repository and trained models have been made publicly available on \href{https://github.com/ebalogun01/TemperatureGAN}{github}.

\paragraph{Ethical Standards}
The research meets all ethical guidelines, including adherence to the legal requirements of the study country.

\paragraph{Author Contributions}
\textbf{E.B}: Conceptualization, Methodology, Software, Investigation, Validation, Visualization, Formal analysis, Writing --- original draft.  \textbf{R.R}: Supervising, funding. \textbf{A.M}: Supervising, funding, writing. All authors approved the final submitted draft.

\paragraph{Supplementary Material}
All supplementary material is contained in the appendix or linked in the main text. Code and trained models are accessible on \href{https://github.com/ebalogun01/TemperatureGAN}{github}. Training data is accessible \href{https://data.mendeley.com/datasets/9k892pzkfx/1}{here}.

\begin{appendix}\appheader
\section{Model architecture}\label{appendixA}
Tables \ref{G-table}, \ref{Dt-table}, and \ref{Ds-table} describe the neural network architectures implemented in this paper.

\begin{table}[h!]
  \caption{Generator $G$ Architecture}
  \label{G-table}
  \centering
  \begin{tabular}{lll}
    \toprule
        Layer Type & Parameters & Output Shape \\ 
        \hline
        Input (noise) & $z\in\mathbb{R}^{100}$, concat with transformed labels & (N, 200)  \\
        FC Linear & Neurons = 400 & (N, 400) \\
        FC Linear & Neurons = 800 & (N, 800) \\
        FC Linear & Neurons = 100 & (N, 100) \\
        Unsqueeze & - & (N, 1, 100) \\
        ConvTranspose1D & Filters = 10, kernel size= 3, BatchNorm, ReLU & (N, 10, 102)  \\
        ConvTranspose1D & Filters = 10, kernel size= 3, BatchNorm, ReLU & (N, 10, 104)\\
        ConvTranspose1D & Filters = 64, kernel size= 5, BatchNorm, ReLU & (N, 64, 108)  \\
        ConvTranspose1D & Filters = 112, kernel size= 5 BatchNorm, ReLU & (N, 112, 112) \\
        Conv1D & Filters = 28, kernel = 1, stride= 4 & (N, 28, 28)\\
        Unsqueeze & - & (N, 1, 28, 28)\\
        Conv2D & Filters = 2, kernel = (5, 5), BatchNorm, ReLU & (N, 2, 24, 24)\\
        Conv2D & Filters = 4, kernel = (5, 5), BatchNorm, Tanh & (N, 4, 20, 20)\\
        Conv2D & Filters = 16 kernel = (5, 5), BatchNorm, Tanh & (N, 16, 16, 16)\\
        Conv2D & Filters = 24,  kernel = (5, 5),  BatchNorm,Tanh & (N, 24, 12, 12)\\
        Output Conv2D & Filters = 24 kernel = (5, 5) & (N, 24, 8, 8) \\
    \bottomrule
  \end{tabular}
\end{table}

\begin{table}[h!]
  \caption{Discriminator $D_{s}$ Architecture}
  \label{Ds-table}
  \centering
  \begin{tabular}{lll}
    \toprule
        Layer Type & Parameters & Output Shape \\ 
        \hline
        Input (3D-Image) & $img\in\mathbb{R}^{24X8X8}$ & (N, 24, 8, 8)  \\ 
        Conv2D & Filters=24, kernel size=3, padding, LeakyReLu(0.2) & (N, 24, 8, 8) \\
        Conv2D & Filters=24, kernel size=3, padding, LeakyReLu(0.2) & (N, 24, 8, 8)  \\
        Conv2D & Filters=24, kernel size=3, LeakyReLu(0.2) & (N, 24, 6, 6)  \\
        Conv2D & Filters=24, kernel size=3, LeakyReLu(0.2) & (N, 24, 4, 4) \\
        Conv2D & Filters=24, kernel size=3, LeakyReLu(0.2) & (N, 24, 2, 2) \\
        Conv2D & Filters=24, kernel size=2, LeakyReLu(0.2) & (N, 24, 1, 1) \\
        Flatten & - & (N, 24) \\
        FC Linear & Neurons=100, LeakyReLu(0.2) & (N, 100) \\
        Concat embedding & - & (N, 200) \\
        FC Linear & Neurons=150, LeakyReLu(0.2) & (N, 150) \\
        FC Linear & Neurons=50, LeakyReLu(0.2) & (N, 50) \\
        Output FC Linear & Neurons=1 & (N, 1)\\
    \bottomrule
  \end{tabular}
\end{table}

\begin{table}[h!]
  \caption{Discriminator $D_{t}$ Architecture}
  \label{Dt-table}
  \centering
  \begin{tabular}{lll}
    \toprule
        Layer Type & Parameters & Output Shape \\ 
        \hline
        Input (3D-Image) & $img\in\mathbb{R}^{24X8X8}$ & (N, 24, 8, 8)  \\ 
        Gradient Computation & - & (N, 23, 8, 8) \\
        Conv2D & Filters=16, kernel size=(3, 3), LeakyReLu(0.2) & (N, 16, 6, 6)  \\
        Conv2D & Filters=4, kernel size=(3, 3), LeakyReLu(0.2) & (N, 4, 4, 4)  \\
        Flatten & - & (N, 64) \\
        Concat embedding & - & (N, 164)\\
        FC Linear & Neurons=164, LeakyReLu(0.2) & (N, 164) \\
        FC Linear & Neurons=200, LeakyReLu(0.2) & (N, 200) \\
        FC Linear & Neurons=100, LeakyReLu(0.2) & (N, 100) \\
        Output FC Linear & Neurons=1 & (N, 1)\\
    \bottomrule
  \end{tabular}
\end{table}
\FloatBarrier

In all models, the labels are mapped from $\mathbb{R}^{15}$ to $\mathbb{R}^{100}$ learned embedding during training.

\subsection{GAN Conditioning}
The month is represented as a discrete one-hot vector, which suggests that we cannot continuously vary its representation to interpolate between months. It is also an intuitive decision as the number of months in a year is not changing, so it is convenient to think of each month as one class from a multi-class size of 12.
It can be desirable to generate temperatures for a given month because downstream impacts may also depend on the month/time of the year. One example is on grid reliability or resilience studies. Energy usage patterns are typically seasonal and can change depending on the month, thus, the impacts of temperatures can be much more severe during months when demand is also high. Additionally, the type of demand can be critical. For example, HVAC vs EV charging, or a mix of both. Being able to understand times of the year that pose higher risk can be particularly desirable with the proliferation of long-duration or seasonal storage and DERs. Energy transition planners or utilities can estimate potential risks for each month and strategically plan resources accordingly.
We include month as a conditional variable because months are universal and the number of months each year is fixed; intuitively, it makes the model more generalizable because seasons are region-dependent. Additionally, we carried out experiments using seasons (both continuous and discrete) as a conditional variable and found that the model did not perform as well as using one-hot encoded months variable. We also note that once we are able to model each month accurately, it is much easier to aggregate the different months (see Figure \ref{fig:ECDF_1}) into desired seasons (barring some slight inaccuracies due to smoothness of seasonal transitions). The converse is much harder, if not impossible. That is, once we aggregate certain months during training, we lose information of month-to-month variations that may exist and it will be very challenging to recover.

The region is represented as a dual-axis variable (X, Y), representing its relative position to the the specified origin, which is selected to be the southwest (SW) corner of the dataset. The GAN is conditioned on these relative positions during training. A depiction of this is shown in Figure \ref{fig:condition}.

\begin{figure}[h!]
    \centering
    \includegraphics[width=10cm]{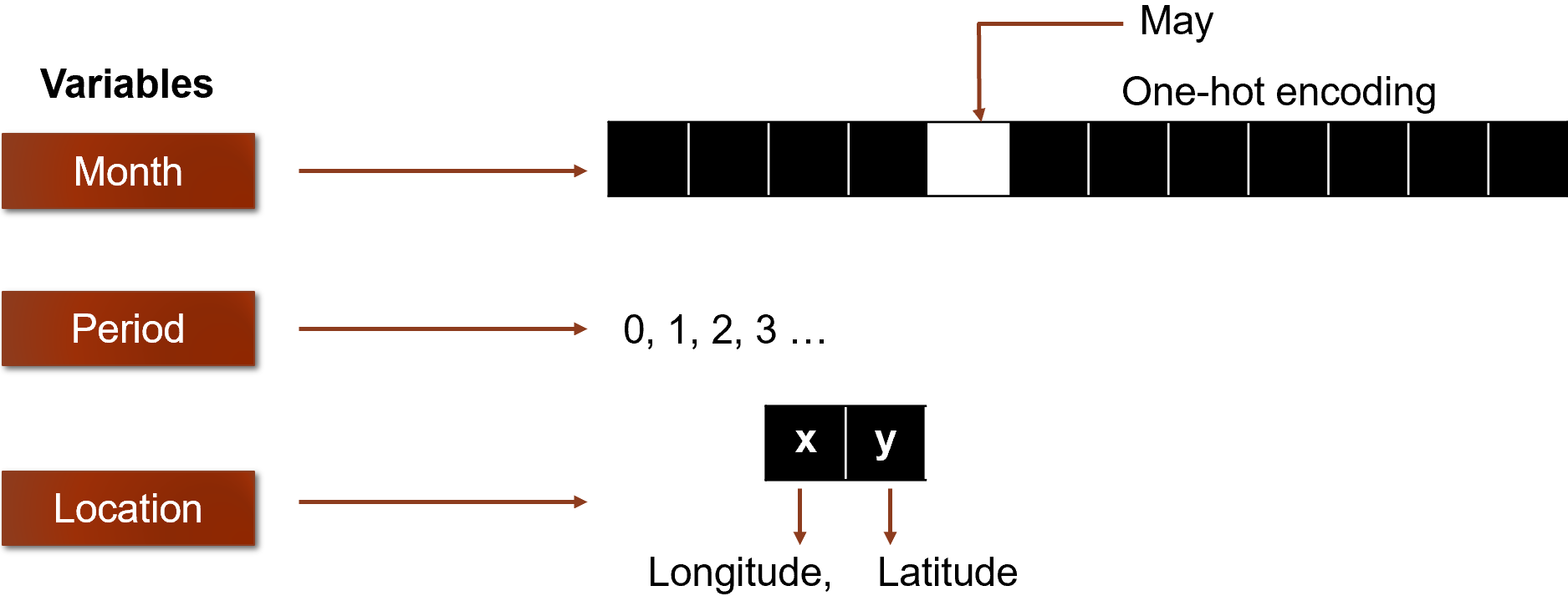}
    \caption{Each conditional variable is passed into a model block that transforms it into a higher dimensional learned embedding, resulting in the labels being mapped from $\mathbb{R}^{15}$ to $\mathbb{R}^{100}$}
    \label{fig:condition}
\end{figure}

\newpage

\section{Model Evaluation}\label{appendixb}
\subsection{SPAC'D}\label{appendixSPACD}
In the main text, we briefly touched on a key idea of this metric and the rationale for choosing the L1-norm as the preferred distance measure for this metric. We will now discuss the details for implementation. As shown in figure \ref{SPACD_process} below, each video frame is unraveled into a vector whose length is equal to the total number of pixels within the frame (here we have 64 pixels per $1\degree \times 1\degree$ region. Each pixel can be viewed as a feature within the sample, and the goal is to calculate the Pearson product-moment correlation coefficient (PPCC) matrices for each month and then calculate the distance between these matrices for the ground-truth and generated data. The choice for the L1-norm was not arbitrary but a more natural choice for this distance measure. If the Frobenius norm distance was used, then SPAC'D is no longer implicitly bounded by a fixed interval. Because the PPCC takes on values between -1 and 1, the maximum difference for any set of two PPCC matrices is 2. The frobenius norm of a matrix X is

\begin{equation}
    \norm{X}_{F} = \sqrt{\sum_{i=1}^{m}\sum_{j=1}^{n}\abs{x_{ij}}^2}
\end{equation}

implying that for a given set of samples, the maximum sum of distance of all the pixels must be 2N, because the maximum distance per pixel is 2. This means that the average per pixel distance depends on N, because the PPCC matrix has a shape of  $N \times N$ where $N$ is the total number of pixels per sample. Therefore by using the frobenius norm, the maximum average distance between the PPCC matrices is $\frac{\sqrt{2^2 \times N^2}}{N^2} = \frac{2N}{N^2} = \frac{2}{N}$, making it size-variant. However, the L1-norm does not suffer from this, making it fairly straightforward to implement as the metric remains bounded [0, 2] for any $N$.

\begin{figure}[h!]
    \centering
    \includegraphics[width=0.7\linewidth]{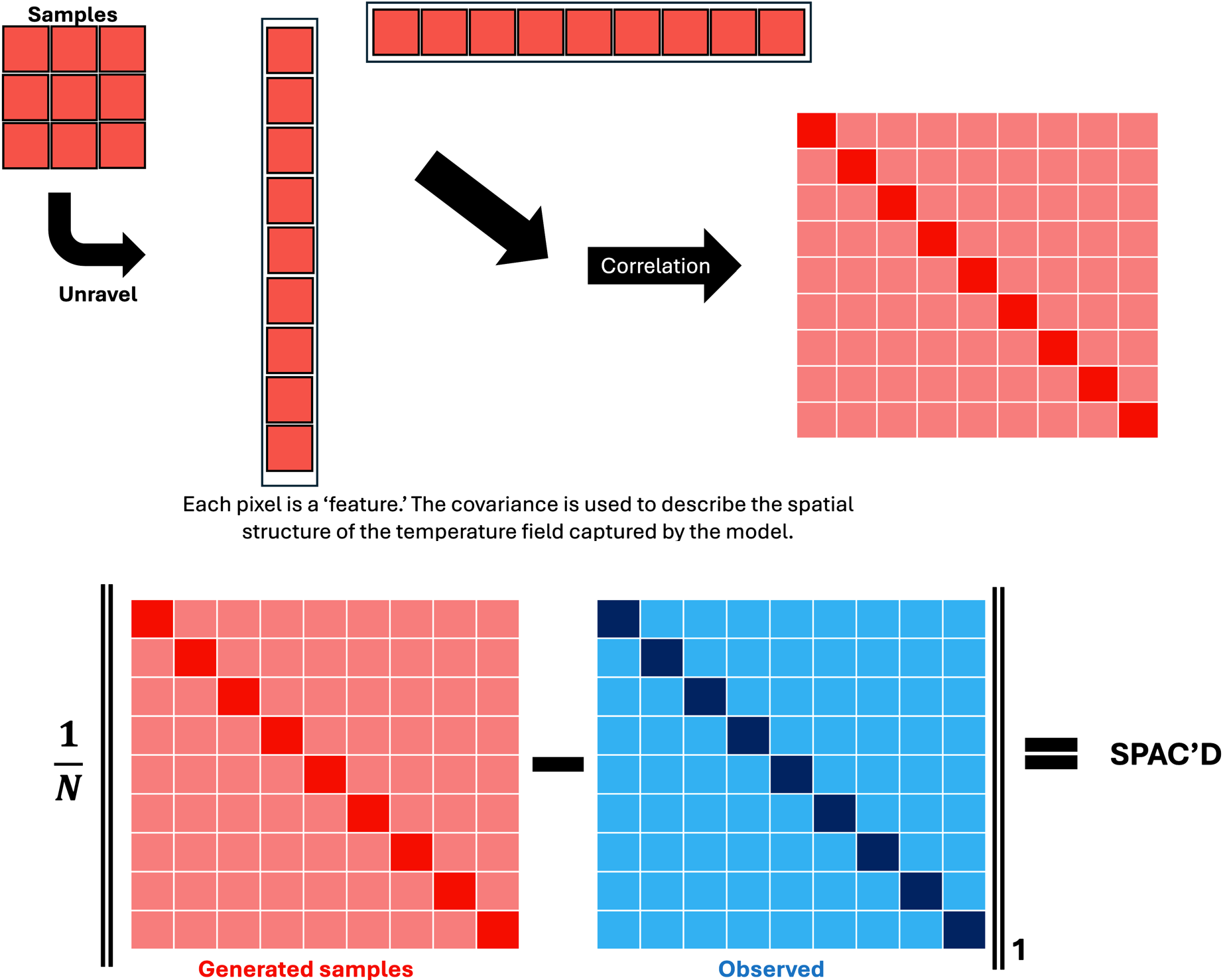}
    \caption{SPAC'D computation description. Using the feature vector, a correlation matrix is computed. The matrix L1 norm distance between the true and generated correlation matrices is calculated and reported as SPAC'D}
    \label{SPACD_process}
\end{figure}

\subsection{Temporal Integrity}\label{temporal_integrity}

\begin{figure}[h!]
    \centering
    \includegraphics[width=5.5cm]{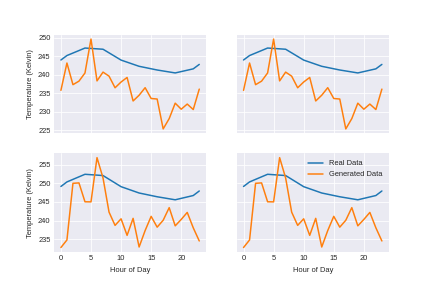}
    \includegraphics[width=5.5cm]{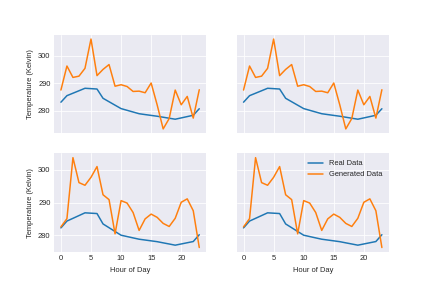}
    \includegraphics[width=5.5cm]{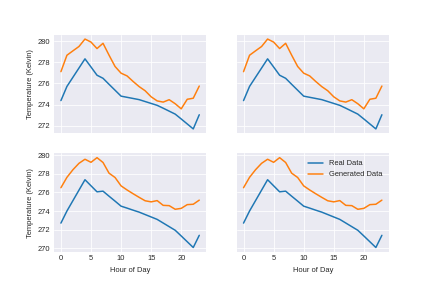}
    \includegraphics[width=5.5cm]{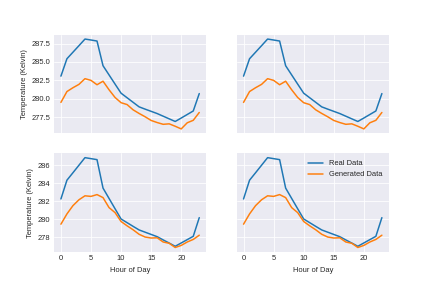}
    \caption{Visually observe that diurnal cycles generated which includes a temporal gradient penalty (rows 3 \& 4) are of higher quality than the models without the temporal gradient penalty (rows 1 \& 2). Note that the actual temperature values for these plots do not matter as we randomly sample from the real and generated examples to visually compare only the temporal patterns from the real data and the generated data}
    \label{fig:Temporal_penalty}
\end{figure}

For many practical purposes, visual inspection may be sufficient for evaluating the validity of timeseries generation, but for model comparisons and evaluation compared to a baseline, a standard process for comparison is integral. 
\begin{table}[h!]
  \caption{Effect of Temporal Gradient Penalty}
  \label{T_pen-table}
  \centering
  \begin{tabular}{lll}
    \toprule
    \multicolumn{3}{c}{Overall Gradient Distribution Statistics} \\ \hline
    Sample & Mean (K) & Standard Deviation (K) \\ \hline
    Real Data & -0.0056  &  0.9525\\
    Generated Data & -0.0973 & 2.5576\\
    Generated Data with Temporal Penalty & \textbf{-0.0071} & \textbf{1.1023}\\
    \bottomrule
  \end{tabular}
\end{table}

The table above shows the temporal gradient distribution values from the real and generated dataset for the period from 1979-1982. We observe that training with one discriminator suffices for the GAN to learn the overall temperature distributions, but the samples produced have temporal gradients that are not as true to the real data distribution, though we see that the overall diurnal cycle patterns are captured. Given this observation, we explored two variants for training the GAN. In the first variant, which we call ExWGAN-TGP, we add to the cost function a temporal gradient penalty, which yields the Generator (G) cost function:

\begin{equation}
    L_{g} = \underset{\tilde{\mathbf{T}}\sim\mathbb{P}_{g}}{\mathbb{E}}[D(\tilde{\mathbf{T}})] + \lambda_{tp} \norm{\frac{T(s, t) - T(s, t-1)}{\Delta t}}_{F} 
\end{equation}

where $\Delta t = 1$ hour for this work. $\lambda_{TP}$ represents the hyperparameter that can be adjusted for penalizing the temporal gradients directly. The second term in the G cost above represents the Frobenius Norm of the 3D temporal gradient matrix. Plots below \ref{fig:Temporal_penalty} show diurnal cycles without and with temporal penalty.

\subsection{FDTD tables}\label{FDTD tables}

\begin{table}[h!]
  \caption{FDTD (K) for Period 0 (1979 - 1982), Nevada area. Average distance 0.6 K/\degree C}
  \label{FDTD_NV_0}
  \centering
  \begin{tabular}{llllll}
    \toprule        
        Month & Real Mean & Generated Mean & Real STDEV & Generated STDEV & FDTD\\ 
        \hline
        January & 277.6915 & 278.1642 & 3.1502 & 3.2268 & 0.4789\\
        February & 279.9620 & 279.7024 & 3.4725 & 3.3151 & 0.3036\\
        March & 281.5201 & 281.7916 & 3.1634 & 3.0203 & 0.3069\\
        April & 287.1678 & 286.3463 & 4.2334 & 4.1516 & 0.8256\\
        May & 292.0256 & 291.6092 & 4.0923 & 3.6950 & 0.5755\\
        June & 298.0948 & 297.1719 & 4.4016 & 4.0736 & 0.9795\\
        July & 301.6974 & 301.0059 & 3.8504 & 3.5962 & 0.7368\\
        August & 300.8320 & 300.7011 & 3.6720 & 3.5070 & 0.2105\\
        September & 297.4704 & 296.1691 & 3.7490 & 3.5836 & 1.3118\\
        October & 289.1882 & 288.8969 & 4.4955 & 4.2833 & 0.3604\\
        November & 282.3828 & 281.6065 & 3.8058 & 3.3115 & 0.9204\\
        December & 279.8768 & 279.4155 & 3.2901 & 3.2362 & 0.4644\\
    \bottomrule
  \end{tabular}
\end{table}

\begin{table}
\centering
  \caption{FDTD (K) for Period 8 (2011 - 2014), Nevada area. Avg. distance 0.5192 K/\degree C}
  \label{FDTD_NV_8}
\begin{tabular}{l l l l l l } 
\toprule 
Month & Real Mean & Generated Mean & Real STDEV & Generated STDEV & FDTD\\ 
\hline
January & 277.8183& 277.1743& 2.7726 & 2.8178 & 0.6457\\ 
February & 280.2512& 279.6572 & 3.1188 & 3.1590 & 0.5953\\ 
March & 283.0497 & 282.7167 & 3.4341 & 3.4655 & 0.3345\\ 
April & 288.0169 & 287.5964 & 3.9681 & 4.0082 & 0.4225\\ 
May & 292.9910 & 292.4301 & 4.1578 & 4.1915 & 0.5619\\ 
June & 297.6616 & 297.3702 & 4.3225 & 4.2530 & 0.2995\\ 
July & 301.0671 & 300.9308 & 3.7384 & 3.7202 & 0.1374\\ 
August & 301.2599 & 300.7459 & 3.6445 & 3.6193 & 0.5146\\ 
September & 297.2854 & 296.2444 & 3.6689& 3.6605 & 1.0410\\ 
October & 290.4260 & 289.2941 & 4.1681& 4.2065 & 1.1325\\ 
November & 280.7239 & 280.6468 & 3.8598 & 3.7307 & 0.1503\\ 
December & 277.7062 & 277.3202 & 2.7934 & 2.8802 & 0.3956\\ 
\bottomrule
\end{tabular}
\end{table}

\begin{table}
\caption{FDTD (K) for Period 3 (1991 - 1994), Portland, Oregon Area. Avg. distance 0.2731 K/\degree C}
\label{FDTD_POR_3}
\centering
\begin{tabular}{l l l l l l}
\toprule
Month & Real Mean & Generated Mean & Real STDEV & Generated STDEV & FDTD\\
\hline
January & 277.6905 & 277.5804 & 2.8445 & 2.6762 & 0.2010 \\
February & 279.2674 & 278.8922 & 2.6814 & 2.6864 & 0.3753 \\
March & 281.3473 & 281.1332 & 2.8626 & 2.6976 & 0.2704 \\
April & 282.6404 & 282.9748 & 2.8337 & 2.7582 & 0.3428 \\
May & 286.3006 & 286.3374 & 3.3921 & 3.3240 & 0.0775 \\
June & 287.5147 & 288.0268 & 3.2757 & 3.2406 & 0.5134 \\
July & 290.4595 & 290.7853 & 3.3119 & 3.3702 & 0.3309 \\
August & 291.1990 & 291.4042 & 3.3491 & 3.4130 & 0.2149 \\
September & 289.7109 & 289.4515 & 3.6540 & 3.6439 & 0.2596 \\
October & 285.3286 & 285.3490 & 3.2192 & 3.2973 & 0.0806 \\
November & 279.2227 & 279.7910 & 2.5087 & 2.4875 & 0.5687 \\
December & 277.7254 & 277.7549 & 2.2562 & 2.2863 & 0.0421 \\
\bottomrule
\end{tabular}
\end{table}
\FloatBarrier

\subsection{More Q-Q Envelopes}\label{more_QQ_plots}
\begin{figure}
    \centering 
    \includegraphics[width=0.7\linewidth]{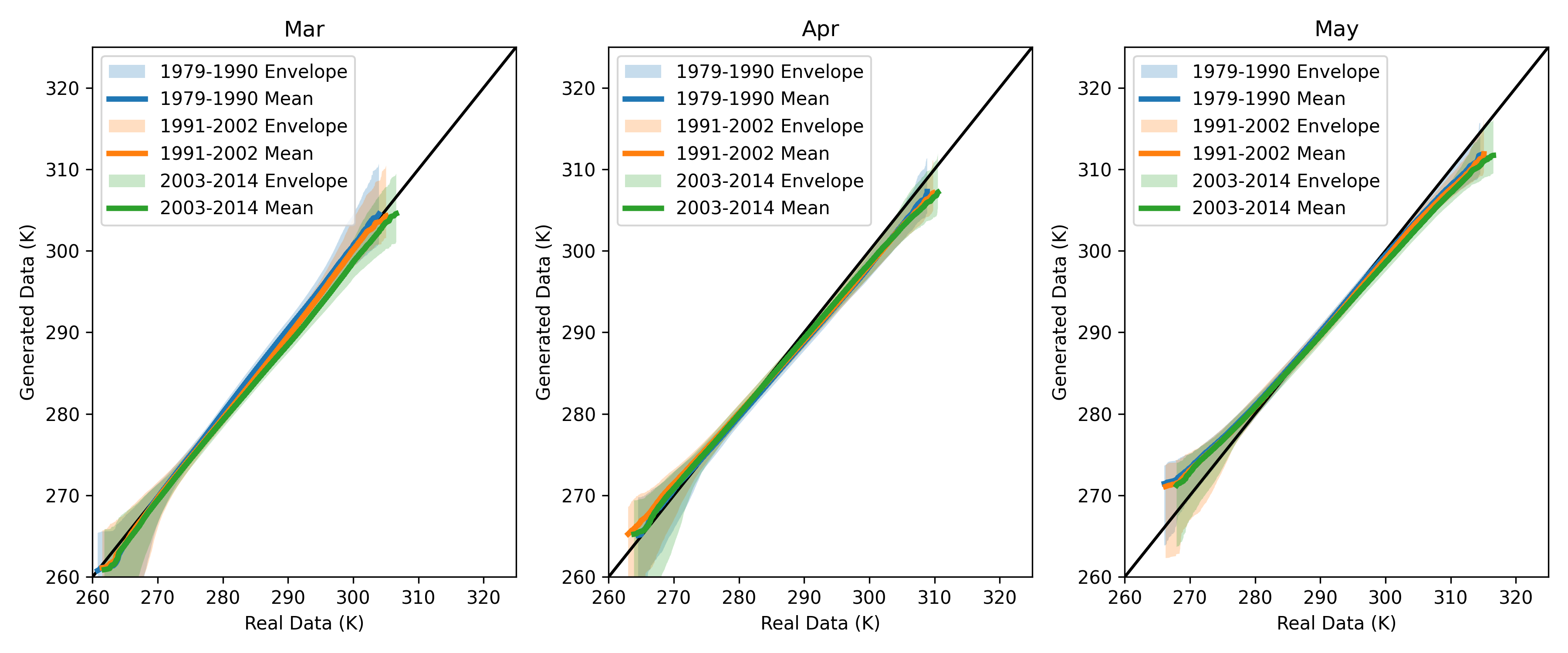}
    \includegraphics[width=0.7\linewidth]{Images/qq_new_Nevada_Summer.png}
    \includegraphics[width=0.7\linewidth]{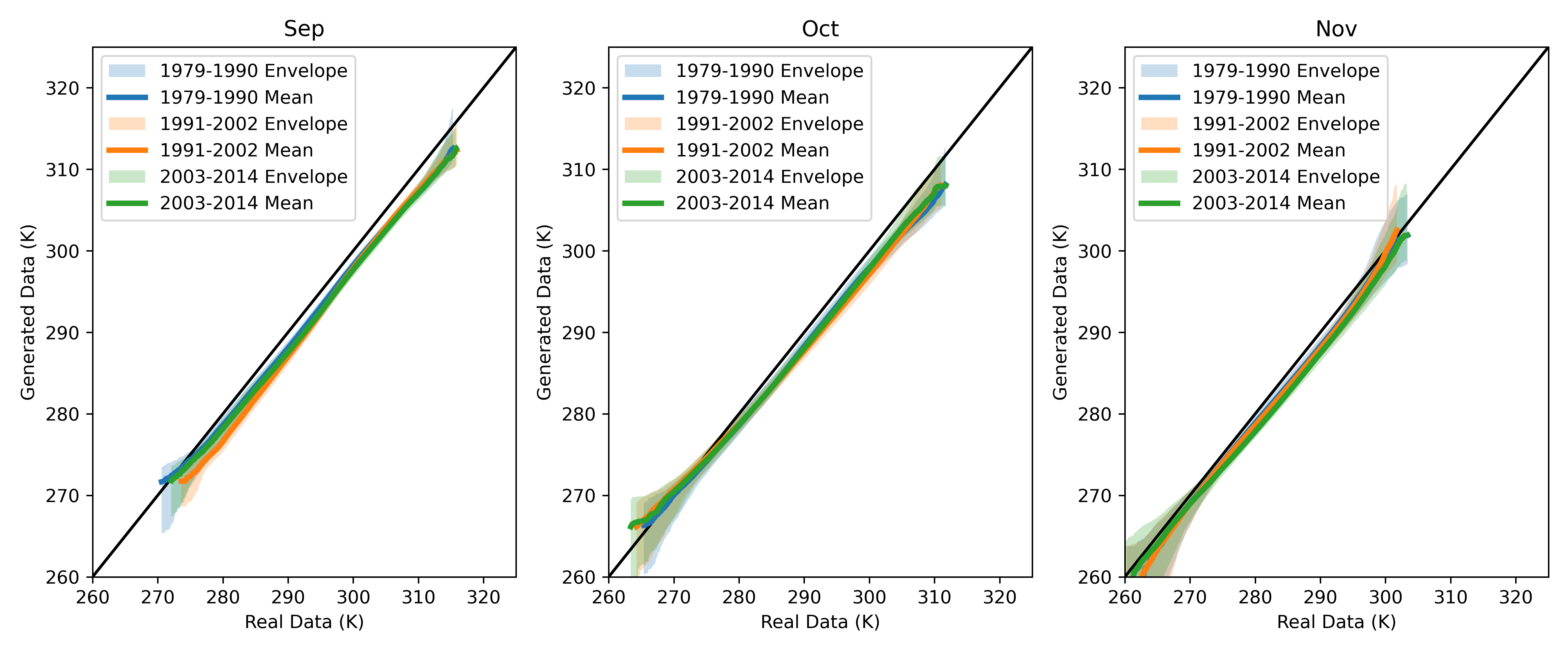}
    \includegraphics[width=0.7\linewidth]{Images/qq_new_Nevada_Winter.png}
    \caption{Q-Q envelopes for Nevada region using 100 TemperatureGAN generated samples}
    \label{fig:qq_nv_all}
\end{figure}

\begin{figure}
    \centering 
    \includegraphics[width=0.7\linewidth]{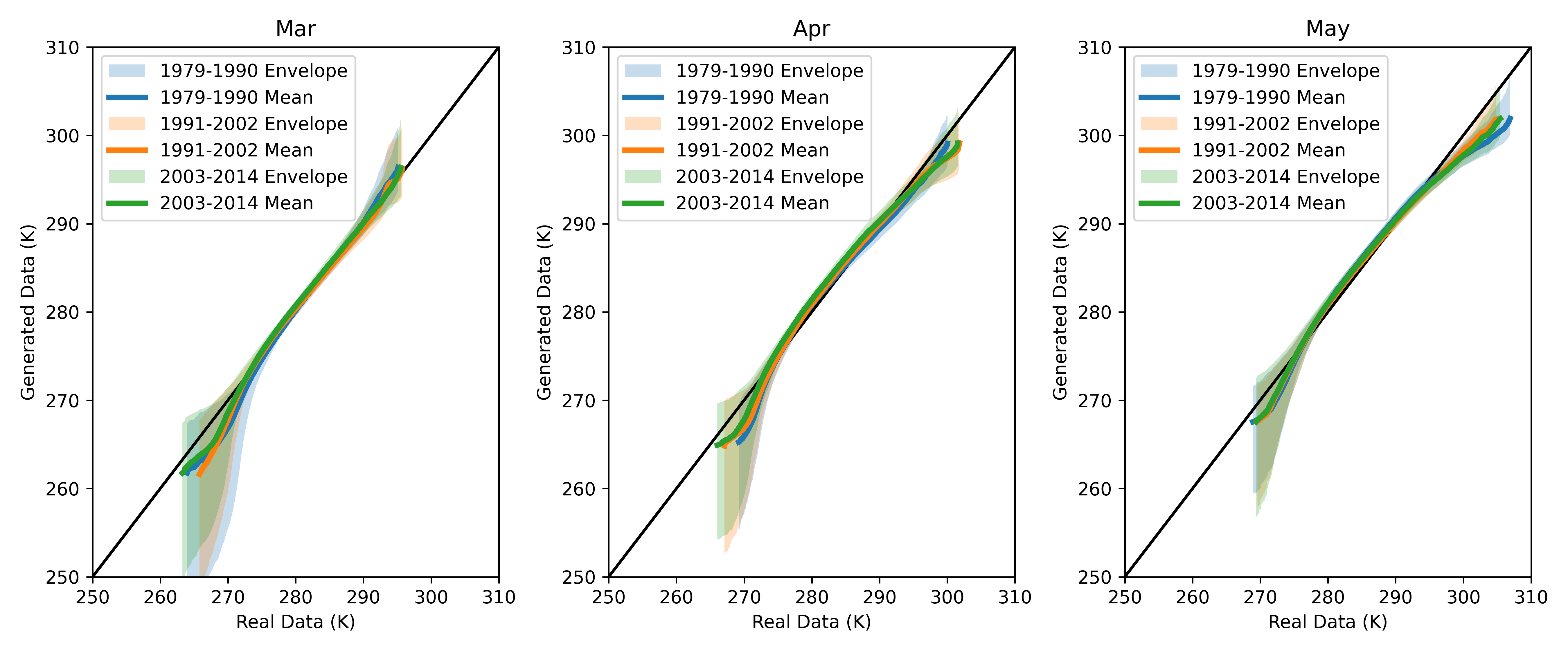}
    \includegraphics[width=0.7\linewidth]{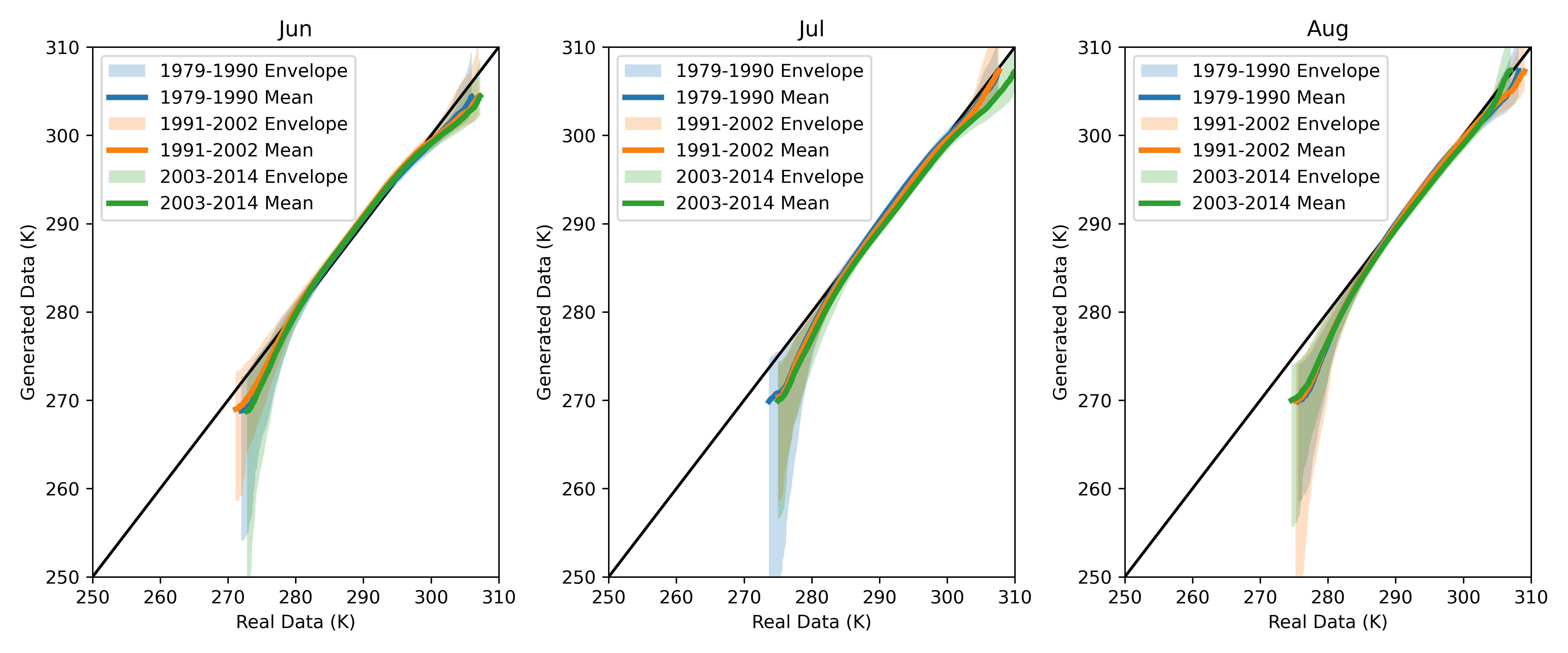}
    \includegraphics[width=0.7\linewidth]{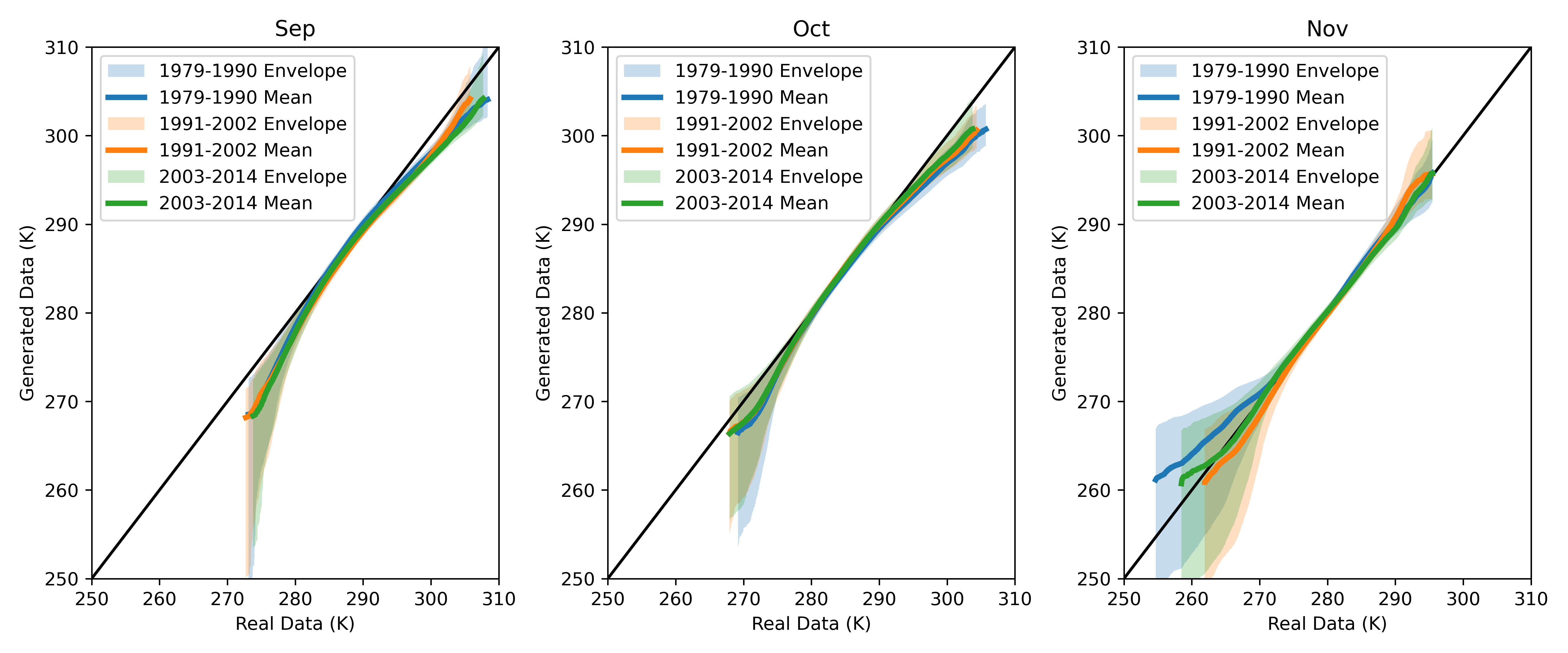}
    \includegraphics[width=0.7\linewidth]{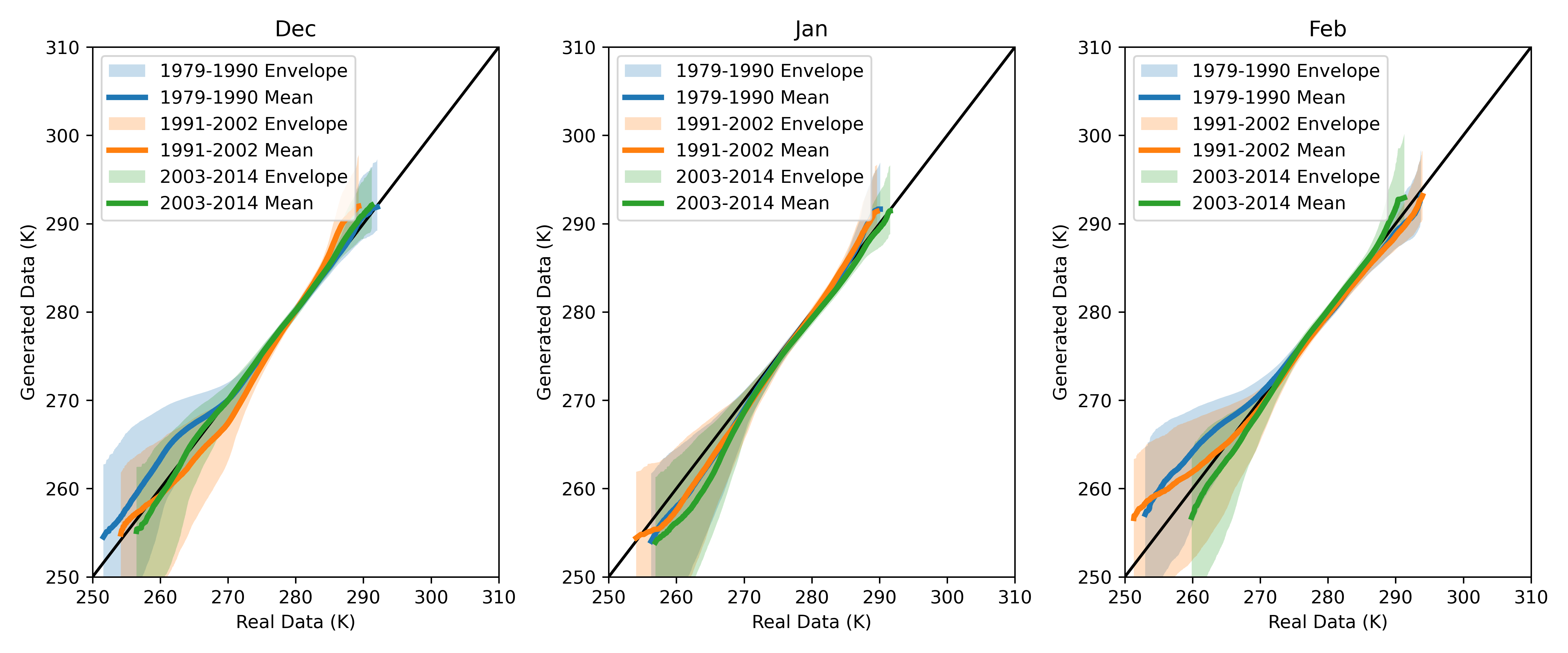}
    \caption{Q-Q envelopes for Portland region using 100 TemperatureGAN generated samples}
    \label{fig:qq_pt_all}
\end{figure}

\begin{figure}
    \centering 
    \includegraphics[width=0.7\linewidth]{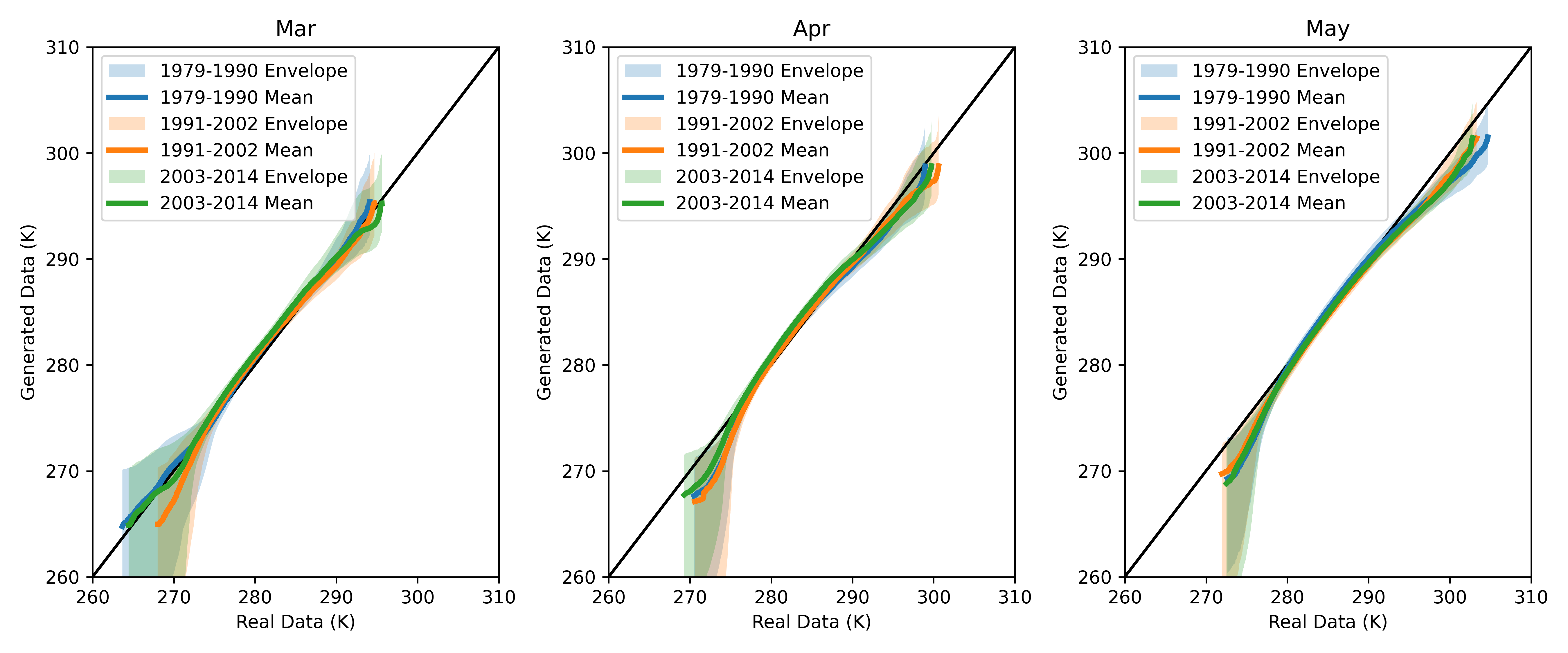}
    \includegraphics[width=0.7\linewidth]{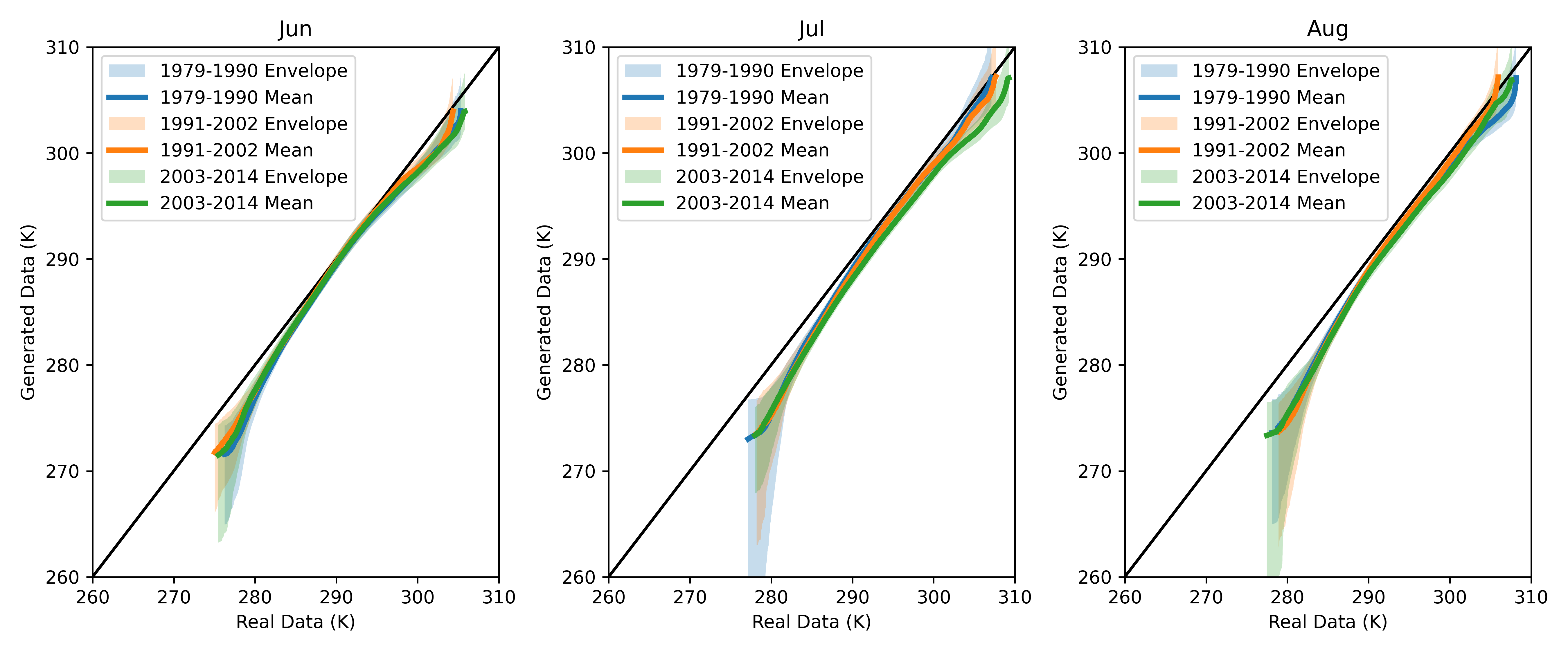}
    \includegraphics[width=0.7\linewidth]{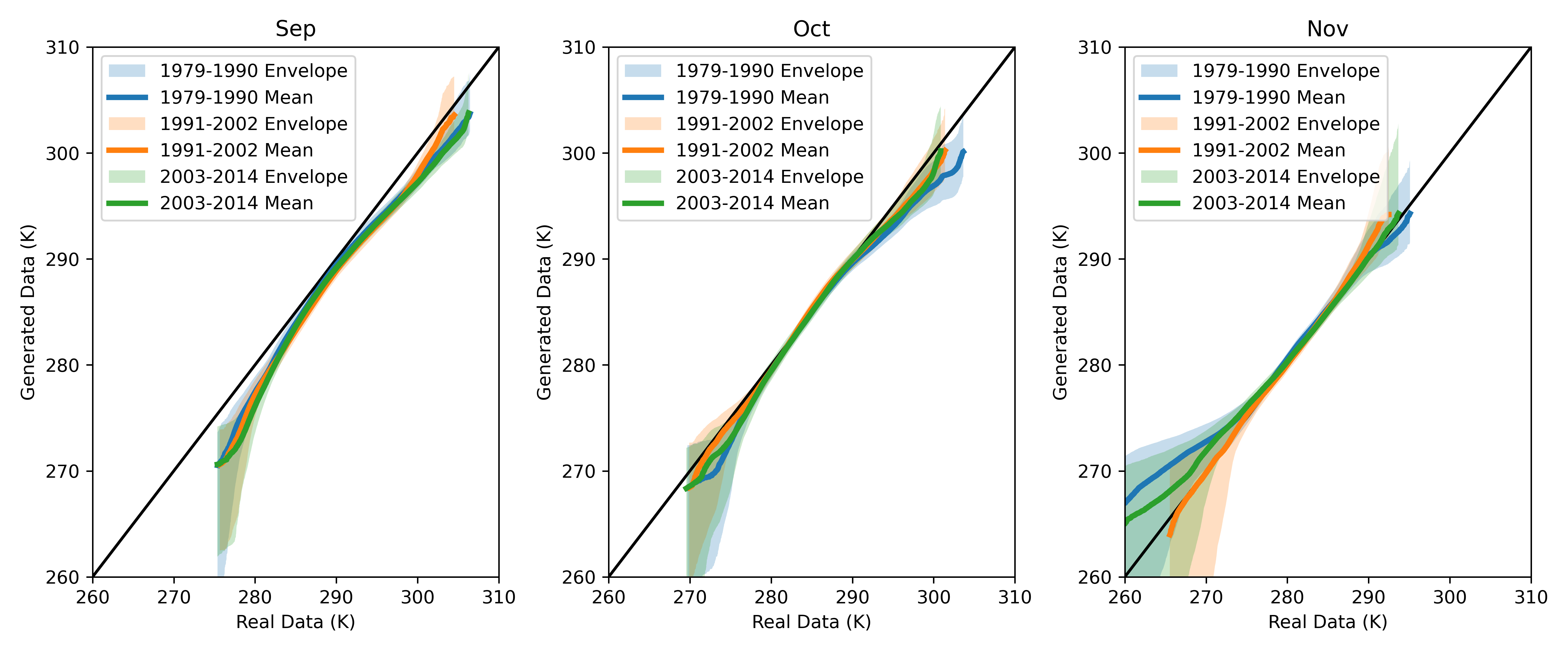}
    \includegraphics[width=0.7\linewidth]{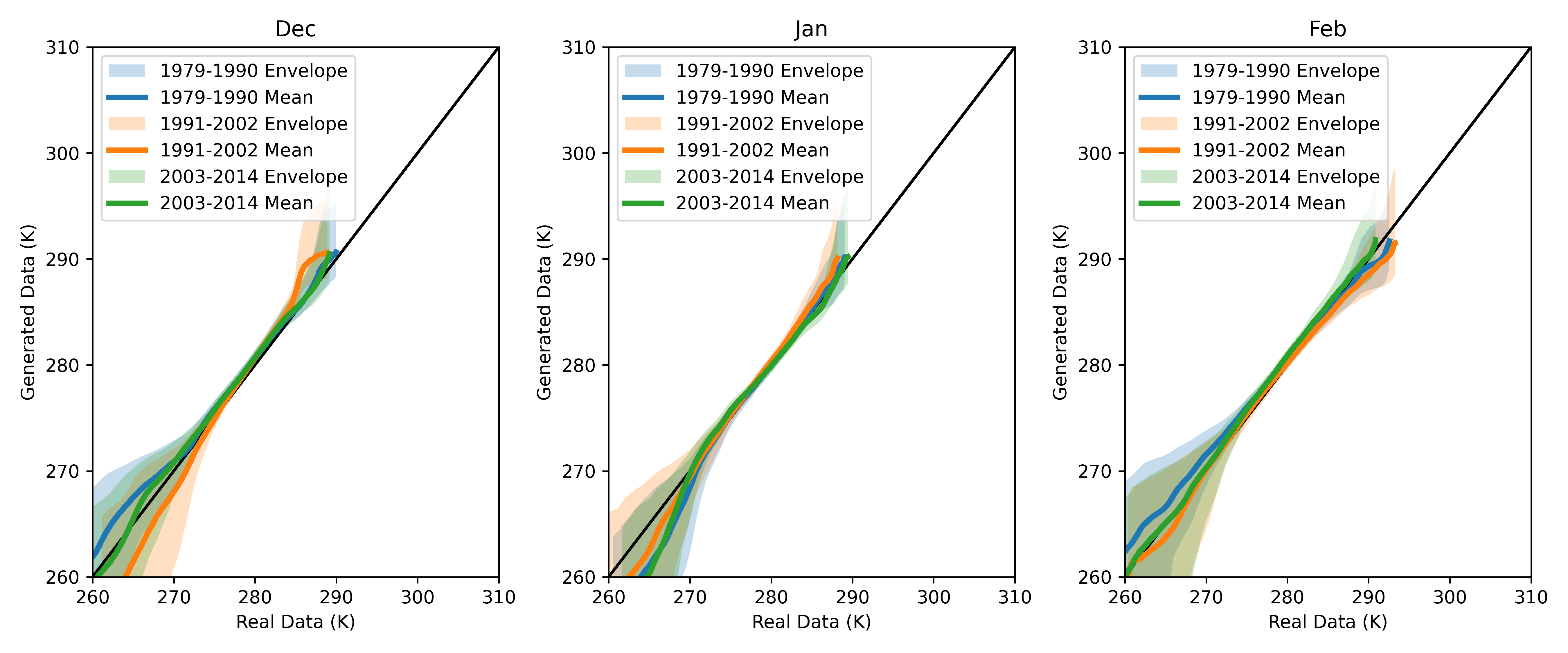}
    \caption{Q-Q Plot envelopes for Washington region using 100 TemperatureGAN generated samples}
    \label{fig:qq_wa_all}
\end{figure}

\begin{figure}
    \centering 
    \includegraphics[width=0.7\linewidth]{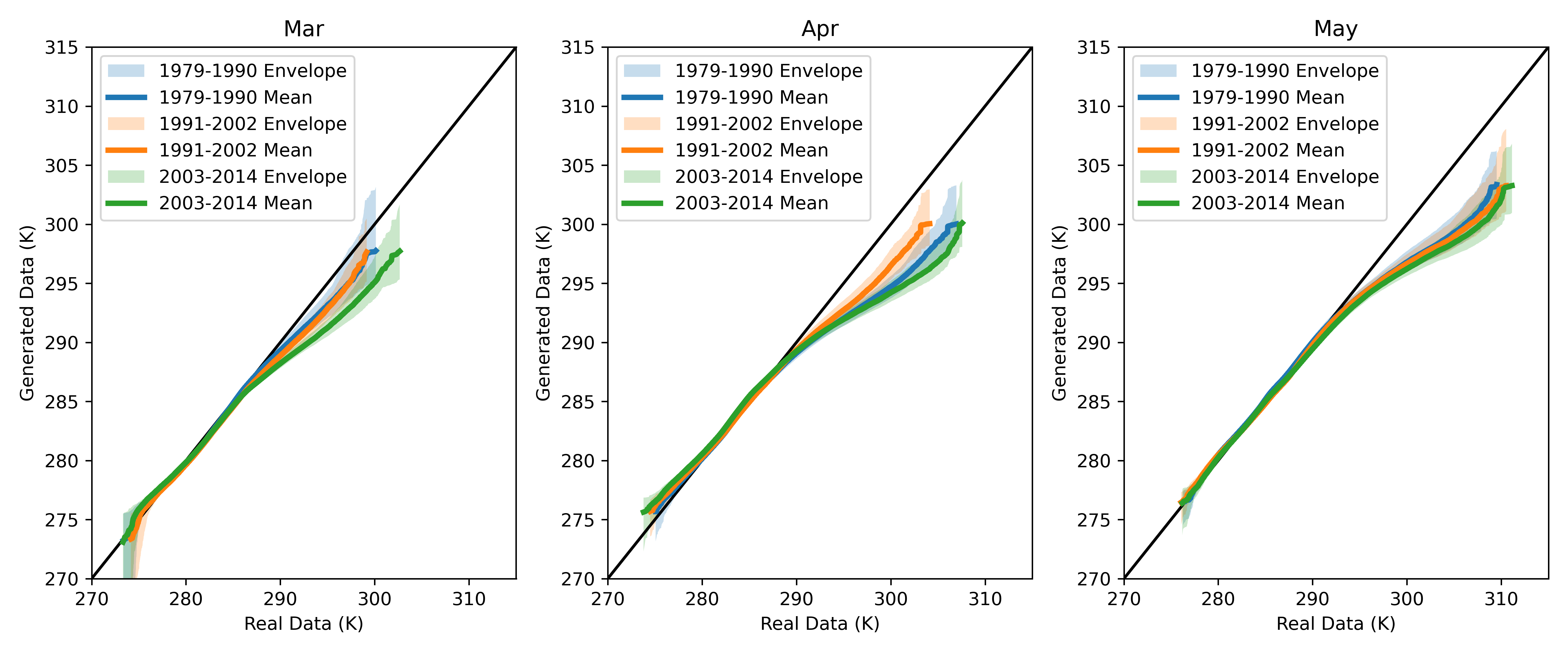}
    \includegraphics[width=0.7\linewidth]{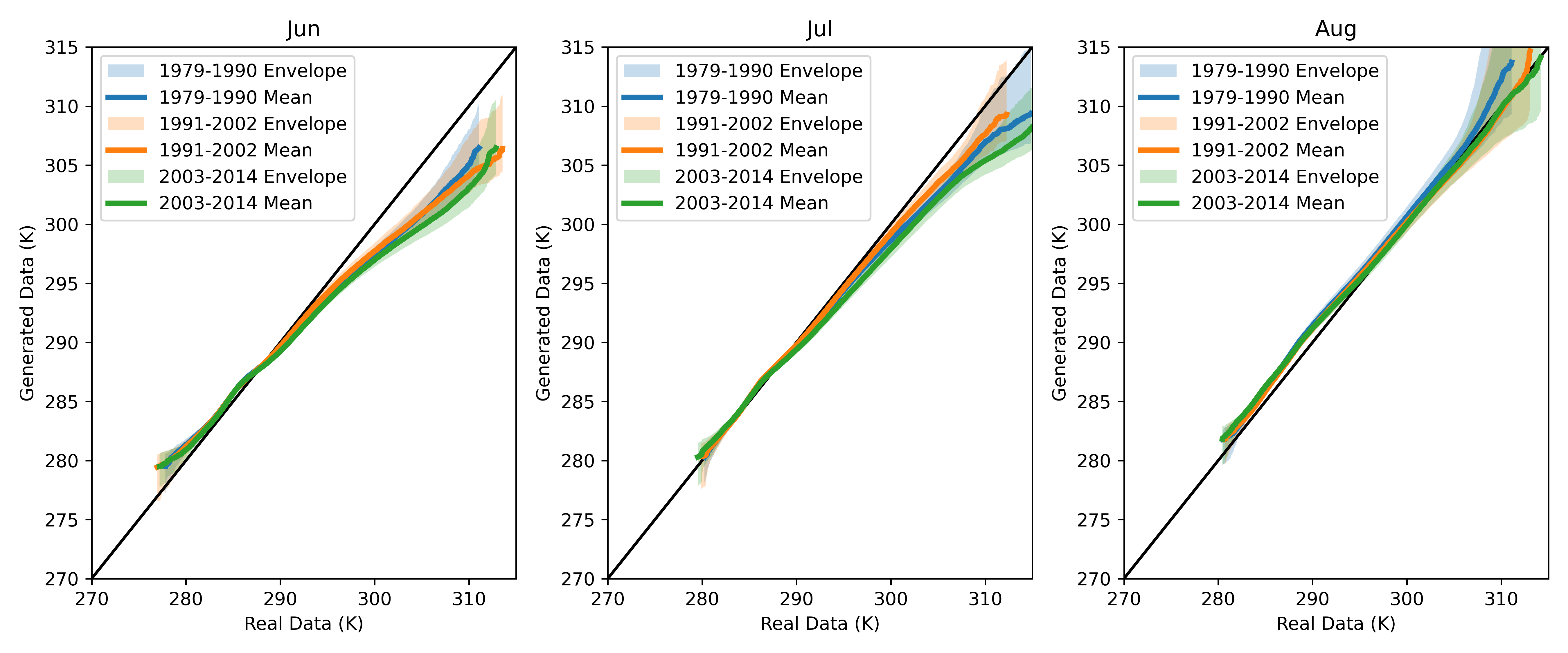}
    \includegraphics[width=0.7\linewidth]{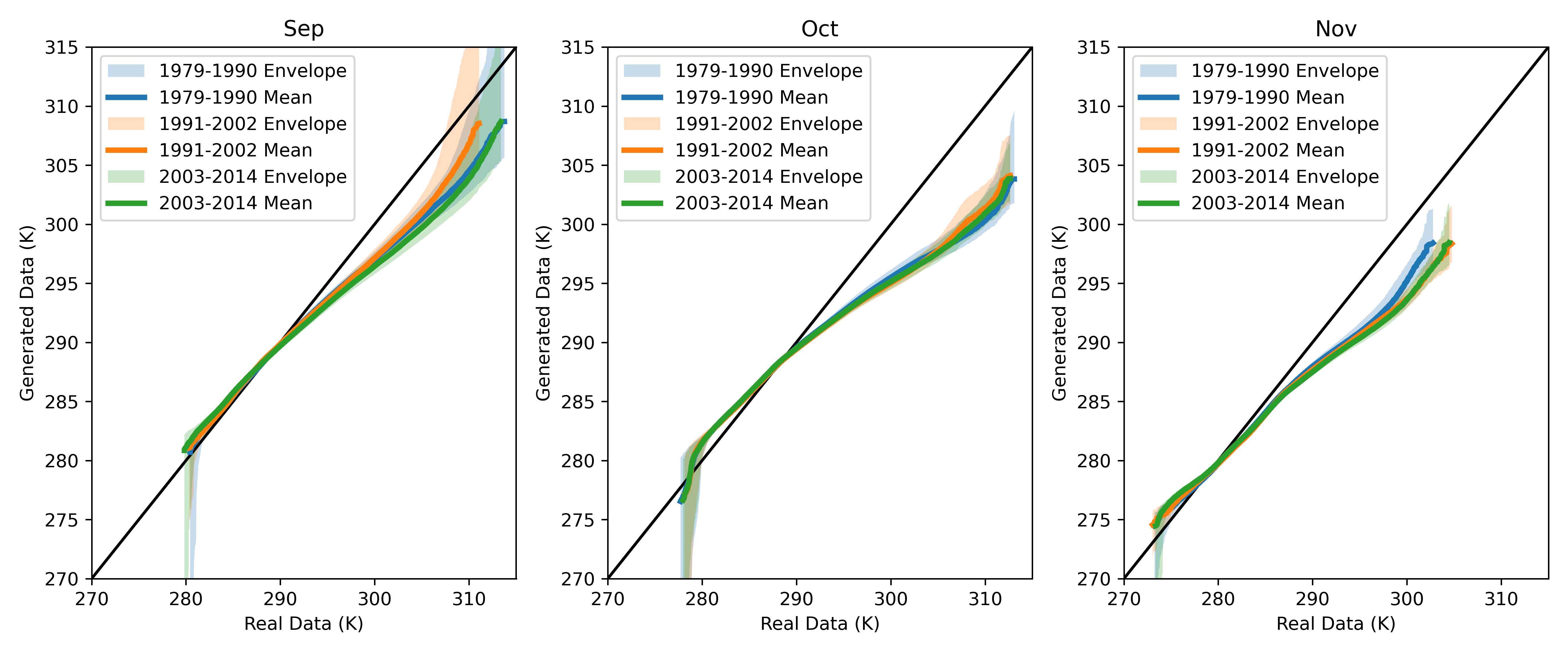}
    \includegraphics[width=0.7\linewidth]{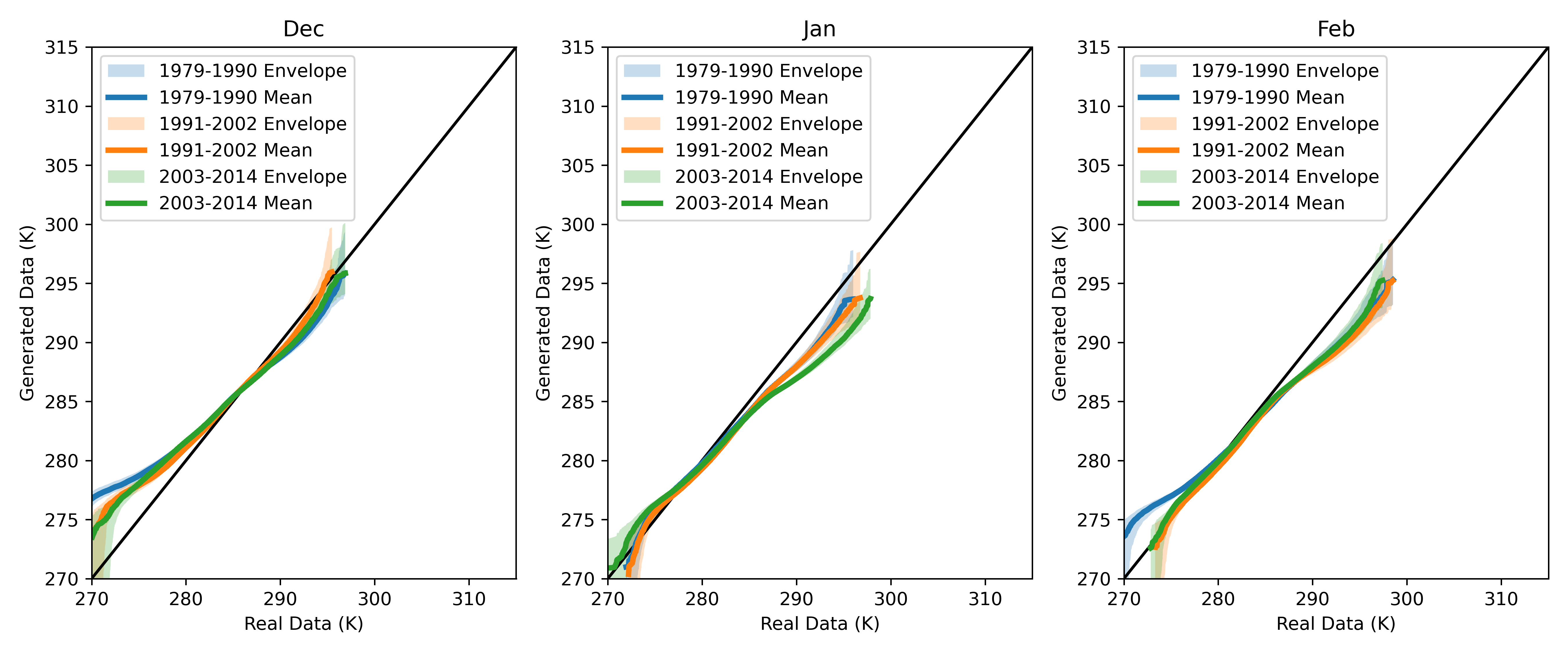}
    \caption{Q-Q Plot envelopes for San Francisco Bay region using 100 TemperatureGAN generated samples}
    \label{fig:qq_sf_all}
\end{figure}
\FloatBarrier

\subsection{Sampled distributions}\label{distribution_plots}
\begin{figure}[h!]
    \centering
    \includegraphics[width=6cm]{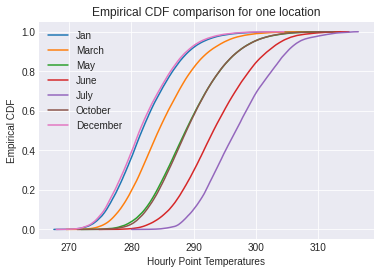}
     \includegraphics[width=6cm]{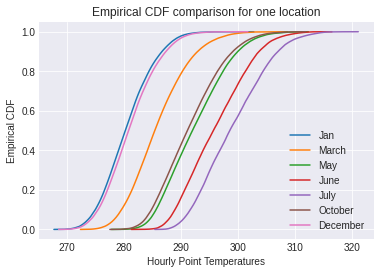}
    \caption{Monthly ECDF plots for Los Angeles (LA). Observe the distributions exhibit a rightward shift for hotter months, especially with the tails stretching, indicating more warming over 24 years}
    \label{fig:ECDF_1}
\end{figure}

\begin{figure}[h!]
    \centering
    \includegraphics[width=4cm]{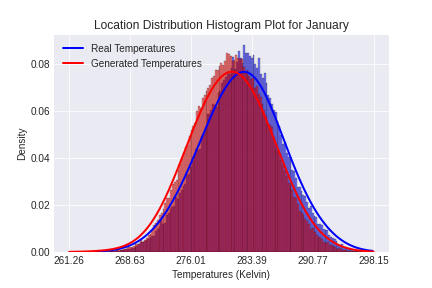}
    \includegraphics[width=4cm]{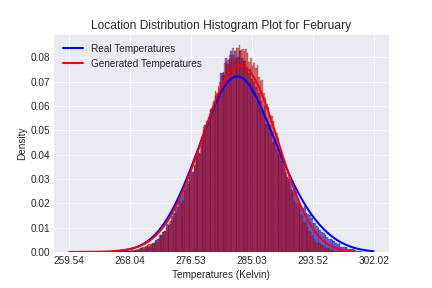}
    \includegraphics[width=4cm]{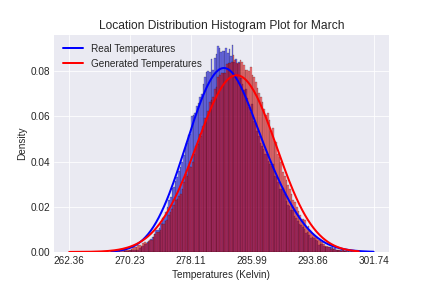}
    \includegraphics[width=4cm]{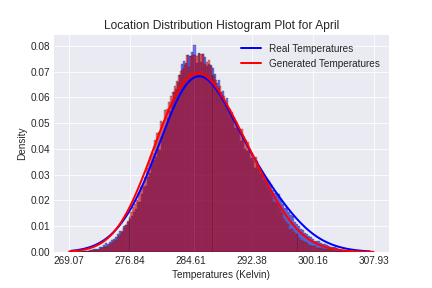}
    \includegraphics[width=4cm]{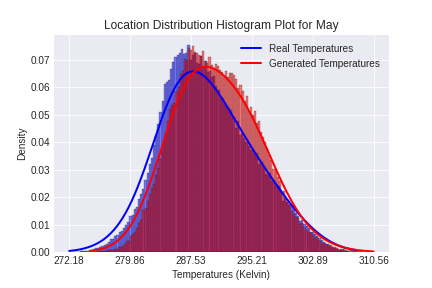}
    \includegraphics[width=4cm]{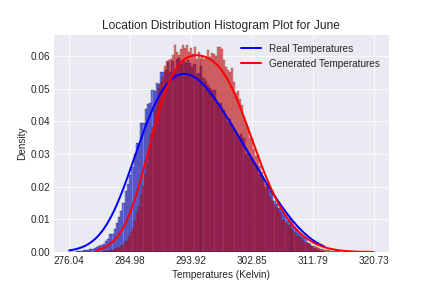}
    \includegraphics[width=4cm]{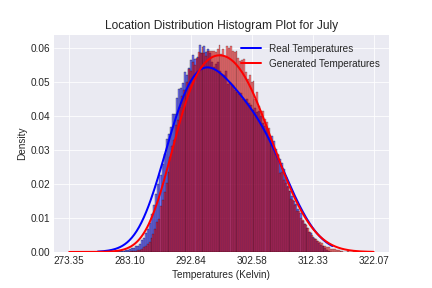}
    \includegraphics[width=4cm]{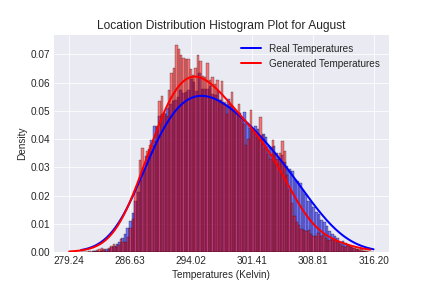}
    \includegraphics[width=4cm]{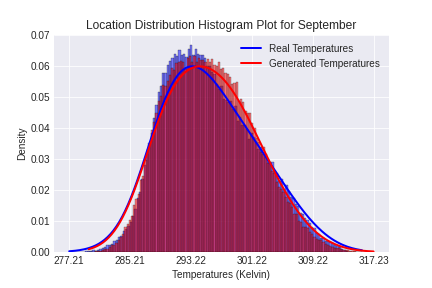}
    \includegraphics[width=4cm]{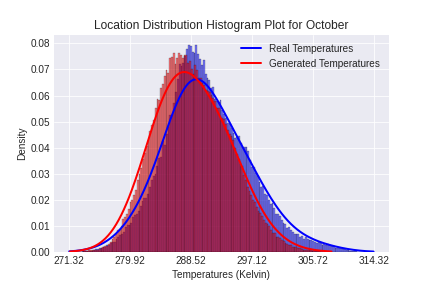}
    \includegraphics[width=4cm]{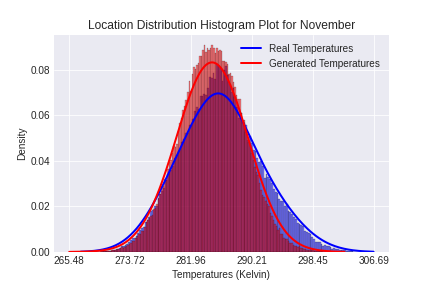}
    \includegraphics[width=4cm]{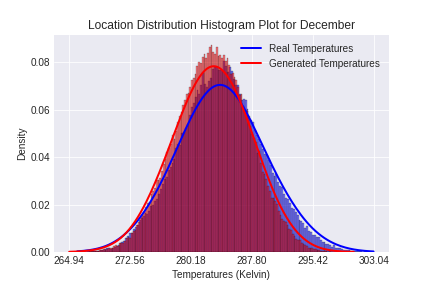}
    \caption{Los Angeles county temperature sample distributions for each month using the kernel density estimate plots for period $k_0$ (1979 - 1982)}
    \label{fig:base_lbl_dist_plots_LA}
\end{figure}
\FloatBarrier

\begin{figure}[h!]
    \centering
    \includegraphics[width=4cm]{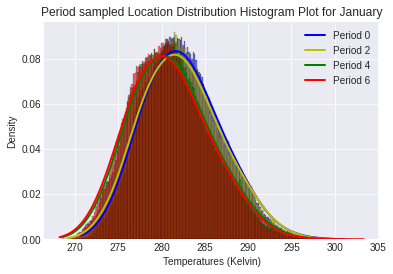}
    \includegraphics[width=4cm]{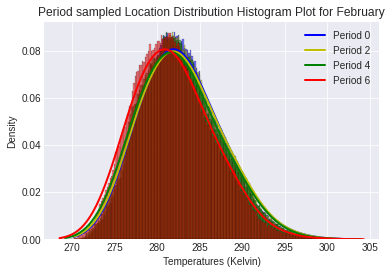}
    \includegraphics[width=4cm]{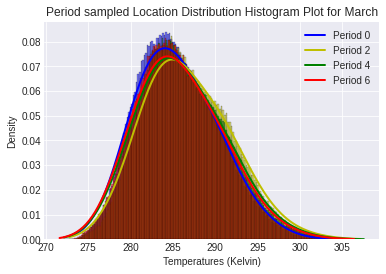}
    \includegraphics[width=4cm]{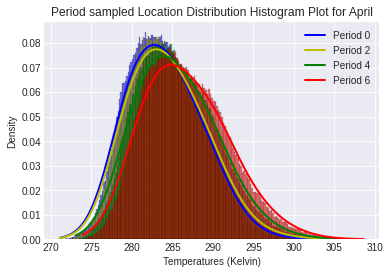}
    \includegraphics[width=4cm]{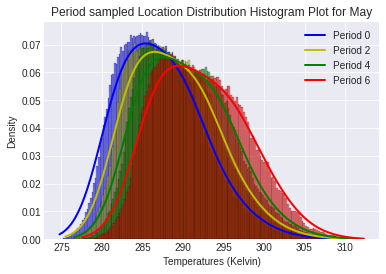}
    \includegraphics[width=4cm]{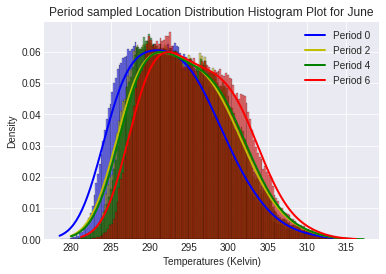}
    \includegraphics[width=4cm]{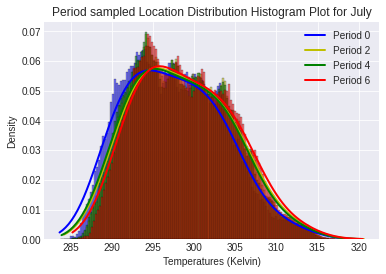}
    \includegraphics[width=4cm]{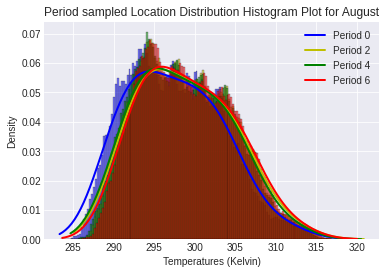}
    \includegraphics[width=4cm]{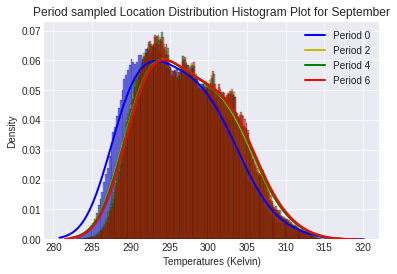}
    \includegraphics[width=4cm]{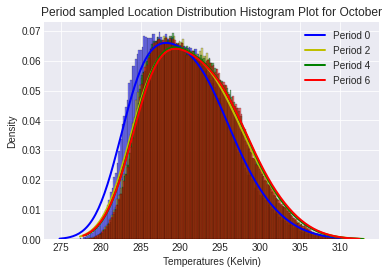}
    \includegraphics[width=4cm]{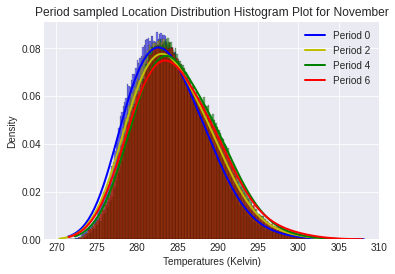}
    \includegraphics[width=4cm]{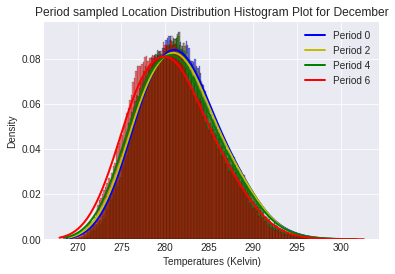}
    \caption{Generated monthly distributions while varying the period variable for Los Angeles (LA) County region. Plots show $k_0, k_2, k_4, k_6$}
    \label{fig:period_sampling}
\end{figure}

\begin{figure}[h]
    \centering
    \includegraphics[width=4cm]{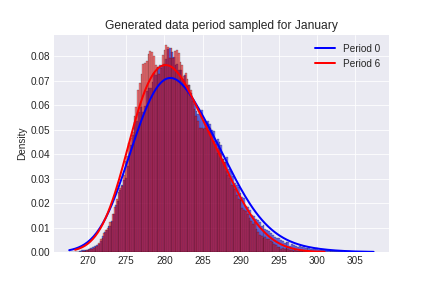}
    \includegraphics[width=4cm]{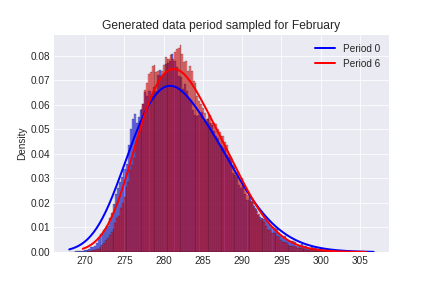}
    \includegraphics[width=4cm]{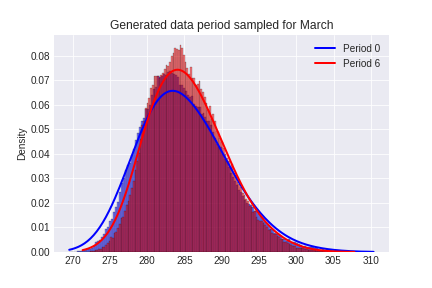}
    \includegraphics[width=4cm]{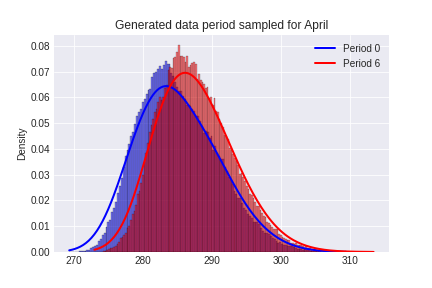}
    \includegraphics[width=4cm]{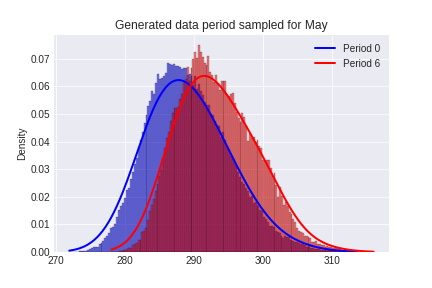}
    \includegraphics[width=4cm]{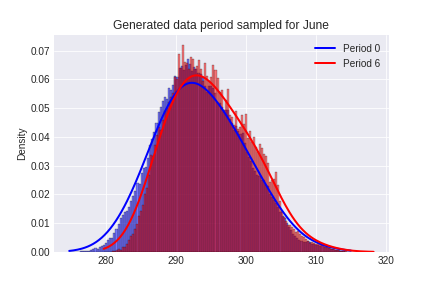}
    \includegraphics[width=4cm]{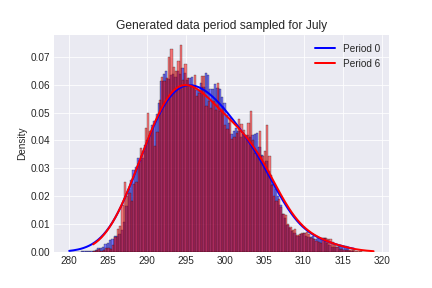}
    \includegraphics[width=4cm]{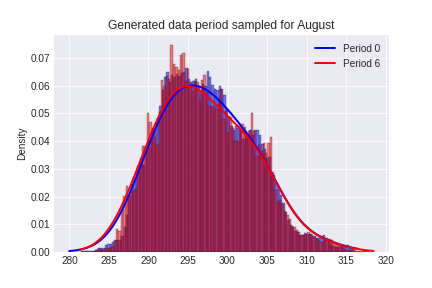}
    \includegraphics[width=4cm]{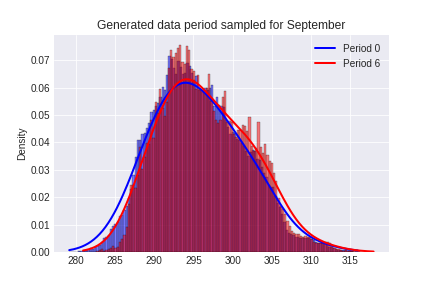}
    \caption{Images show month-region-period based sample distributions from the TemperatureGAN. One can observe a positive distribution shift for some months. Comparing to real data distribution below, one can see the generative model captures the distribution shifts from $k_0$ to $k_6$ that exist in specific months within that region. $k_0$ (blue, 1979-1982), $k_6$ (red, 2003 - 2006) for LA County region}
    \label{fig:dis_shift_gen1}
\end{figure}

\begin{figure}[h]
    \centering
    \includegraphics[width=4cm]{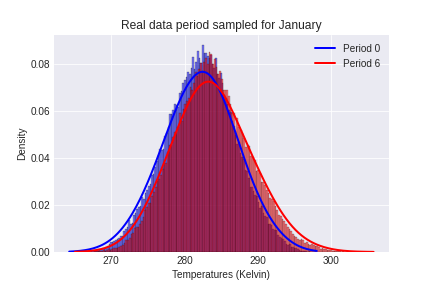}
    \includegraphics[width=4cm]{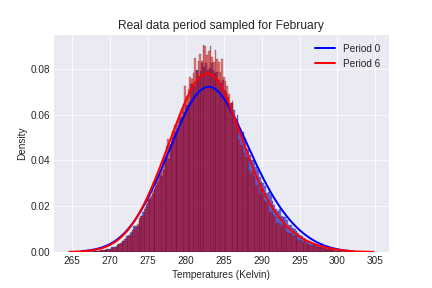}
    \includegraphics[width=4cm]{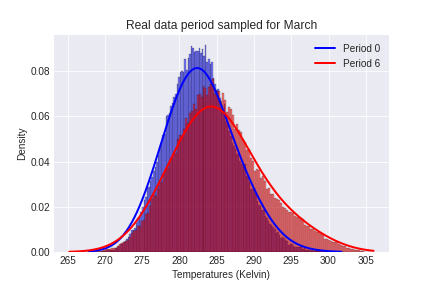}
    \includegraphics[width=4cm]{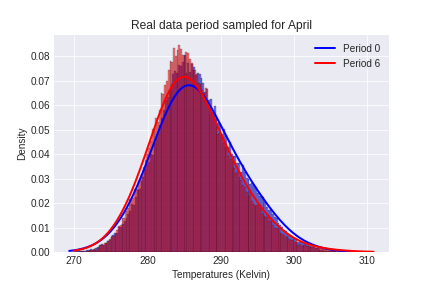}
    \includegraphics[width=4cm]{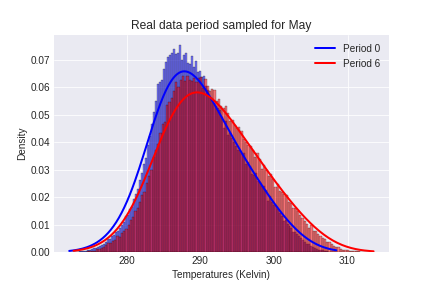}
    \includegraphics[width=4cm]{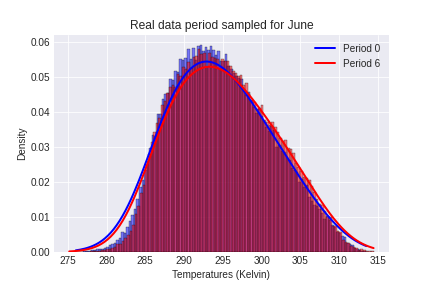}
    \includegraphics[width=4cm]{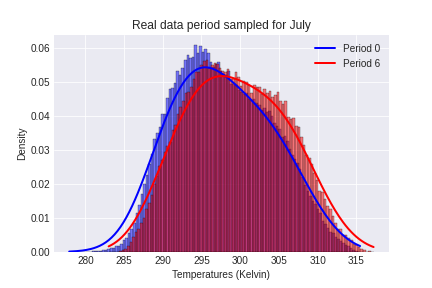}
    \includegraphics[width=4cm]{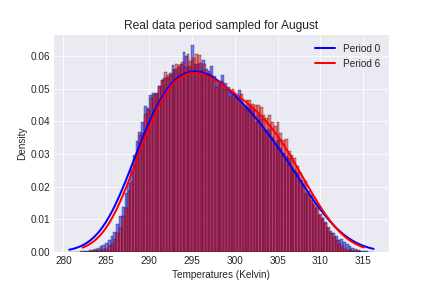}
    \includegraphics[width=4cm]{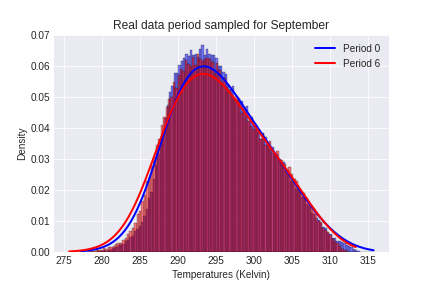}
    \caption{Images show month-region-period based sample distributions from the ground-truth data. $k_0$ (blue, 1979-1982), $k_6$ (red, 2003 - 2006) in LA County region}
    \label{fig:approach4}
\end{figure}


\end{appendix}







\end{Backmatter}
\end{document}